%% file: main.tex
\def\isarxiv{1} %%% for icml submission version, we comment this line

\ifdefined\isarxiv
\documentclass[11pt]{article}

\usepackage[numbers]{natbib}

\else

\documentclass{article} % For LaTeX2e
\usepackage[submission]{colm2025_conference}
\fi

\ifdefined\isarxiv

\usepackage{amsmath}
\usepackage{amsthm}
\usepackage{amssymb}
\usepackage{algorithm}
\usepackage{subfig}
\usepackage{algpseudocode}
\usepackage{graphicx}
\usepackage{grffile}
\usepackage{wrapfig,epsfig}
\usepackage{url}
\usepackage{xcolor}
\usepackage{epstopdf}

\usepackage{bbm}
\usepackage{dsfont}
\usepackage{todonotes} % Jiahao: Added for CoLM deadline, Mar 24, 2025
\usepackage{multirow} % Jiahao: Added for CoLM deadline, Mar 24, 2025
\usepackage{booktabs} % Jiahao: Added for CoLM deadline, Mar 24, 2025

\else
\usepackage{amsmath}
\usepackage{amsthm}
\usepackage{amssymb}
\usepackage{algorithm}
\usepackage{subfig}
\usepackage{algpseudocode}
\usepackage{graphicx}
\usepackage{grffile}
\usepackage{wrapfig,epsfig}
\usepackage{microtype}
\usepackage{hyperref}
\usepackage{url}
\usepackage{xcolor}
\usepackage{epstopdf}
\usepackage{bbm}
\usepackage{dsfont}
\usepackage{booktabs}
\usepackage{lineno}

\usepackage{todonotes} % Jiahao: Added for CoLM deadline, Mar 24, 2025
\usepackage{multirow} % Jiahao: Added for CoLM deadline, Mar 24, 2025

\fi
 
\allowdisplaybreaks

\ifdefined\isarxiv

\usepackage{tikz}
\usepackage{hyperref}  %%% arxiv don't allow this.
\hypersetup{colorlinks=true,citecolor=blue,linkcolor=blue} %%% Zhao : maybe we should comment this in submission.
\usetikzlibrary{arrows}
\usepackage[margin=1in]{geometry}

\else

\definecolor{darkblue}{rgb}{0, 0, 0.5}
\hypersetup{colorlinks=true, citecolor=darkblue, linkcolor=darkblue, urlcolor=darkblue}

\fi

\graphicspath{{./figs/}}

\theoremstyle{plain}
\newtheorem{theorem}{Theorem}[section]

\newtheorem{observation}[theorem]{Observation}

\newcounter{promptcounter} % Jiahao: added on March 27, 2025, for CoLM due
\newcounter{pexamplecounter}[promptcounter] % Jiahao: added on March 27, 2025, for CoLM due
\renewcommand{\thepexamplecounter}{\thepromptcounter.\arabic{pexamplecounter}} % Jiahao: added on March 27, 2025, for CoLM due

\newenvironment{ptemplate}[1][]{
  \refstepcounter{promptcounter}
  \setcounter{pexamplecounter}{0}
  \todo[inline, color=gray!10]{\textbf{Prompt Template \thepromptcounter}: #1}
}{}
\newenvironment{pexample}[1][]{
  \refstepcounter{pexamplecounter}
  \todo[inline, color=gray!10]{\textbf{Example Prompt \thepexamplecounter}: #1}
}{}
\newcommand{\wh}{\widehat}

\renewcommand{\hat}{\wh}

\makeatletter
\newcommand*{\RN}[1]{\expandafter\@slowromancap\romannumeral #1@}
\makeatother

\ifdefined\isarxiv 
\else 
\renewcommand{\cite}{\citep}
\fi
\usepackage{lineno}

\begin{document}

\ifdefined\isarxiv

\date{}

\title{Can You Count to Nine? A Human Evaluation Benchmark for Counting Limits in Modern Text-to-Video Models}
\author{
Xuyang Guo\thanks{\texttt{ gxy1907362699@gmail.com}. Guilin University of Electronic Technology.}
\and
Zekai Huang\thanks{\texttt{ zekai.huang.666@gmail.com}. The Ohio State University.}
\and
Jiayan Huo\thanks{\texttt{ jiayanh@arizona.edu}. University of Arizona.}
\and 
Yingyu Liang\thanks{\texttt{
yingyul@hku.hk}. The University of Hong Kong. \texttt{
yliang@cs.wisc.edu}. University of Wisconsin-Madison.} 
\and 
Zhenmei Shi\thanks{\texttt{
zhmeishi@cs.wisc.edu}. University of Wisconsin-Madison.}
\and
Zhao Song\thanks{\texttt{ magic.linuxkde@gmail.com}. The Simons Institute for the Theory of Computing at the UC, Berkeley.}
\and
Jiahao Zhang\thanks{\texttt{ ml.jiahaozhang02@gmail.com}. Independent Researcher.}
}

\else

\title{Can You Count to Nine? A Human Evaluation Benchmark for Counting Limits in Modern Text-to-Video Models}

\author{Antiquus S.~Hippocampus, Natalia Cerebro \& Amelie P. Amygdale \thanks{ Use footnote for providing further information
about author (webpage, alternative address)---\emph{not} for acknowledging
funding agencies.  Funding acknowledgements go at the end of the paper.} \\
Department of Computer Science\\
Cranberry-Lemon University\\
Pittsburgh, PA 15213, USA \\
\texttt{\{hippo,brain,jen\}@cs.cranberry-lemon.edu} \\
\And
Ji Q. Ren \& Yevgeny LeNet \\
Department of Computational Neuroscience \\
University of the Witwatersrand \\
Joburg, South Africa \\
\texttt{\{robot,net\}@wits.ac.za} \\
\AND
Coauthor \\
Affiliation \\
Address \\
\texttt{email}
}

\ifdefined\colmsubmissiontrue
\linenumbers
\fi

\maketitle

\fi

\ifdefined\isarxiv
\begin{titlepage}
  \maketitle
  \begin{abstract}
\input{00_abstract}

  \end{abstract}
  \thispagestyle{empty}
\end{titlepage}

{\hypersetup{linkcolor=black}
\tableofcontents
}
\newpage

\else

\begin{abstract}
\input{00_abstract}
\end{abstract}

\fi

\input{01_introduction} %%% Section 1. Introduction
\input{02_related_works}
\input{03_benchmark}

\input{04_experiments}

\input{05_conclusion}

%\newpage
%\input{deprecated_detailed_prompts}

\ifdefined\isarxiv
%\section*{Acknowledgments}
\bibliographystyle{alpha}
\bibliography{ref}

\else

\bibliography{ref}
\bibliographystyle{colm2025_conference}

\fi

\newpage
\onecolumn
\appendix
\begin{center}
    \textbf{\LARGE Appendix}
\end{center}

\paragraph{Roadmap.}
\input{07_append_impl_details}
\input{08_append_additional_experiments}
\input{09_append_additional_qual_study}

\input{10_append_video_examples}

%%%% Cut-line between first 10 pages and appendix

%%% some writing rules

%% Writing rule for creating tags.
%% Tags :
%% Theorem    \ref{thm:bla_bla}
%% Lemma      \ref{lem:bla_bla}
%% Claim      \ref{cla:bla_bla}
%% Corollary  \ref{cor:bla_bla}
%% Fact       \ref{fac:bla_bla}
%% Definition \ref{def:bla_bla}
%% Section    \ref{sec:bla_bla}
%% Subsection \ref{sub:bla_bla}
%% Equation   \ref{eq:bla_bla}

\end{document}

%% file: 00_abstract.tex
Generative models have driven significant progress in a variety of AI tasks, including text-to-video generation, where models like Video LDM and Stable Video Diffusion can produce realistic, movie-level videos from textual instructions. Despite these advances, current text-to-video models still face fundamental challenges in reliably following human commands, particularly in adhering to simple numerical constraints. In this work, we present \textbf{T2VCountBench}, a specialized benchmark aiming at evaluating the counting capability of SOTA text-to-video models as of 2025. Our benchmark employs rigorous human evaluations to measure the number of generated objects and covers a diverse range of generators, covering both open-source and commercial models. Extensive experiments reveal that all existing models struggle with basic numerical tasks, almost always failing to generate videos with an object count of 9 or fewer. Furthermore, our comprehensive ablation studies explore how factors like video style, temporal dynamics, and multilingual inputs may influence counting performance. We also explore prompt refinement techniques and demonstrate that decomposing the task into smaller subtasks does not easily alleviate these limitations. Our findings highlight important challenges in current text-to-video generation and provide insights for future research aimed at improving adherence to basic numerical constraints.

%% file: 01_introduction.tex
\section{Introduction}

Generative models have long been at the core of many of today's successes in the AI research community. By leveraging cross-modal, large-scale pretraining on language, visual, and speech data, these models have demonstrated significant progress across a wide scale of problems, including text-to-image generation, text-to-audio generation, natural language synthesis, and so on. In particular, text-to-video generation has emerged as one of the most impressive applications in recent years. Many stunning models, such as Sora~\cite{sora}, Kling~\cite{kling}, Pika~\cite{pika2.2}, and many so on, are capable of producing realistic, movie-level videos based on human instructions. This capability is powered by representative language-video models such as Video LDM~\cite{brl+23} and Stable Video Diffusion~\cite{bdk+23}.
Despite these rapid developments, text-to-video models still exhibit fundamental limitations in generating trustworthy videos that precisely follow human instructions. Challenges remain in producing coherent motions between frames~\cite{jxth24,llz+24}, adhering to real-world physical constraints~\cite{lhy+24,xyyg24}, and reliably refusing to generate offensive content~\cite{mzy+24,dcw+24}. Previous work has provided deep insights into these high-level issues and suggested promising directions for closing the research gap in text-to-video generation. However, most of these studies focus on the overall quality and coherence of the generated videos while overlooking some of the simpler, basic aspects.

In this work, we draw inspiration from prior observations that CLIP-based~\cite{rkh+21,pet+23} and text-to-image models~\cite{hwrl24} face difficulties with minimalist counting problems, and we extend this perspective to examine text-to-video models. Our focus is on assessing whether these models can adhere to basic numerical constraints as specified in user prompts. To achieve this, we present a specialized benchmark, \textbf{T2VCountBench}, aiming at evaluating the counting ability of SOTA text-to-video models as of 2025. Our benchmark employs rigorous human evaluations to count accurately the objects generated, and it covers a diverse range of generators, including both open-sourced and proprietary systems. Unlike earlier studies that emphasize high-level attributes such as video coherence and fidelity, our approach explicitly isolates counting performance from other generative capabilities. To the best of our knowledge, this represents one of the earliest efforts to systematically benchmark counting ability in modern text-to-video models.

With the proposed T2VCountBench, we conduct an extensive evaluation to probe the counting capability of text-to-video generative models. The experimental results indicate that all existing models exhibit clear failures when it comes to simple numerical constraints, almost always failing to generate videos with an object count of 9 or fewer. We also perform a comprehensive ablation study on various factors that may influence the counting ability of these models, such as video style, temporal dynamics, and multilingual inputs. To further illustrate the non-trivial nature of such a negative result, we examine simple prompt refinement techniques and show that the counting limitations cannot be easily alleviated by decomposing the task into smaller subtasks. The contributions of this paper are summarized as follows:
\begin{itemize}
    \item We introduce T2VCountBench, the first specialized benchmark that systematically evaluates counting ability in modern text-to-video generation models.
    \item We demonstrate through extensive human evaluations that all SOTA text-to-video models consistently face difficulties at basic numerical constraints.
    \item We conduct comprehensive ablation studies examining how factors such as video style, temporal dynamics, and multilingual inputs affect counting performance.
    \item We explore prompt engineering techniques showing that counting limitations are inherent to the models and cannot be easily overcome through task decomposition.
\end{itemize}

\paragraph{Roadmap.} In Section~\ref{sec:rel_work}, we discuss related works of this paper. In Section~\ref{sec:bench}, we present the basic information of our proposed T2VCountBench benchmark. In Section~\ref{sec:experiments}, we show the experiment results and our key observation. We conclude our paper in Section~\ref{sec:conclusion}. 

%% file: 02_related_works.tex
\section{Related Works}\label{sec:rel_work}

\paragraph{Text-to-Video Benchmarks.} 
As one of the most impactful applications of generative AI, text-to-video generative models have deeply revolutionized the process of visual art creation and have shown astonishing potential to create film-level video samples. To evaluate the effectiveness of such models, a variety of benchmarking papers have been involved~\cite{llr+23,bmv+23}, which systematically probe the capabilities of text-to-video generative models. 
These benchmarks have been becoming increasingly critical after the proposal and wide use of the text-to-video diffusion models~\cite{hsg+22,wgw+23,ytz+24}, which allows us to generate high fidelity and human instruction-aligned videos and makes these benchmarks make more sense. One of the earliest and most representative evaluation benchmarks is FETV~\cite{llr+23}, which considers fine-grained evaluation spectrum of different types of text prompts and also discusses effective automatic metrics for video generation, such as FID~\cite{hru+17}, FVD~\cite{usk+18} for video quality, and CLIPScore~\cite{hhf+21} to ensure video-text alignment. %\Jiahao{Zekai, plz refer to p2 of~\cite{llr+23} here, cite fid, fvd, clipscore.}.\Zekai{Done} 
Next, StoryBench~\cite{bmv+23} extends the dimension of text-to-video benchmarks and considers a novel and different perspective, which focuses on video-based storytelling capabilities, taking both action generation, and story continuation into consideration. 

\paragraph{Diffusion Models for Text-to-Video Generation.}
Video diffusion models~\cite{hsg+22,bdk+23} have been widely used in business nowadays, such as Sora~\cite{sora}, Kling~\cite{kling}, Pika~\cite{pika2.2}, and many so on.
Most current state-of-the-art image or video generation models are based on the diffusion model~\cite{hja20,hs21}. 
To achieve text-to-image or video generation \cite{hsc+22,wyc+23}, the generation models need to equip multi-modality understanding ability. The most popular technique is learning representation alignment for different modalities based on CLIP~\cite{rkh+21,rdn+22,bjg+23}, where Stable Diffusion \cite{rbl+22} achieves this by such a technique. 
On the other hand, thanks to the scalability of Transformers~\cite{vsp+17}, the transformer-based video generation models can easily extend their size and ability~\cite{yzas21,dzht22,px23} and make themselves successful.  
Furthermore, one line of work is to improve the video generation quality~\cite{sph+22,wcm+24,wyt+25}, while another line of work is to accelerate the generation speed \cite{zwy+22,sh22,hyz+22,wgw+23,kpct23}. 
Other text-to-video models are based on GAN-based methods~\cite{dfhp19} or flow matching-based methods~\cite{jsl+24}. Our insights from this work could enlighten future advancements in text-to-video and text-to-image generative models, with a particular focus on enhancing controlled generation~\cite{wxz+24,wsd+24,ccb+25,cgh+25,czz+25} and expressive power~\cite{cgl+25,gkl+25,cgl+25_rich,csy25,gll+25}.

%% file: 03_benchmark.tex
\section{The T2VCountBench Benchmark}\label{sec:bench}

In this section, we commence by introducing some baseline generators evaluated in the benchmark in Section~\ref{sec:base_models}, and then present the text prompt template used for generating videos in Section~\ref{sec:prompts}. Next, we discuss our evaluation protocol in Section~\ref{sec:eval_prot}.

\subsection{Baseline Models}\label{sec:base_models}

\begin{table}[!ht]
    \centering
    
    \resizebox{0.95\linewidth}{!}{ 
    \begin{tabular}{|c|c|c|c|c|}
        \hline
        \textbf{Model Name} & \textbf{Organization} & \textbf{Year} & \textbf{\# Params} & \textbf{Open}\\
        \hline
       Kling~\cite{kling} & Kuai & 2024 & N/A & No \\
        \hline
       Wan2.1~\cite{wan2.1} & Alibaba &2025 & 14B & Yes \\
        \hline
       Sora~\cite{sora} & OpenAI & 2024 & N/A & No \\
        \hline
       Mochi-1~\cite{mochi1} & Genmo & 2024 & 10B & Yes \\
        \hline
       LTX Video~\cite{hcb+24} & Lightricks & 2024 & 2B & Yes  \\
        \hline
       Pika 2.2~\cite{pika2.2} & Pika Labs & 2025 & N/A & No \\
        \hline
       Dreamina~\cite{dreamina} & ByteDance & 2024 & N/A & No  \\
        \hline
       Qingying~\cite{qingying} & Zhipu & 2024 & 5B & Yes  \\
        \hline
       Gen 3 Alpha~\cite{g24} & RunwayML & 2024 & N/A & No  \\
        \hline
       Hailuo~\cite{hailuo} & MiniMax & 2025 & N/A & No  \\
        \hline

    \end{tabular}
    }
    \caption{\textbf{Key Details of the 10 Assessed Text-to-Video Diffusion Models.}}
    \label{tab:models}
\end{table}

Our benchmark covers a wide range of text-to-video generative models, focusing on modern systems released between 2024 and 2025. We evaluate the counting ability of 10 models, including both open-source and proprietary commercial generators with API access. This selection guarantees the trustworthiness and timeliness of our benchmark. Basic information of the tested models are provided in Table~\ref{tab:models}.

For generation settings, we use the smallest available resolution (typically 720p) to reduce the workload on fidelity control and to focus on counting accuracy. We adopt a 16:9 aspect ratio and select the shortest available video duration (typically 4 seconds) to further concentrate on counting. For additional implementation details, please refer to Appendix~\ref{sec:append_impl_detail}.

\subsection{Generation Prompts}\label{sec:prompts}

The selection of text prompts is critical for a trustworthy assessment of the inherent counting capabilities of text-to-video models. While many current benchmarks~\cite{llr+23,hhy+24,yhx+24} do not explicitly evaluate counting abilities, or mix counting with other less relevant tasks~\cite{lcl+24,shl+24}, our approach isolates counting for a direct and effective assessment.

We use the following template of text prompts in most evaluations:
\begin{ptemplate}[\textless scene transition\textgreater, \textless number\textgreater~\textless object\textgreater~doing something with \textless motion constraint\textgreater~in \textless style\textgreater.]\label{pro:most_general}\end{ptemplate}

Here, \textless number\textgreater~represents the number of objects to be generated in the video, taking values in $\{1,3,5,7,9\}$ to cover a range of difficulty levels. The \textless object\textgreater~can be one of three categories: \texttt{'human'}, \texttt{'nature'}, or \texttt{'artifact'}. To assess the stability of counting under temporal dynamics, we include both a \textless scene transition\textgreater~and a \textless motion constraint\textgreater~in the prompt. We also examine the impact of style by adding the \textless style\textgreater~constraint. For scene, motion, and style, three options are provided for each entry, resulting in a total of 165 unique text prompts. Two examples of these prompts are shown below:

\begin{pexample}[Five students walking alongside the road, in cartoon style.]\label{pro:general_example_1}\end{pexample}
\begin{pexample}[Seven kites soar through the sky, in graceful circles.]\label{pro:general_example_2}\end{pexample}

\subsection{Evaluation Protocol}\label{sec:eval_prot}

In this paper, we adopt a fully human evaluation protocol to ensure a rigorous and error-free evaluation. Five AI-knowledgeable undergraduate and graduate students evaluate all the generated videos visually. We evaluate two metrics: \textbf{Counting Accuracy} and \textbf{Object Fidelity}. Counting Accuracy measures whether the exact number of objects is generated, while Object Fidelity assesses whether the generated objects are genuine and recognizable.

Let $N$ denote the number of required objects specified in the prompt, and let $\hat{N}$ denote the number of generated target objects in the video. We denote the set of all text prompts as $P$, and a single prompt as $p \in P$. Our metrics are computed as follows:

\paragraph{Counting Accuracy.}  
Let $P_0 \subseteq P$ be the subset of prompts evaluated in a specific experiment (e.g., all prompts with \texttt{'human'} objects or those with an object count of 5). Counting Accuracy is defined as:
\begin{align*}
%$
    \mathsf{CountAcc}(P_0) := \frac{1}{|P_0|}\sum_{p \in P_0} \mathbf{1}[\hat{N}_p = N_p],
%    $
\end{align*}
where $\hat{N}_p$ is the number of objects generated for prompt $p$, and $N_p$ is the ground truth number of objects. Here ${\bf 1}[E]$ denote the variable that outputs $1$ if event $E$ is true, and $0$ otherwise. We do not require every generated object to be perfectly faithful. If a human annotator considers an object generally similar to the target, it is counted. In cases of ambiguity, such as when different annotators report different values for $\hat{N}$ for the same prompt, the result is deemed correct if at least one annotator reports the correct count. This design separates the evaluation of object fidelity from counting.

\paragraph{Object Fidelity.}  
Object Fidelity measures the proportion of generated objects that are both genuine and recognizable. It is computed as:
\begin{align*}
%$
    \mathsf{AvgFidelity}(P_0) := \frac{1}{|P_0|}\sum_{p \in P_0}  {\hat{M}_p}/{\hat{N}_p},
%$
\end{align*}
where $\hat{M}_p$ is the number of trustworthy objects among the $\hat{N}_p$ generated for prompt $p$. In cases where annotators provide different values for $\hat{M}_p$, we use the highest reported value.

%% file: 04_experiments.tex
\vspace{-2mm}
\section{Experiments}\label{sec:experiments}
\vspace{-2mm}
We discuss the experimental observations obtained with our T2VCountBench in this section. Section~\ref{sec:exp_main} evaluates the overall counting performance for a wide range of baseline generators, while Section~\ref{sec:exp_abl} examines the impact of multiple factors on the counting capability of text-to-video generative models, followed by Section~\ref{sec:exp_mult_ling} explore multilingual abilities of text-to-video diffusion models. Finally, Section~\ref{sec:exp_refine} analyzes the impact of prompt refinement. 

\vspace{-3mm}
\subsection{Overall Counting Results}\label{sec:exp_main}

\begin{table}[!ht]
\centering
\resizebox{1\linewidth}{!}{
\begin{tabular}{|c|cccccccc|}
\toprule
\multirow{2}{*}{\centering\textbf{Model}} &  
\multicolumn{2}{c}{\textbf{Human}} &      
\multicolumn{2}{c}{\textbf{Nature}} & 
\multicolumn{2}{c}{\textbf{Artifact}} & 
\multicolumn{2}{c|}{\textbf{Overall}} \\ 
\cmidrule(lr){2-3} \cmidrule(lr){4-5} \cmidrule(lr){6-7} \cmidrule(lr){8-9}
& \textbf{Count Acc} & \textbf{Fidelity Avg} & \textbf{Count Acc} & \textbf{Fidelity Avg} & \textbf{Count Acc} & \textbf{Fidelity Avg} & \textbf{Count Acc} & \textbf{Fidelity Avg} \\
\midrule
Mochi-1      & 0.20 & {\color{red} 0.98} & 0.25 & {\color{red} 0.89} & 0.31 & {\color{red} 0.93} & 0.25 & {\color{red} 0.93} \\
Gen 3 Alpha  & 0.29 & 0.61 & 0.25 & 0.63 & 0.33 & 0.74 & 0.29 & 0.66 \\
Dreamina     & 0.27 & 0.72 & 0.33 & 0.75 & 0.33 & 0.75 & 0.31 & 0.74 \\
Kling        & 0.42 & 0.93 & 0.27 & {\color{red} 0.92} & 0.38 & 0.83 & 0.36 & 0.89 \\
Sora         & 0.36 & {\color{red} 0.96} & {\color{blue} 0.45} & 0.86 & 0.33 & 0.85 & 0.38 & {\color{red} 0.89} \\
Hailuo       & 0.42 & 0.89 & {\color{blue} 0.42} & 0.85 & 0.38 & 0.63 & 0.41 & 0.79 \\
Qingying     & {\color{blue} 0.53} & 0.94 & 0.33 & 0.80 & {\color{blue} 0.40} & 0.80 & 0.42 & 0.85 \\
Wan2.1       & 0.51 & {\color{red} 1.00} & 0.36 & {\color{red} 0.93} & {\color{blue} 0.40} & {\color{red} 0.92} & {\color{blue} 0.42} & {\color{red} 0.95} \\
LTX Video    & {\color{blue}0.60} & 0.71 & 0.29 & 0.74 & {\color{blue}0.40} & {\color{red} 0.91} & {\color{blue} 0.43} & 0.79 \\
Pika 2.2     & {\color{blue} 0.67} & 0.68 & {\color{blue} 0.45} & 0.82 & 0.36 & 0.74 & {\color{blue} 0.50} & 0.74 \\
\bottomrule
\end{tabular}
}
\caption{{\bf Overall Counting Accuracy and Fidelity Across various Object Types.} We highlight the three generators with the best counting accuracy in {\color{blue} blue}, and the three models with the best average fidelity in {\color{red} red}. 
}
\label{tab:count_main_acc_fid}
\vskip -0.2in
\end{table}

We use the universal prompt template shown in Prompt Template~\ref{pro:most_general} to examine the inherent counting capability of text-to-video models. We instantiate this template with the following options:
\begin{itemize}
    \item \textless number\textgreater: $\{1,3,5,7,9\}$;
    \item \textless object\textgreater: \texttt{'Human'}, \texttt{'Nature'}, \texttt{'Artifact'};
    \item \textless scene transition\textgreater: \texttt{'None'}, \texttt{'Home to City'}, \texttt{'Home to Nature'};
    \item \textless motion\textgreater: \texttt{'None'}, \texttt{'Turn'}, \texttt{'Rotation'};
    \item \textless style\textgreater: \texttt{'Plain'}, \texttt{'Cartoon'}, \texttt{'Watercolor'}.
\end{itemize}

For each model, we consider all compositions of \textless number\textgreater~and \textless object\textgreater, while ablating one property among scene, motion, and style at a time and keeping the others fixed. All results are obtained through the human evaluation process described in Section~\ref{sec:eval_prot}. Detailed experimental results are presented in Table~\ref{tab:count_main_acc_fid}, with models sorted in ascending order by overall counting accuracy.

Our experiments reveal several key findings. First, most models struggle with counting objects, even when the number of objects is low (from 1 to 9). For example, even the strongest model, Pika 2.2, achieves only 50\% overall counting accuracy, highlighting a fundamental limitation in current text-to-video models. We also observe significant variations in counting accuracy across different object classes. For instance, while Pika 2.2 and LTX Video achieve over 60\% accuracy in the Human category, their accuracy drops below 40\% in the Artifact category. We summarize this observation as follows:

\begin{observation}
    The overall counting performance of SOTA text-to-video generators is unsatisfactory, and counting accuracy varies significantly across object categories.
\end{observation}

In addition to counting accuracy, we evaluate object fidelity. Our results show that both per-class and overall fidelity scores are generally higher than the counting accuracy. For example, Wan2.1 achieved a perfect 100\% fidelity in the Human category across all 165 prompts, and Mochi-1 reached a fidelity score of 0.98 in the same category. Interestingly, the rankings for counting accuracy and object fidelity are not strongly correlated. Pika 2.2, which has the best counting performance, does not rank among the top three in average fidelity. Conversely, Mochi-1 ranks in the top three for fidelity across all classes but has the lowest overall counting accuracy. Only Wan2.1 shows a balanced performance by ranking in the top three for both counting and fidelity. This leads to the following observation:

\begin{observation}\label{obs:fid_acc_irrelevance}
    Although most text-to-video models achieve high object fidelity, this does not directly translate into accurate counting performance.
\end{observation}

These findings motivate us to treat counting as a distinct challenge from fidelity, as improvements in fidelity alone do not necessarily lead to better counting capabilities.

\subsection{Ablation Study}\label{sec:exp_abl}

In this subsection, we study the impact of different relevant factors on model's counting accuracy, as well as the fidelity of the generated videos. Specifically, we consider the difficulty levels of counting tasks, the impact of video art style, and the impact of videos' temporal dynamics, such as scene transition or object motion. Due to space limitations, ablation study on style and object motion are delayed to Appendix~\ref{sec:append_more_experiments}.

\begin{table}[!ht]
\centering
\resizebox{\linewidth}{!}{
\begin{tabular}{|c|cccccccccc|}
\toprule
\multirow{2}{*}{\textbf{Model}} & 
\multicolumn{2}{c}{\textbf{1}} & 
\multicolumn{2}{c}{\textbf{3}} & 
\multicolumn{2}{c}{\textbf{5}} & 
\multicolumn{2}{c}{\textbf{7}} & 
\multicolumn{2}{c|}{\textbf{9}} \\
\cmidrule(lr){2-3} \cmidrule(lr){4-5} \cmidrule(lr){6-7} \cmidrule(lr){8-9} \cmidrule(lr){10-11}
& \textbf{Count Acc} & \textbf{Fidelity Avg} & \textbf{Count Acc} & \textbf{Fidelity Avg} & \textbf{Count Acc} & \textbf{Fidelity Avg} & \textbf{Count Acc} & \textbf{Fidelity Avg} & \textbf{Count Acc} & \textbf{Fidelity Avg} \\
\midrule
Mochi-1     & 0.97 & 1.00 & 0.27 & 0.96 & 0.03 & 0.88 & 0.00 & 0.89 & 0.00 & 0.93 \\
Gen 3 Alpha & 0.91 & 0.67 & 0.39 & 0.64 & 0.12 & 0.65 & 0.03 & 0.65 & 0.00 & 0.69 \\
Dreamina    & 0.85 & 0.82 & 0.21 & 0.73 & 0.21 & 0.73 & 0.21 & 0.72 & 0.06 & 0.68 \\
Kling       & 0.91 & 0.95 & 0.39 & 0.88 & 0.24 & 0.87 & 0.18 & 0.89 & 0.06 & 0.87 \\
Sora        & 0.94 & 0.92 & 0.48 & 0.95 & 0.18 & 0.84 & 0.18 & 0.85 & 0.12 & 0.89 \\
Hailuo      & 0.91 & 0.93 & 0.48 & 0.72 & 0.15 & 0.80 & 0.30 & 0.80 & 0.18 & 0.71 \\
Qingying    & 0.91 & 0.91 & 0.58 & 0.86 & 0.39 & 0.83 & 0.12 & 0.87 & 0.09 & 0.77 \\
Wan2.1      & 0.91 & 0.97 & 0.61 & 0.97 & 0.42 & 0.93 & 0.18 & 0.91 & 0.00 & 0.96 \\
LTX Video   & 0.85 & 0.88 & 0.55 & 0.84 & 0.39 & 0.80 & 0.27 & 0.76 & 0.09 & 0.65 \\
Pika 2.2    & 0.91 & 0.86 & 0.61 & 0.82 & 0.58 & 0.76 & 0.27 & 0.65 & 0.12 & 0.63 \\
\bottomrule
\end{tabular}
}
\caption{\textbf{Counting Accuracy and Object Fidelity Across Different Difficulty Levels.}}
\label{tab:count_difficult_acc_fid}
\vskip -0.1in
\end{table}

\paragraph{Impact of Difficulty Levels.}  
In this experiment, we analyze how the models' counting capabilities change as the task becomes more challenging, ranging from counting 1 object to counting 9 objects. Using the same prompt settings as in Section~\ref{sec:exp_main}, Table~\ref{tab:count_difficult_acc_fid} shows the aggregated metrics for different difficulty levels.

The results clearly indicate that counting accuracy drops significantly as the required number of objects increases. For instance, most models achieve around 90\% accuracy when generating a single object, but accuracy falls to less than 10\% when generating nine objects, with some models like Mochi-1 failing on all prompts. Even for a moderate task, such as generating five objects, more than half of the models only succeed in about 20\% of the prompts, demonstrating a substantial challenge in counting. In contrast, object fidelity remains relatively stable regardless of the number of objects, indicating that current text-to-video models are robust in fidelity even when counting becomes difficult. Our key observation is summarized as follows:

\begin{observation}
    Models' counting accuracy drops rapidly as the number of objects increases, often reaching unsatisfactory levels even at moderate counts (e.g., only 5 objects). In contrast, object fidelity remains largely unaffected by the number of objects.
\end{observation}
%\vspace{-1mm}
\paragraph{Impact of Scene Transition.}  
In this study, we use the general Prompt Template~\ref{pro:most_general} and vary only the scene transition of the generated videos. Specifically, we use the following options:

\begin{itemize}
    \item \textless number\textgreater: $\{1,3,5,7,9\}$;
    \item \textless object\textgreater: \texttt{'Human'}, \texttt{'Nature'}, \texttt{'Artifact'};
    \item \textless scene transition\textgreater: \texttt{'None'}, \texttt{'Home to City'}, \texttt{'Home to Nature'}.
\end{itemize}

To isolate the impact of scene transition, we fix the other options by setting \textless style\textgreater~to \texttt{'Plain'} and \textless motion\textgreater~to \texttt{'None'}. The counting accuracy results for this ablation study are shown in Figure~\ref{fig:scene_acc}, and the object fidelity results are discussed in Appendix~\ref{sec:append_more_experiments} (Figure~\ref{fig:scene_fid}).

Our results show that, with a few exceptions such as Kling and Dreamina, most models do not exhibit a noticeable change in counting accuracy when the scene transition varies. For example, Hailuo achieves a counting accuracy of 0.47 for both the \texttt{'Plain'} and \texttt{'Home to City'} settings, while \texttt{'Home to Nature'} only slightly improves the accuracy to 0.53. Based on these findings, we conclude:

\begin{observation}
    Scene transition in videos does not significantly affect models' counting capability, suggesting that models' temporal dynamics may have little impact on counting performance.
\end{observation}

\begin{figure}[!ht]
    \centering
    \includegraphics[width=1.0\linewidth]{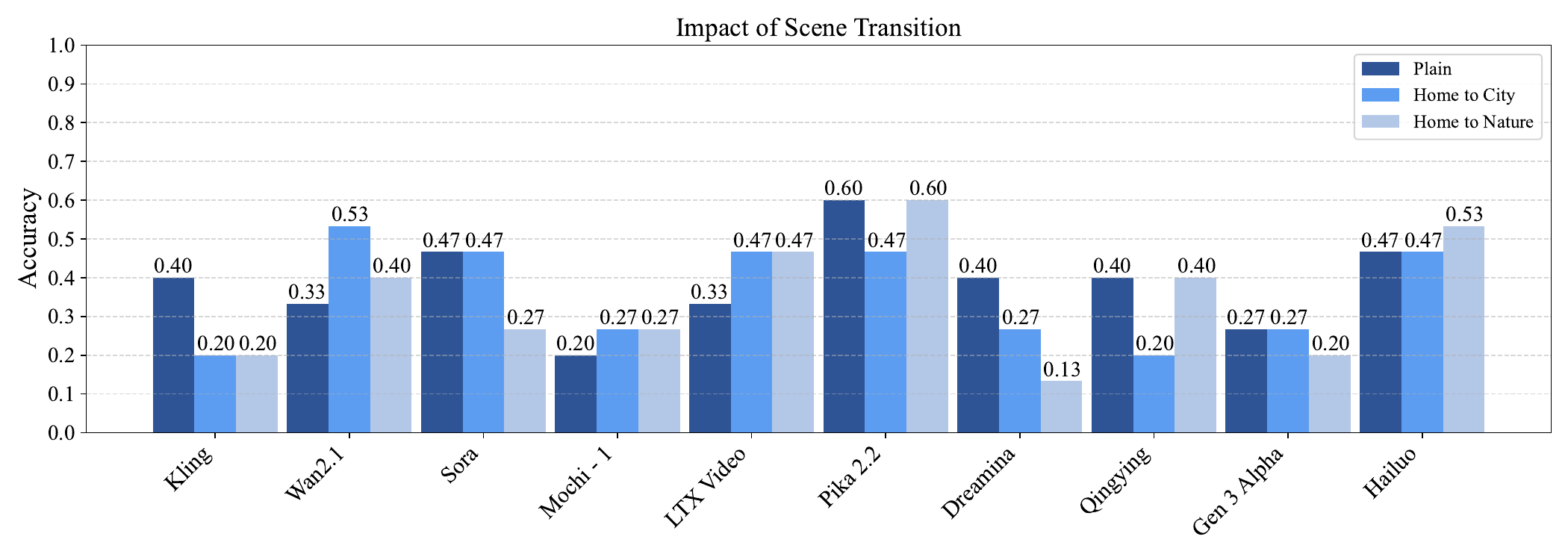}
    \vskip -0.18in
    \caption{
    {\bf Impact of Scene Transition on Counting Accuracy}.}
    \label{fig:scene_acc}
    \vskip -0.15in
\end{figure}

\begin{figure}[!ht]
    \centering
    \includegraphics[width=1.0\linewidth]{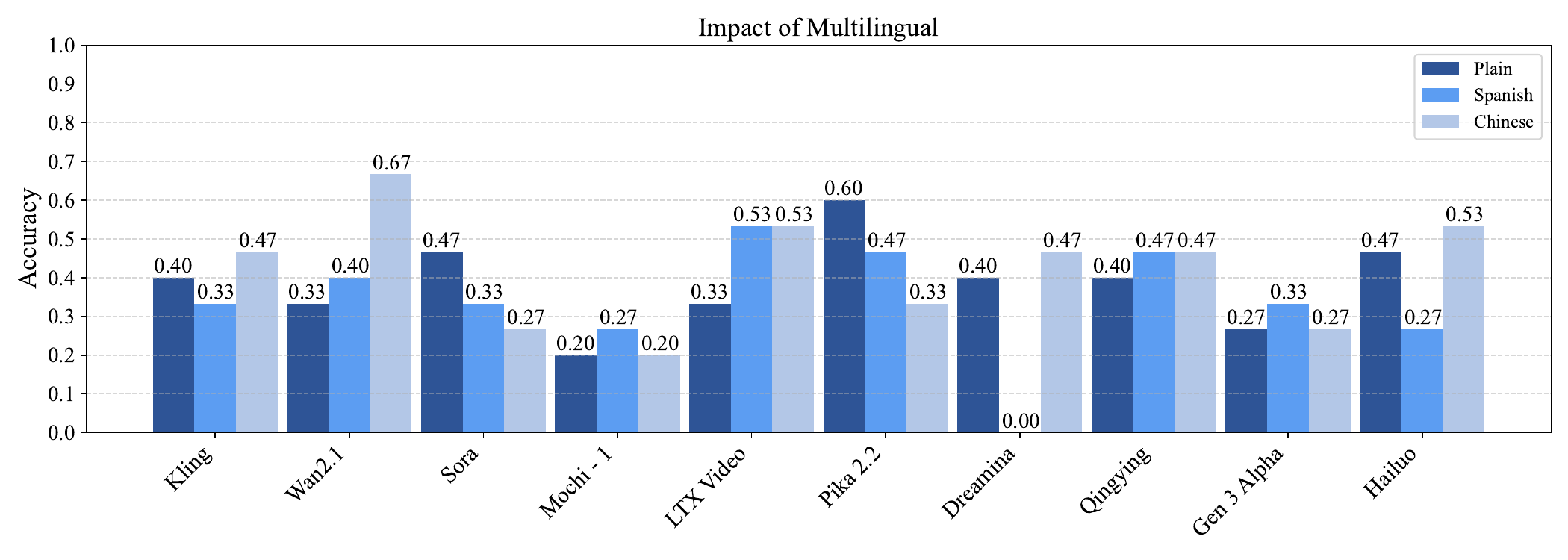}
    \vskip -0.18in
    \caption{
    {\bf Impact of Multilingual Prompts on Counting Accuracy}.}
    \label{fig:multilingual_acc}
\end{figure}

\subsection{Multilingual Abilities}\label{sec:exp_mult_ling}

In this subsection, we investigate the multilingual counting ability of text-to-video generators by testing how well they follow quantity constraints when the prompt language changes. This is important because many users express ideas more naturally in their native language rather than in English. To isolate the language factor, we use a minimalist prompt template:

\begin{ptemplate}[(Translate to \textless language\textgreater) \textless number\textgreater~\textless object\textgreater~doing something.]\label{pro:lang_temp}\end{ptemplate}

The prompt is originally written in English and then translated into the target language using the standard Google Translate API. The options for the prompt are:

\begin{itemize}
    \item \textless number\textgreater: $\{1,3,5,7,9\}$; \quad
    \item \textless object\textgreater: \texttt{'Human'}, \texttt{'Nature'}, \texttt{'Artifact'}; \quad
    \item \textless language\textgreater: \texttt{'English'}, \texttt{'Spanish'}, \texttt{'Chinese'}.
\end{itemize}

Our experimental results, shown in Figure~\ref{fig:multilingual_acc} and Figure~\ref{fig:multilingual_fid}, reveal that models exhibit significant variance when counting objects across different languages. For example, models such as Sora and Pika perform best with English prompts. In contrast, models developed by Chinese teams, including Kling, Wan2.1, and Hailuo, achieve the best results in Chinese. For Spanish, most models perform worse compared to the other two languages, with Dreamina even notably refusing to respond to Spanish prompts. Notably, Qingying delivers consistent performance across languages while maintaining good overall results. These findings underscore the need to enhance the multilingual capabilities of text-to-video models to promote digital fairness.

\begin{observation}
    Counting accuracy varies significantly across languages. Some models excel in English, others in Chinese, and several struggle with Spanish, highlighting fairness concerns in current multilingual capabilities.
\end{observation}

\begin{figure}[!ht]
    \centering
    \includegraphics[width=1.0\linewidth]{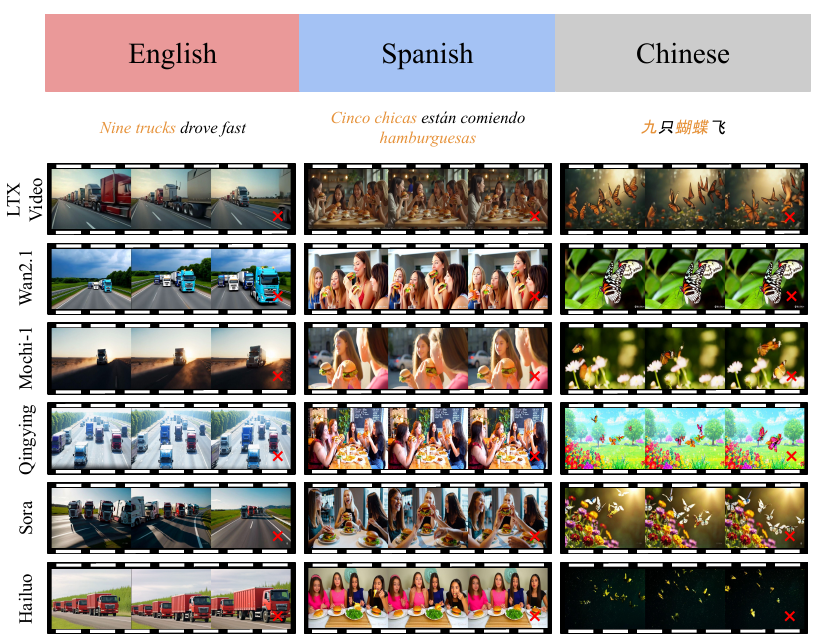}
    \caption{
    {\bf Qualitative Study on Multilingual Prompts.} The Spanish prompt describes ``five girls eating burgers'', while the Chinese prompt describes ``nine butterflies flying''.}
    \label{fig:7_4}
\end{figure}

To demonstrate the impact of multilingual counting ability more thoroughly, we also present a qualitative study in Figure~\ref{fig:7_4}. From the qualitative study, we observe that all models exhibit a failure in generating nine objects in both English and Chinese results. While the fidelity is desirable, the count is entirely incorrect, matching our findings in Table~\ref{tab:count_difficult_acc_fid}, which highlight the models' inability to generate a large numer of objects. For Spanish, we selected a prompt designed to generate five objects, but all models surprisingly failed in this task. This strengthens the inherent limitations of text-to-video generators in counting when faced with multilingual challenges.

\subsection{Prompt Refinement}\label{sec:exp_refine}

\begin{figure}[!ht]
    \centering
    \includegraphics[width=1.0\linewidth]{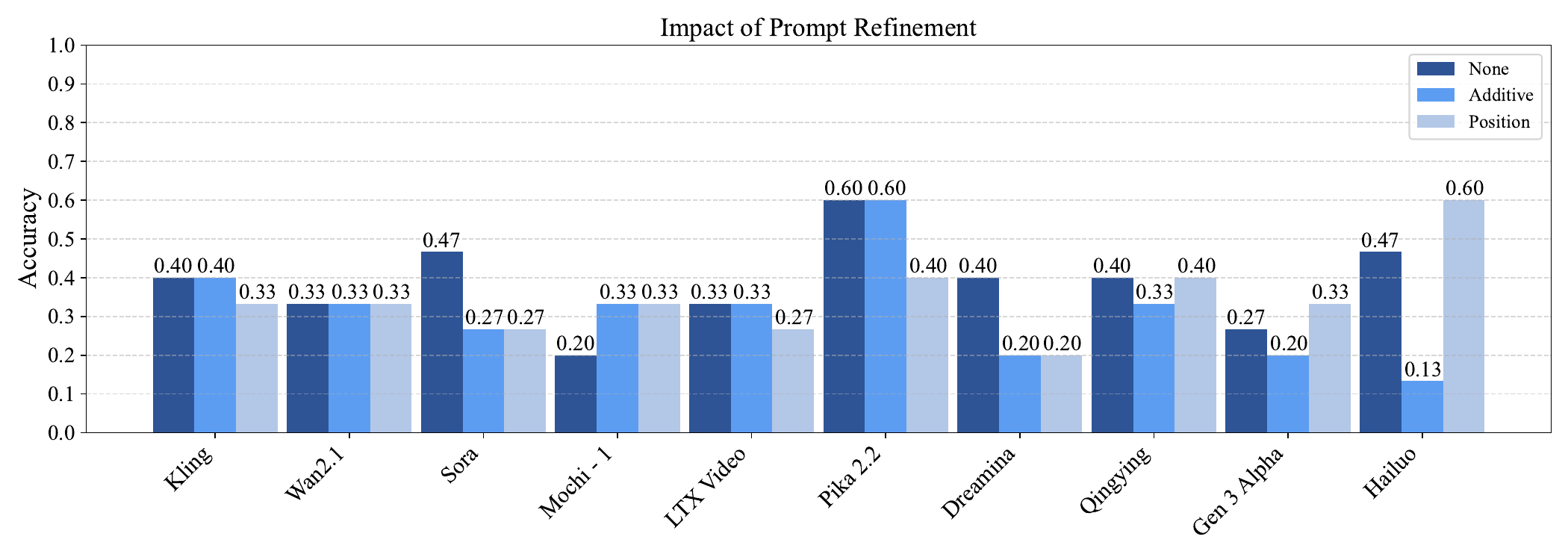}
    \vskip -0.2in
    \caption{
    {\bf Impact of Prompt Refinement on Counting Accuracy}.}
    \label{fig:prompt_refinement_acc}
    \vskip -0.2in
\end{figure}

\begin{figure}[!ht]
    \centering
    \includegraphics[width=1.0\linewidth]{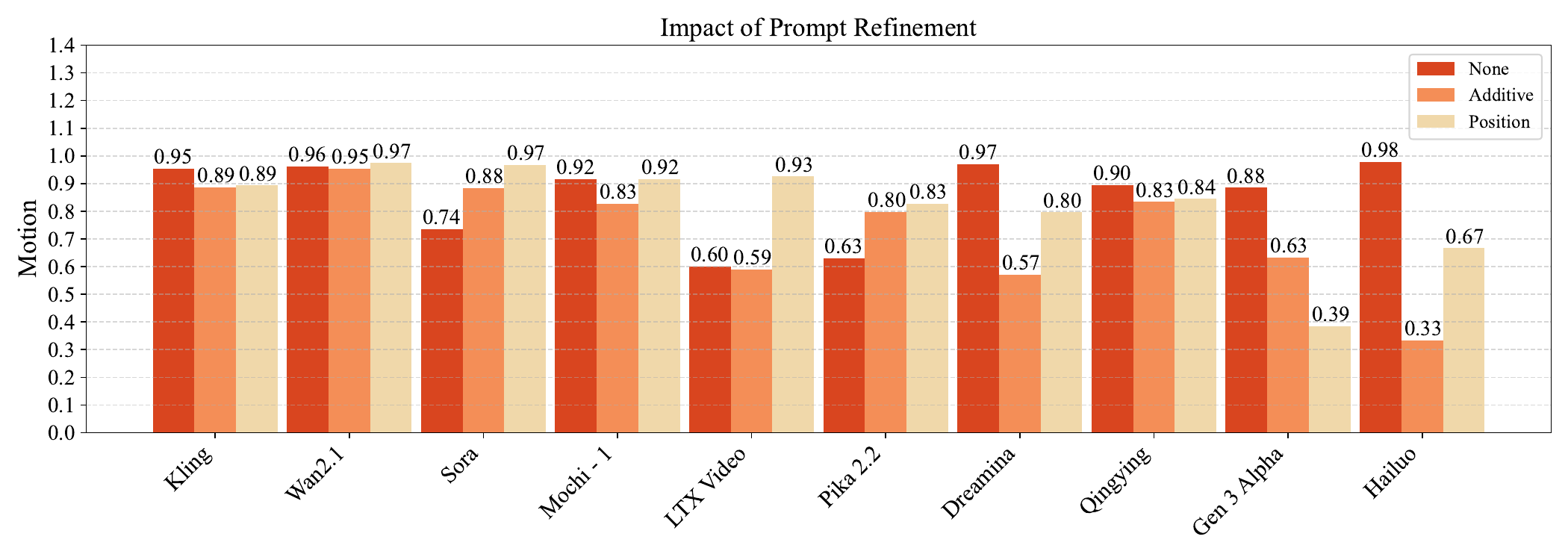}
    \vskip -0.2in
    \caption{
    {\bf Impact of Prompt Refinement on Object Fidelity}. }
    \label{fig:prompt_refinement_fid}
    \vskip -0.2in
\end{figure}

In this experiment, we examine whether simple prompt refinement can alleviate the intrinsic counting drawbacks of text-to-video models. Inspired by how humans break down tasks when counting a large number of objects, we consider two types of prompt refinements: additive decomposition and position guidance.

\paragraph{Prompt Design.}  
Following a minimalist design similar to Section~\ref{sec:exp_mult_ling}, we focus solely on prompt refinement. Our intuition is -- when counting a large group, a person might naturally break it into two smaller groups. Let $N$ be the desired number of objects, and let $\lfloor \cdot \rfloor$ denote the floor operation. Our first refined prompt template, which uses additive decomposition, is defined as:
\begin{ptemplate}[A group of $\lfloor N/2 \rfloor$ \textless object\textgreater~doing something, with another group of $\lfloor N - N/2 \rfloor$ doing something.]\label{pro:refine_add}\end{ptemplate}
\begin{pexample}[A group of three fishermen fishing, with another group of four fishermen fishing.]\label{pro:refine_add_example1}\end{pexample}

We also introduce a position guidance prompt that explicitly indicates where to place the two groups, reducing ambiguity:
\begin{ptemplate}[A group of $\lfloor N/2 \rfloor$ \textless object\textgreater~doing something on the left side, with another group of $N-\lfloor N/2 \rfloor$ doing something on the right side.]\label{pro:refine_pos}\end{ptemplate}
\begin{pexample}[A group of three bicycles leaning against a wall on the left side, while another group of four bicycles leaning against a wall on the right side.]\label{pro:refine_pos_example1}\end{pexample}

These refined prompts are applied only for cases where $N \geq 2$. For $N=1$, we use the standard prompt as described in Section~\ref{sec:exp_main}.

\paragraph{Experimental Settings.}  
In this experiment, we simplify the prompt options compared to Section~\ref{sec:exp_main}. Specifically, we use:

\begin{itemize}
    \item \textless number\textgreater: $\{1,3,5,7,9\}$; \quad
    \item \textless object\textgreater: \texttt{'Human'}, \texttt{'Nature'}, \texttt{'Artifact'}; \quad
    \item \textless refinement\textgreater: \texttt{'None'}, \texttt{'Additive'}, \texttt{'Position'}.
\end{itemize}

For all other options, default values are used. When the refinement is set to \texttt{'Additive'}, we use Prompt Template~\ref{pro:refine_add}, and when it is set to \texttt{'Position'}, we use Prompt Template~\ref{pro:refine_pos}. Our results are presented in Figure~\ref{fig:prompt_refinement_acc} and Figure~\ref{fig:prompt_refinement_fid}.

\paragraph{Findings.}  
Our results highlight that, in most cases, prompt refinement does not improve counting accuracy, while in some cases, it even degrades performance. For example, WanX2.1 shows similar counting accuracy across all prompt settings, while Hailuo's accuracy drops from 0.47 with no refinement to 0.27 with both additive and position guidance. Only in rare instances, such as Hailuo with position guidance, does counting accuracy improve (from 0.47 to 0.60). Regarding object fidelity, the impact of prompt refinement is also marginal, with improvements observed only in isolated cases (e.g., LTX Video with position guidance). We summarize our observation as follows:

\begin{observation}
    Simple prompt refinement does not consistently improve counting accuracy or object fidelity, and in some cases, it may even reduce performance.
\end{observation}

These findings highlight the inherent challenge of the counting limitation in text-to-video models. Simple, straightforward prompt refinements are insufficient, leaving significant room for future work to address this problem.

%% file: 05_conclusion.tex
\section{Conclusion}\label{sec:conclusion}
Our T2VCountBench reveals fundamental limitations in text-to-video models' counting capabilities. Despite generating visually appealing videos, these models consistently fail to adhere to basic numerical constraints like counting 9 or fewer objects. Our ablation studies across video style, temporal dynamics, and multilingual inputs confirm this pervasive limitation, while prompt refinement attempts yielded minimal improvements. These findings highlight a critical gap in current text-to-video generation that requires additional attention from the research community as we work toward developing truly trustworthy generative video systems that accurately interpret human instructions.

%% file: 07_append_impl_details.tex
In Section~\ref{sec:append_impl_detail}, we provide implementation details for each baseline generator.
we present the observations of additional experiments in Section~\ref{sec:append_more_experiments}. In Section~\ref{sec:append_qual_study}, we illustrate the details of additional qualitative studies.
In Section~\ref{sec:append_video_example}, we present a wide range of generated video examples in our benchmark.

\section{Implementation Details}\label{sec:append_impl_detail}

We present the additional details of the baseline models in this subsection. Specifically, the details of all the 10 text-to-video models are listed as follows:

\begin{itemize}
    \item \textbf{Kling}~\cite{kling}:
    Kling is a private text-to-video model from Kuai, released in 2024. It comes in three versions: Kling 1.6, Kling 1.5, and Kling 1.0. It offers two generation modes—standard and high-quality (the latter is available to members). Kling supports creative parameters: higher settings produce more relevant results, while lower settings yield more creative outputs. It does not support camera movement. It can generate 5-second or 10-second videos, and it supports aspect ratios of 16:9, 1:1, and 9:16. Kling also supports negative prompts, AI-generated prompt hints (powered by DeepSeek), and a prompt dictionary. It can create four videos from the same prompt at once and allows you to set a seed. Each video takes around four minutes to process, and you can batch up to five videos at a time.
    \item \textbf{Wan2.1}~\cite{wan2.1}: 
    Wan2.1 is an open-source text-to-video model~\cite{wan_open} from Alibaba, released in 2025. It comes in two versions: Wan2.1 Fast and Wan2.1 Professional. It supports aspect ratios of 16:9, 9:16, 1:1, 4:3, and 3:4. Wan2.1 also supports expanded prompts, offers an Inspiration Mode, and includes video sound. 
    \item \textbf{Sora}~\cite{sora}: 
    Sora is a private text-to-video generator from OpenAI, opened to the public in 2024. It has a single mode and supports 480p, 720p, and 1080p resolutions, along with 16:9, 1:1, and 9:16 aspect ratios. It can generate videos of 5s, 10s, 15s, or 20s in 30FPS. A monthly subscription of \$20 covers 480p and 720p videos at up to 5 seconds each. [Sora] also supports style presets and can generate four videos from the same prompt at once. For 1080p videos longer than 5 seconds, a \$200 monthly subscription is required. However, since most models only accept 720p requests, the \$20 subscription may be enough for many users. After reaching the daily limit, Sora offers a "relaxed mode," which still processes videos quickly—about 30 seconds per video.
    \item \textbf{Mochi-1}~\cite{mochi1}: 
    Mochi-1 is an open-source text-to-video generator developed by Genmo and opened to the public in 2024. in 2024. It includes various modes and supports 480p resolution, a 16:9 aspect ratio, and 5-second videos at 24FPS. It also offers random prompt ideas and a seed function. Interestingly, when asked to generate three people, Mochi-1 usually only creates two. It can produce two videos at once, with each video taking about three minutes to process.
    \item \textbf{LTX Video}~\cite{hcb+24}: 
    LTX Video is an open-source text-to-video generator developed by Lightricks and opened to the public in 2024. It offers various preset styles and supports 768×512 (512p) resolution. It also supports aspect ratios of 16:9, 1:1, and 9:16, as well as 5-second clips at 24FPS. LTX Video allows you to specify shot type, scene location, style presets, and references, and it supports voiceover scripts. To use it, you first generate the initial scene, then generate motion for that scene.
    \item \textbf{Pika 2.2}~\cite{pika2.2}: 
    Pika 2.2 is a private text-to-video model from Pika Labs, released in 2025. It supports Pikaframes, Pikaaffects, Pikascenes, Pikaaddition, and Pikawaps. You can generate videos in 720p or 1080p, and choose from aspect ratios of 16:9, 9:16, 1:1, 4:5, 4:3, or 5:2. It also lets you create 5-second or 10-second clips, and it supports negative prompts as well as seed inputs. I’ve had an excellent experience with Pika 2.2—its user interface is clear, comfortable, and very responsive. It can produce four videos at once in about 30 seconds each, and you can copy and edit prompts with just one click.
    \item \textbf{Dreamina}~\cite{dreamina}: 
    Dreamina is a private text-to-video model from Bytedance, released in 2024. It has four versions: Video S2.0, Video S2.0 Pro, Video P2.0 Pro, and Video 1.2. It supports Deepseek-R1 to improve prompts, and offers aspect ratios of 16:9, 21:9, 4:3, 1:1, 3:4, and 9:16. Video S2.0, Video S2.0 Pro, and Video P2.0 Pro can generate 5-second videos, with Video P2.0 Pro additionally supporting 10-second clips. Video 1.2 supports 3-, 6-, 9-, and 12-second videos. All versions run at 24FPS.
    \item \textbf{Qingying}~\cite{qingying}: 
    Qingying is the commercial version of the CogVideo family models~\cite{hdz+23,ytz+24}, which are open-source text-to-video models developed by Zhipu, released in 2023 and 2024. It offers two generation modes: Fast and Quality. It supports 5-second videos at either 60FPS or 30FPS, and can produce aspect ratios of 16:9, 9:16, 1:1, 3:4, or 4:3. Qingying also includes three advanced parameters for video style, emotional atmosphere, and camera movement mode, and it supports AI sound and AI effects.
    \item \textbf{Gen 3 Alpha}~\cite{g24}: 
    Gen 3 Alpha is a private text-to-video model developed by RunwayML, and opened to the public in 2024. It includes both Gen 3 Alpha and Gen 3 Alpha Turbo with an intensity of motion (1–10). It supports 720p and 2K resolutions, as well as 16:9 and 9:16 aspect ratios. It can generate 4s videos at 24FPS.
    \item \textbf{Hailuo}~\cite{hailuo}: 
    Hailuo is a private text-to-video model from MiniMax, released in 2025. It includes T2V-01-Director and T2V-01 for text-to-video generation. It supports 720p resolution, likely with a 16:9 aspect ratio, 6s video length, and 24FPS.
\end{itemize}

%% file: 08_append_additional_experiments.tex
\section{Additional Experiments}\label{sec:append_more_experiments}

\begin{figure}[!ht]
    \centering
    \includegraphics[width=1.0\linewidth]{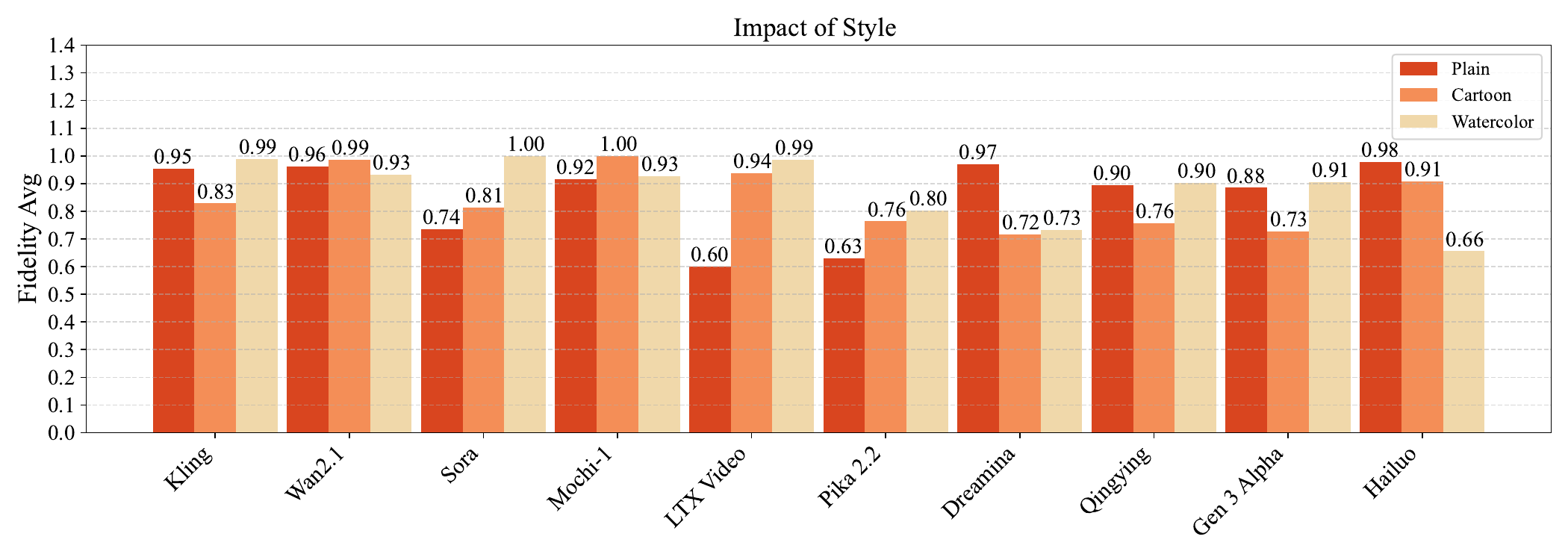}
    \caption{
    {\bf Impact of Style on Object Fidelity}.}
    \label{fig:style_fid}
\end{figure}
\begin{figure}[!ht]
    \centering
    \includegraphics[width=1.0\linewidth]{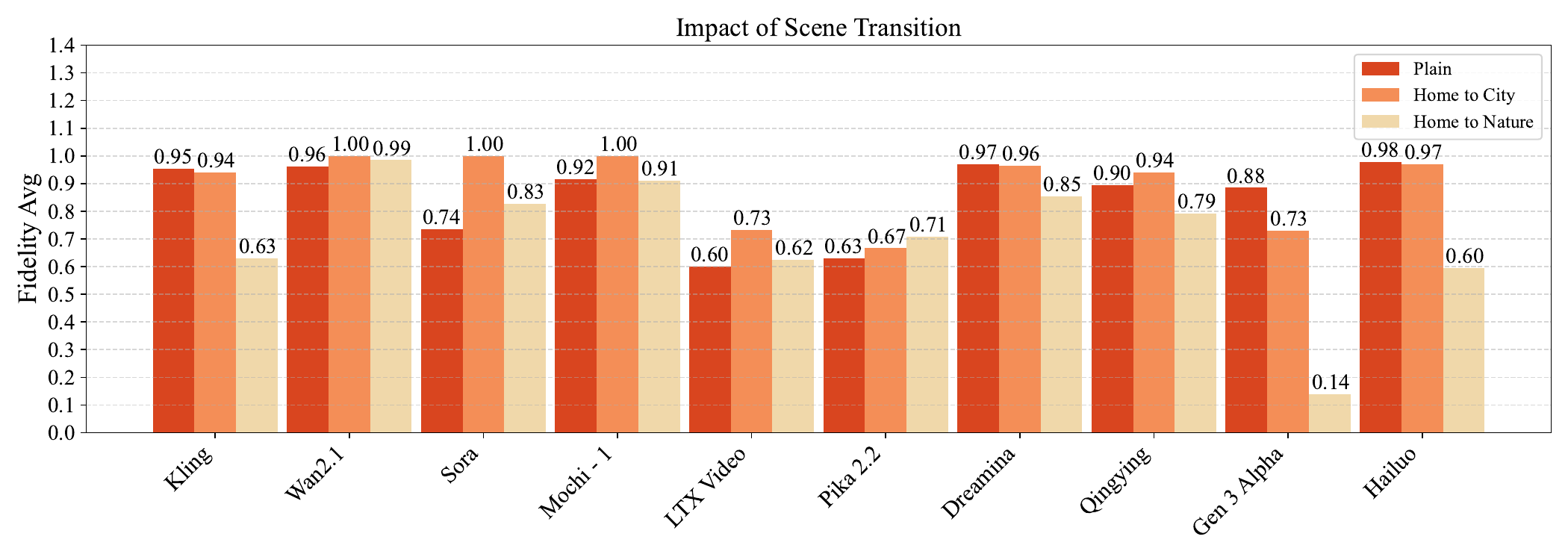}
    \caption{
    {\bf Impact of Scene Transition on Object Fidelity}. }
    \label{fig:scene_fid}
\end{figure}
\begin{figure}[!ht]
    \centering
    \includegraphics[width=1.0\linewidth]{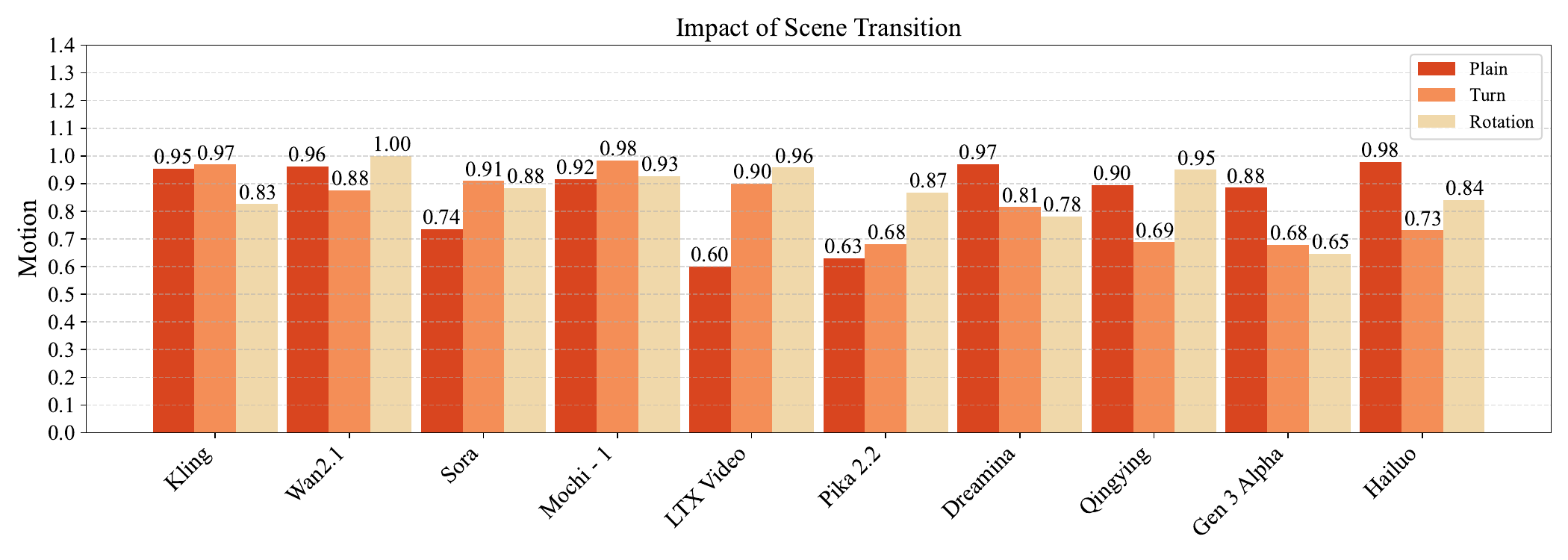}
    \caption{
    {\bf Impact of Motion on Object Fidelity}. }
    \label{fig:motion_fid}
\end{figure}
\begin{figure}[!ht]
    \centering
    \includegraphics[width=1.0\linewidth]{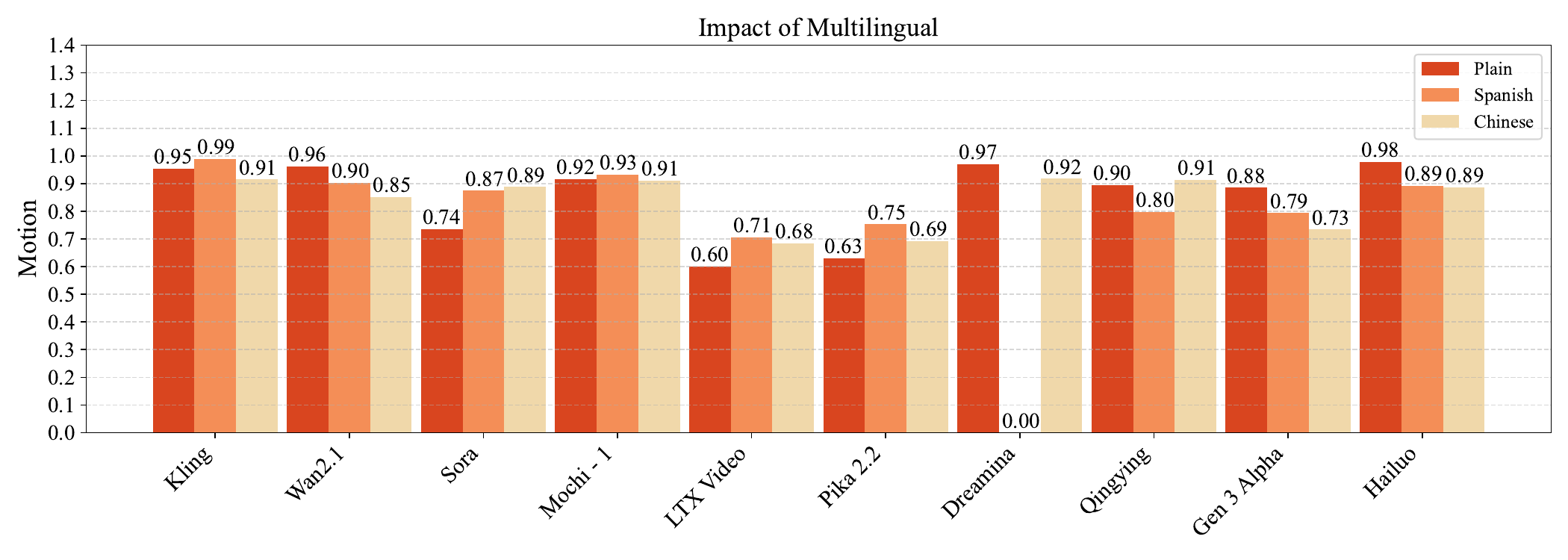}
    \caption{
    {\bf Impact of Multilingual Prompts on Object Fidelity}. }
    \label{fig:multilingual_fid}
\end{figure}

\begin{figure}[!ht]
    \centering
    \includegraphics[width=1.0\linewidth]{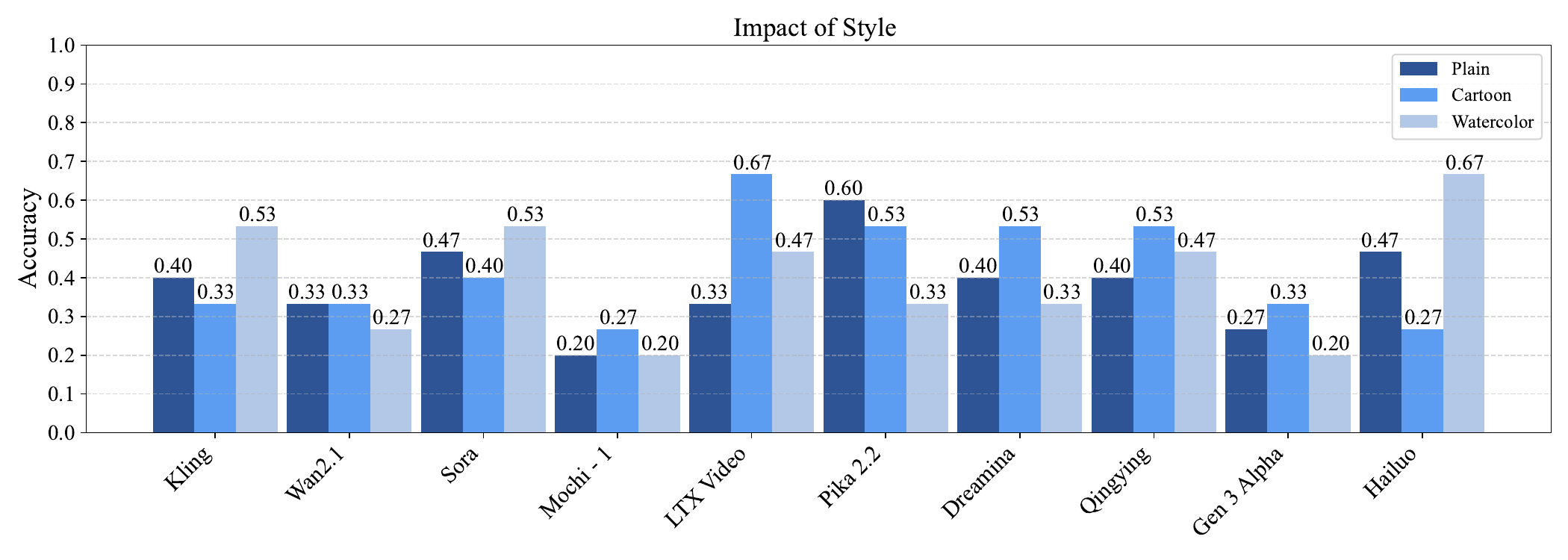}
    \caption{
    {\bf Impact of Style on Counting Accuracy}. }
    \label{fig:style_acc}
\end{figure}

In this subsection, we supplement the missing experiment results due to space limitations in Section~\ref{sec:experiments}. We first supplement the object fidelity results overlooked in ablation study, and then supplement the ablationstudy on art style and impact of object motion. 

\paragraph{Object Fidelity in Ablation Studies.} Due to the less informative nature and the orthogonality between counting and object fidelity, as summarized in Observation~\ref{obs:fid_acc_irrelevance} in Section~\ref{sec:exp_main}, we put all the missing fidelity results in ablation studies in Figures~\ref{fig:style_fid}--\ref{fig:multilingual_fid}. From the fidelity results, we can find that most factors does not have significant impact on the model fidelity, and in most cases the fidelity is significantly better than the counting accuracy. A noticeable drop of fidelity can be seen for Demina under Spanish in Figure~\ref{fig:multilingual_fid}, but this is due to its direct rejection of generating based on Among all the factors, the impact of style is the most significant, while the other factors are less impactful. This matches our previous observation in the main overall results, and our observation can be summarized a follows:

\begin{observation}
    For most factors excluding art style of videos and prompt refinement, model fidelity remains at a high level when combined with counting tasks, the factors does not have signifinat impat on model fidelity, only affecting counting accuracy. 
\end{observation}

\paragraph{Impact of Style.}  
We adopt the general Prompt Template~\ref{pro:most_general} while varying only the style of the generated videos in this ablation study. Specifically, we set the options as follows:
\begin{itemize}
    \item \textless number\textgreater: $\{1,3,5,7,9\}$;
    \item \textless object\textgreater: \texttt{'Human'}, \texttt{'Nature'}, \texttt{'Artifact'};
    \item \textless style\textgreater: \texttt{'Plain'}, \texttt{'Cartoon'}, \texttt{'Watercolor'}.
\end{itemize}
For the remaining options, we fix \textless scene transition\textgreater\ and \textless motion\textgreater\ to \texttt{'None'}. Figure~\ref{fig:style_acc} shows the impact of style on counting accuracy, and Figure~\ref{fig:style_fid} presents its impact on object fidelity.

The results indicate that the effect of style on both counting accuracy and object fidelity varies among models. For example, models like Wan2.1 and Mochi-1 show only marginal differences across styles, with similar performance in \texttt{'Plain'}, \texttt{'Cartoon'}, and \texttt{'Watercolor'} settings. In contrast, LTX Video achieves over 60\% accuracy in the \texttt{'Cartoon'} category, but its accuracy drops below 40\% in the \texttt{'Plain'} category. Hailuo also exhibits a significant change in counting accuracy, with 27\% in \texttt{'Cartoon'} style, surging to 67\% in \texttt{'Watercolor'} style. Based on these observations, we note the following:

\begin{observation}
    The impact of style on counting accuracy varies across models. Some models are highly sensitive to style changes, while others remain relatively unaffected.
\end{observation}

\begin{figure}[!ht]
    \centering
    \includegraphics[width=1.0\linewidth]{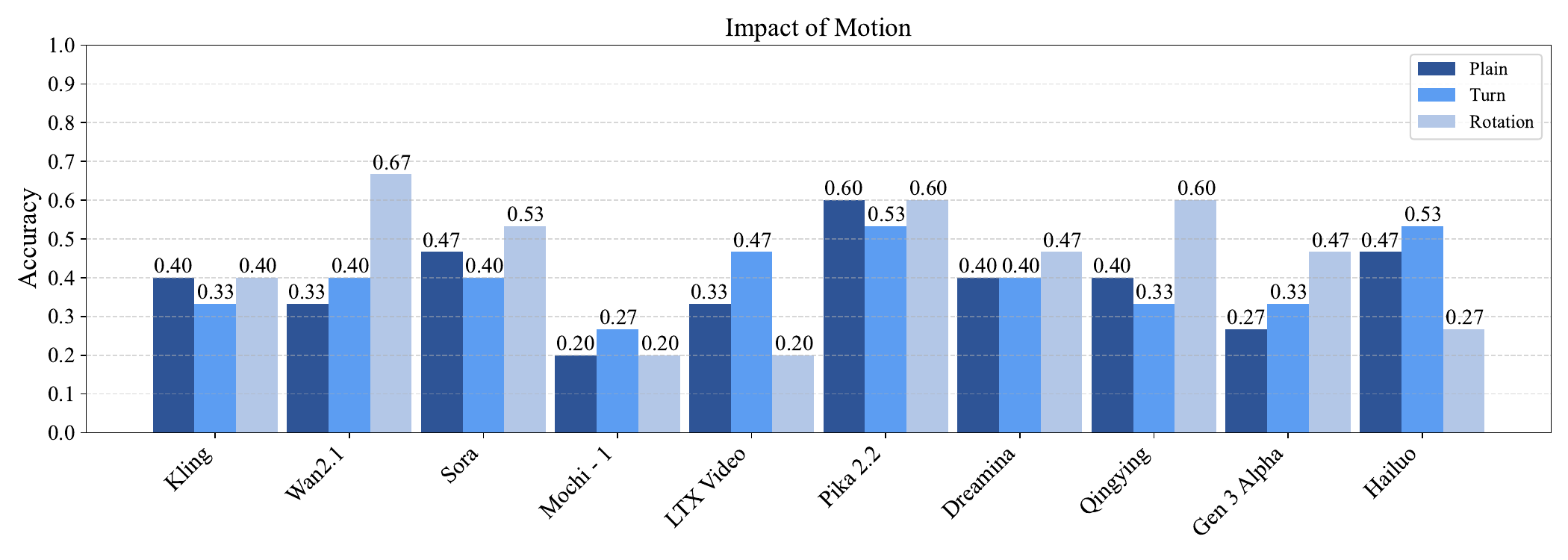}
    \caption{
    {\bf Impact of Motion on Counting Accuracy}.}
    \label{fig:motion_acc}
\end{figure}

\paragraph{Impact of Motion.} This experiment investigates a key aspect of temporal dynamics in generated videos: the motion of target entities. Following the setup of other ablation studies, we use Prompt Template~\ref{pro:most_general} while controlling for irrelevant factors. The specific options considered are:

\begin{itemize}
    \item \textless number\textgreater: $\{1,3,5,7,9\}$;
    \item \textless object\textgreater: \texttt{'Human'}, \texttt{'Nature'}, \texttt{'Artifact'};
    \item \textless motion\textgreater: \texttt{'None'}, \texttt{'Turn'}, \texttt{'Rotation'}.
\end{itemize}

We fix all irrelevant factors to their default values and analyze the impact of motion on counting accuracy, as shown in Figure~\ref{fig:motion_acc}. Overall, most models exhibit minimal sensitivity to object motion. Although exceptions exist, such as Wan2.1, which achieves 0.33 accuracy in the static setting and 0.67 in the rotational setting, these cases are relatively rare. This leads to the following observation:

\begin{observation}
    Counting accuracy remains generally stable across different motion settings, with only a few notable exceptions.
\end{observation}

%% file: 09_append_additional_qual_study.tex
\section{Additional Qualitative Studies}\label{sec:append_qual_study}

In this subsection, we present additional qualitative studies that are not covered in the main body of this paper. Specifically, we provide one qualitative result for an experiment from both Section~\ref{sec:experiments} and Appendix~\ref{sec:append_more_experiments}. The qualitative study on multilingual counting is included in Section~\ref{sec:exp_mult_ling}, while the remaining studies are presented here.

We compare the effects of ablated factors across different experiments, selecting two models with the highest overall counting accuracy from Table~\ref{tab:count_difficult_acc_fid} and two models with the lowest accuracy. Additionally, we include two models that exhibit unusual behaviors when processing these prompts, such as generating low-fidelity videos or displaying unrealistic elements like meaningless icons or faceless avatars. These qualitative analyses provide concrete insight into the limitations of text-to-video models in counting-related tasks.

\begin{figure}[!ht]
    \centering
    \includegraphics[width=1.0\linewidth]{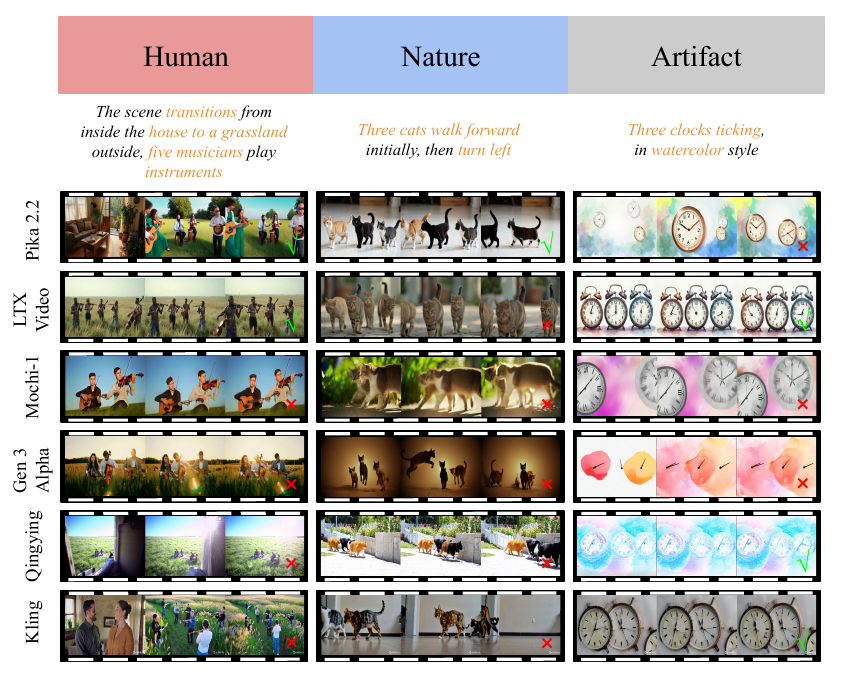}
    \caption{
    {\bf Qualitative Study on Main Results.} }
    \label{fig:7_1}
\end{figure}

\paragraph{Qualitative Study on Main Results.} This study corresponds to the results in Table~\ref{tab:count_main_acc_fid} in Section~\ref{sec:exp_main}. Figure~\ref{fig:7_1} illustrates the effect of different object categories, Human, Nature, and Artifact, on text-to-video generation. For example, when Gen 3 Alpha was prompted with 'Three cats walk forward initially, then turn left,' the output depicted a cat with two tails and no head. Similarly, for the prompt 'Three clocks ticking, in watercolor style,' the model generated a clock with only one hand.

\begin{figure}[!ht]
    \centering
    \includegraphics[width=1.0\linewidth]{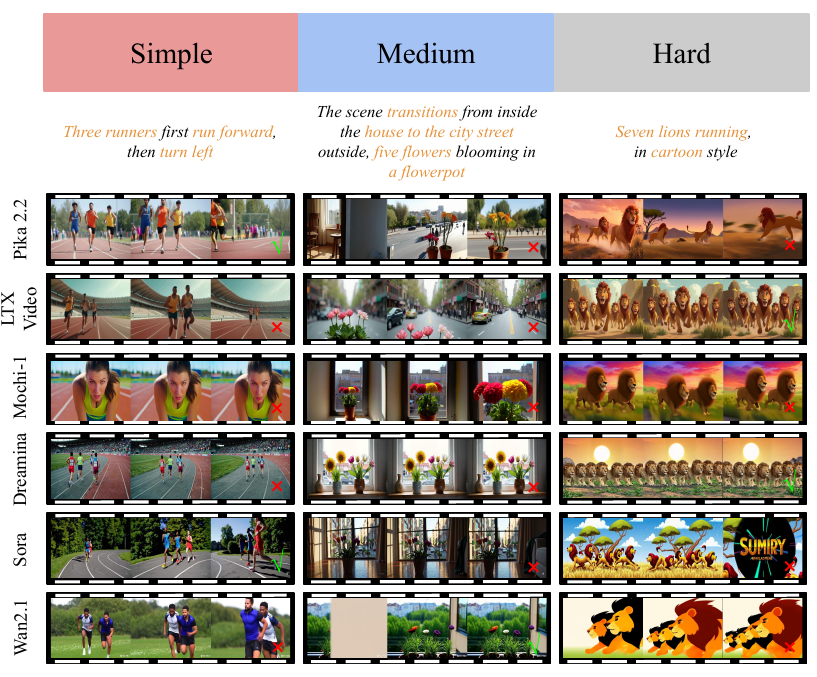}
    \caption{
    {\bf Qualitative Study on Different Difficulty Levels.} }
    \label{fig:7_2}
\end{figure}

\paragraph{Qualitative Study on Different Difficulty Levels.} This study corresponds to the results in Table~\ref{tab:count_difficult_acc_fid} in Section~\ref{sec:exp_abl}. Figure~\ref{fig:7_2} examines the impact of different difficulty levels, Simple, Medium, and Hard, on text-to-video generation. For instance, when Mochi-1 processed the prompt 'Three runners first run forward, then turn left,' only one runner continued facing forward. Similarly, Sora, given 'Seven lions running, in cartoon style,' generated a video ending with an unidentified logo. Wan2.1, when prompted with 'The scene transitions from inside the house to the city street outside, five flowers blooming in a flowerpot,' depicted three flower pots instead of one.

\begin{figure}[!ht]
    \centering
    \includegraphics[width=1.0\linewidth]{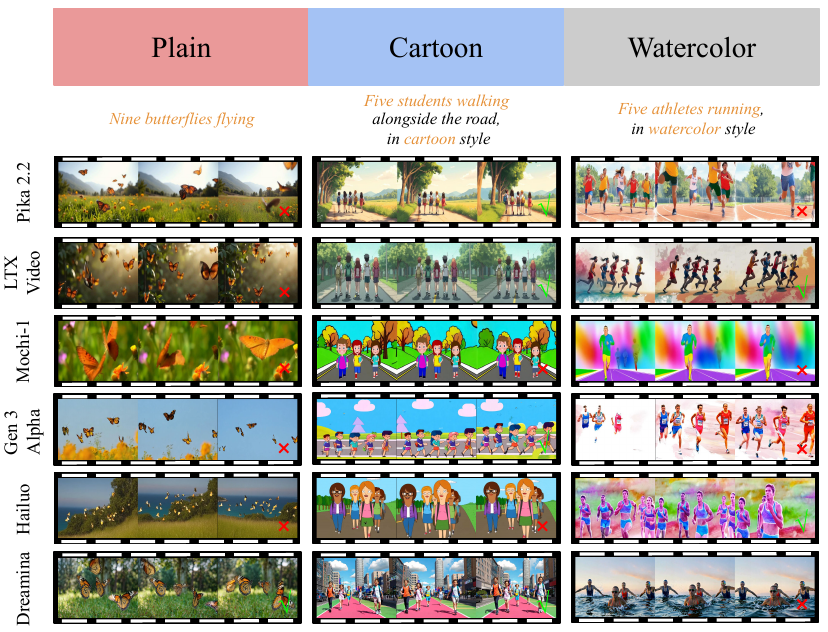}
    \caption{
    {\bf Qualitative Study on the Impact of Style.}}
    \label{fig:7_5}
\end{figure}

\paragraph{Qualitative Study on the Impact of Style.} This study corresponds to the results in Figures~\ref{fig:style_acc} and~\ref{fig:scene_fid} in Section~\ref{sec:append_more_experiments}. Figure~\ref{fig:7_5} illustrates the influence of different styles, Plain, Cartoon, and Watercolor, on text-to-video generation. For instance, Hailuo, prompted with 'Nine butterflies flying,' generated an excessive number of butterflies. Dreamina, when given 'Five students walking alongside the road, in cartoon style,' produced abstract-faced students. However, when asked to generate 'Five athletes running, in watercolor style,' Dreamina instead depicted athletes swimming.

\begin{figure}[!ht]
    \centering
    \includegraphics[width=1.0\linewidth]{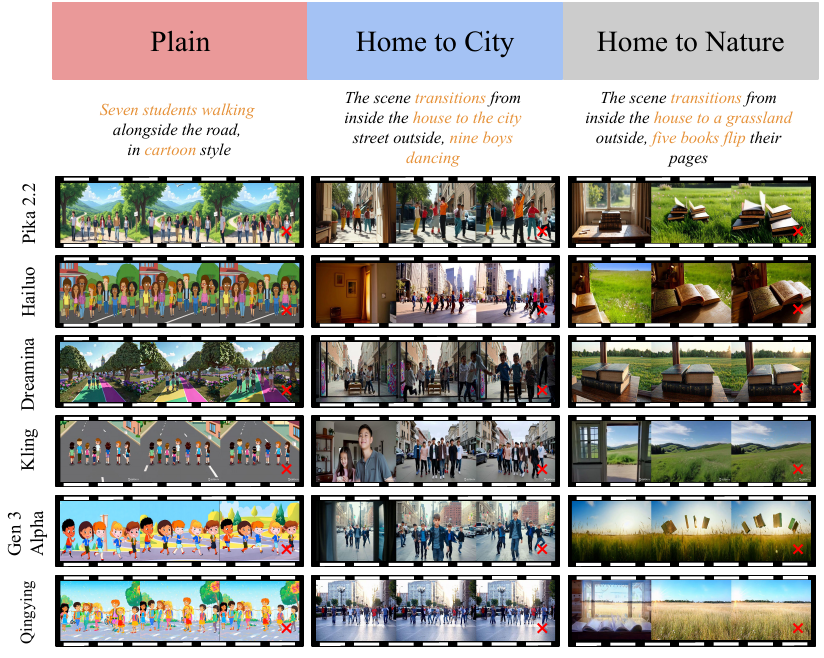}
    \caption{
    {\bf Qualitative Study on Scene Transition Results.} }
    \label{fig:7_6}
\end{figure}

\paragraph{Qualitative Study on the Impact of Scene Transition.} This study corresponds to the results in Figures~\ref{fig:scene_acc} and~\ref{fig:scene_fid} in Section~\ref{sec:exp_abl}. Figure~\ref{fig:7_6} examines the effect of different scene transitions, Plain, Home to City, and Home to Nature, on text-to-video generation. For the prompt 'The scene transitions from inside the house to a grassland outside, five books flip their pages,' Gen 3 Alpha produced books floating in the sky. Similarly, for 'The scene transitions from inside the house to the city street outside, nine boys dancing,' Qingying generated multiple figures with unnatural movements and abstract faces.

\begin{figure}[!ht]
    \centering
    \includegraphics[width=1.0\linewidth]{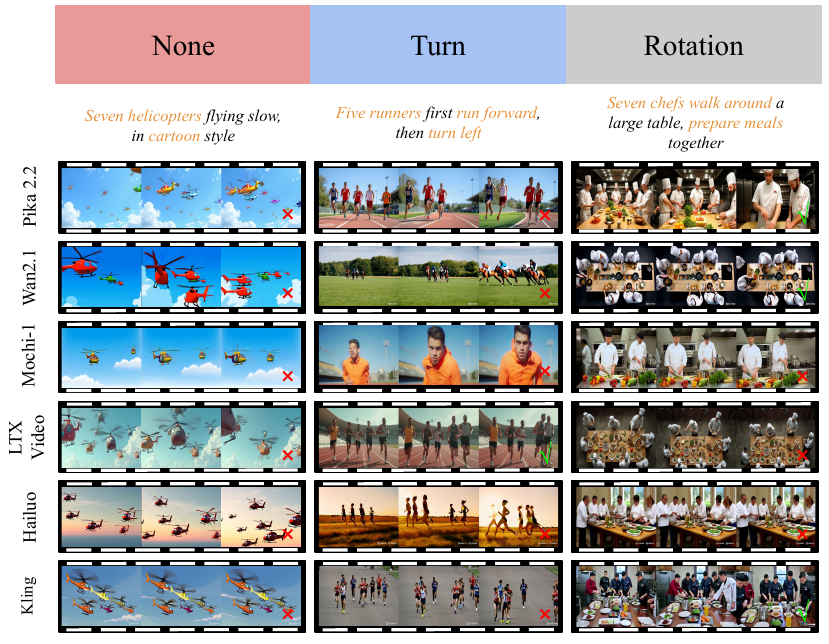}
    \caption{
    {\bf Qualitative Study on Different Motion.} }
    \label{fig:7_7}
\end{figure}

\paragraph{Qualitative Study on the Impact of Object Motion.} This study corresponds to the results in Figures~\ref{fig:motion_acc} and~\ref{fig:motion_fid} in Section~\ref{sec:append_more_experiments}. Figure~\ref{fig:7_7} explores the influence of different motion settings, None, Turn, and Rotation, on text-to-video generation. When given the prompt 'Five runners first run forward, then turn left,' Mochi-1 produced a video where the runner only moved forward. In contrast, Hailuo generated a sequence where a runner suddenly appeared midway through the video.

\begin{figure}[!ht]
    \centering
    \includegraphics[width=1.0\linewidth]{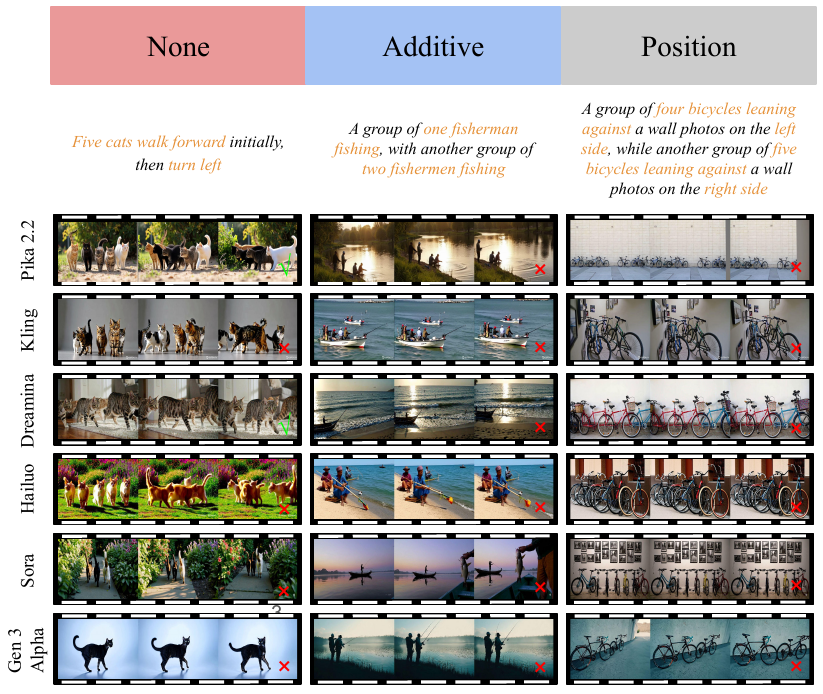}
    \caption{
    {\bf Qualitative Study on Prompt Refinement Results.} }
    \label{fig:7_3}
\end{figure}

\paragraph{Qualitative Study on Prompt Refinement Results.} This study corresponds to the results in Figures~\ref{fig:prompt_refinement_acc} and~\ref{fig:prompt_refinement_fid} in Section~\ref{sec:exp_refine}. Figure~\ref{fig:7_3} investigates the effect of different prompt refinement strategies, None, Additive, and Position, on text-to-video generation. For the prompt 'A group of four bicycles leaning against a wall on the left side, while another group of five bicycles leans against a wall on the right side,' Dreamina generated fragmented bicycles that appeared incomplete. Gen 3 Alpha, processing the same prompt, produced bicycles with distorted handlebars.

%% file: 10_append_video_examples.tex
\section{Video Examples}\label{sec:append_video_example}

\begin{figure}[!ht]
    \centering
    \includegraphics[width=1.0\linewidth]{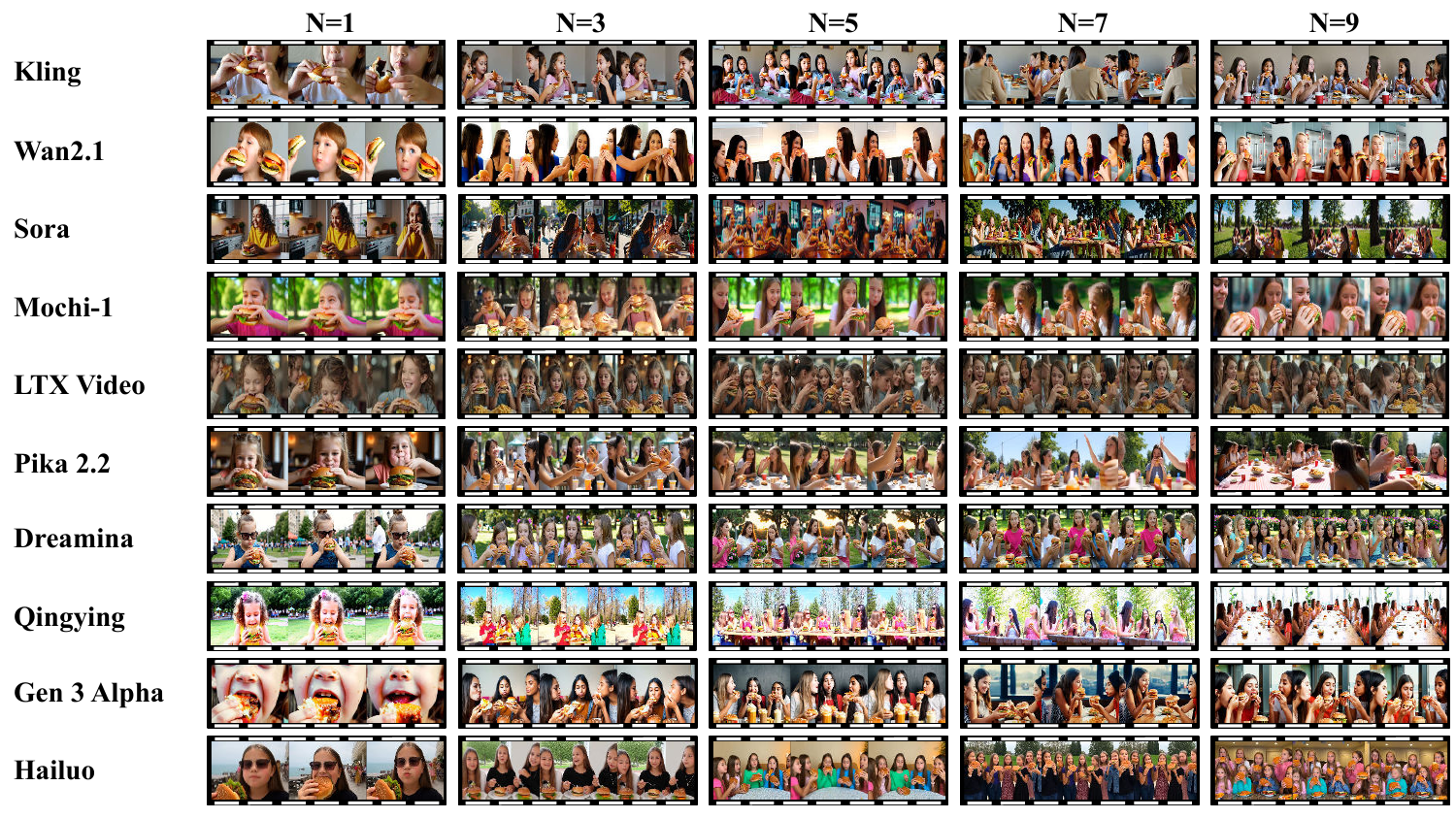}
    \caption{
    {\bf Counting Girls Results on 10 Models}. }
    \label{fig:1_1}
\end{figure}

\begin{figure}[!ht]
    \centering
    \includegraphics[width=1.0\linewidth]{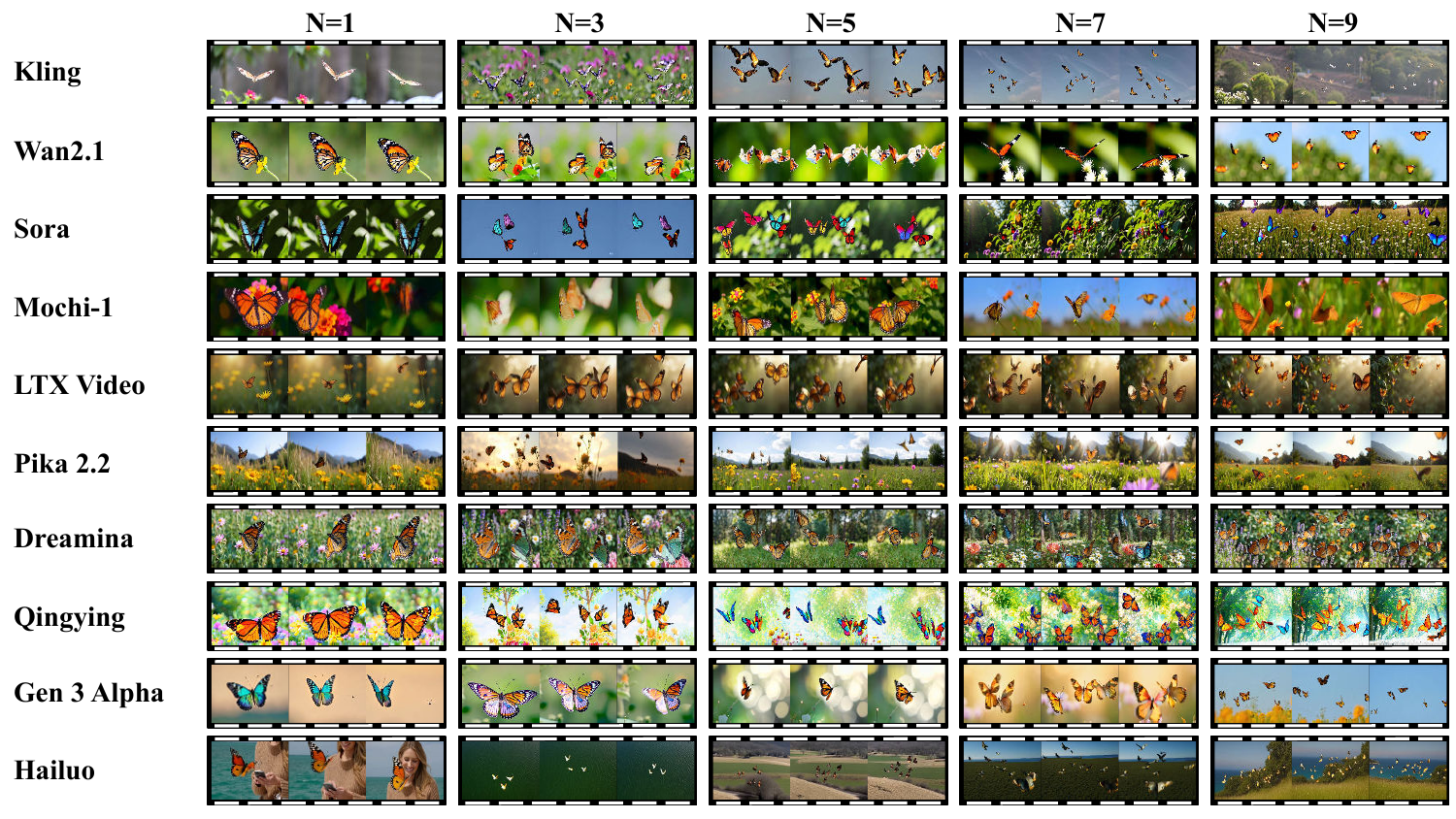}
    \caption{
    {\bf Counting Butterflies Results on 10 Models}. }
    \label{fig:1_2}
\end{figure}

\begin{figure}[!ht]
    \centering
    \includegraphics[width=1.0\linewidth]{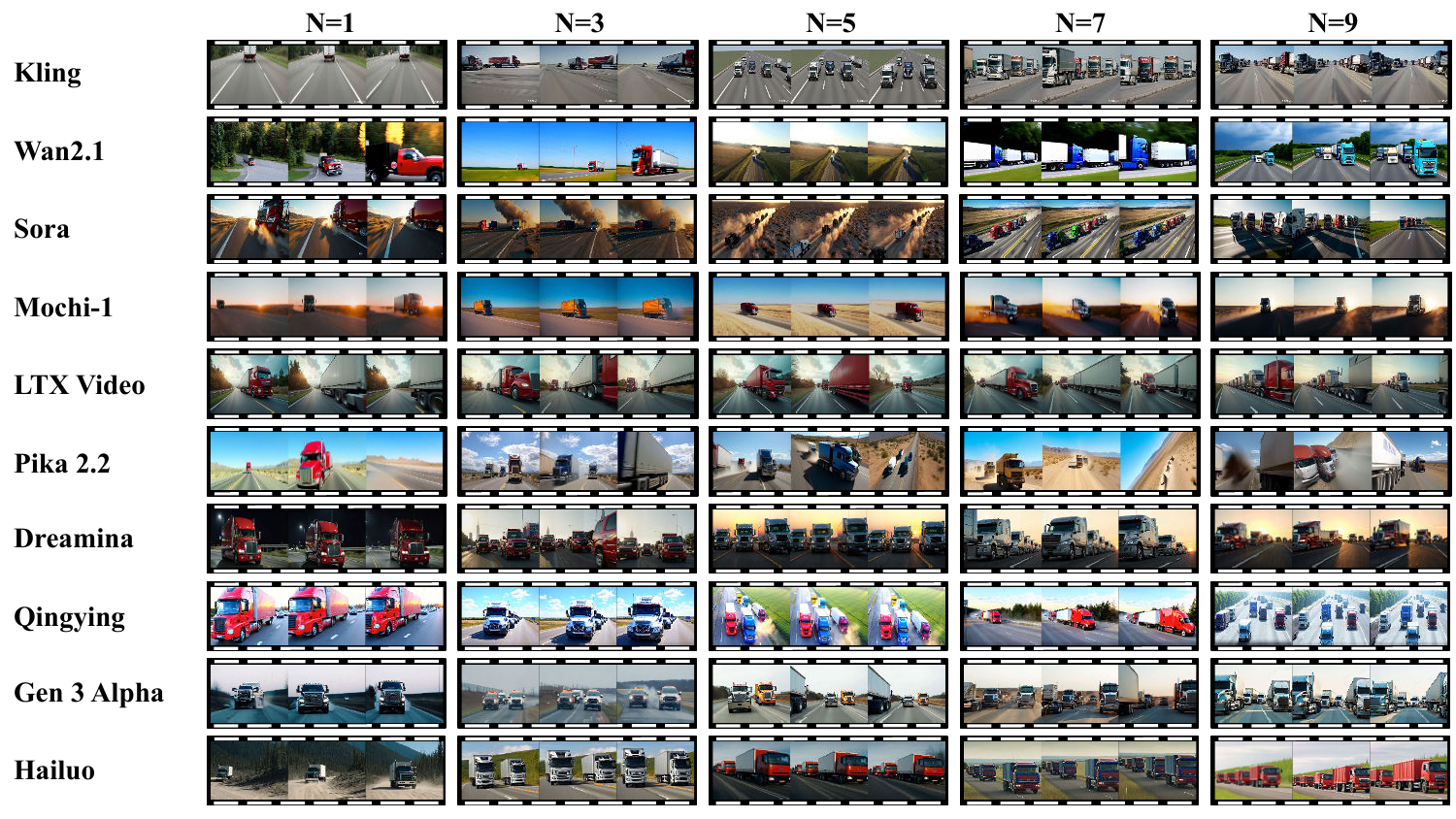}
    \caption{
    {\bf Counting Trucks Results on 10 Models}. }
    \label{fig:1_3}
\end{figure}

\begin{figure}[!ht]
    \centering
    \includegraphics[width=1.0\linewidth]{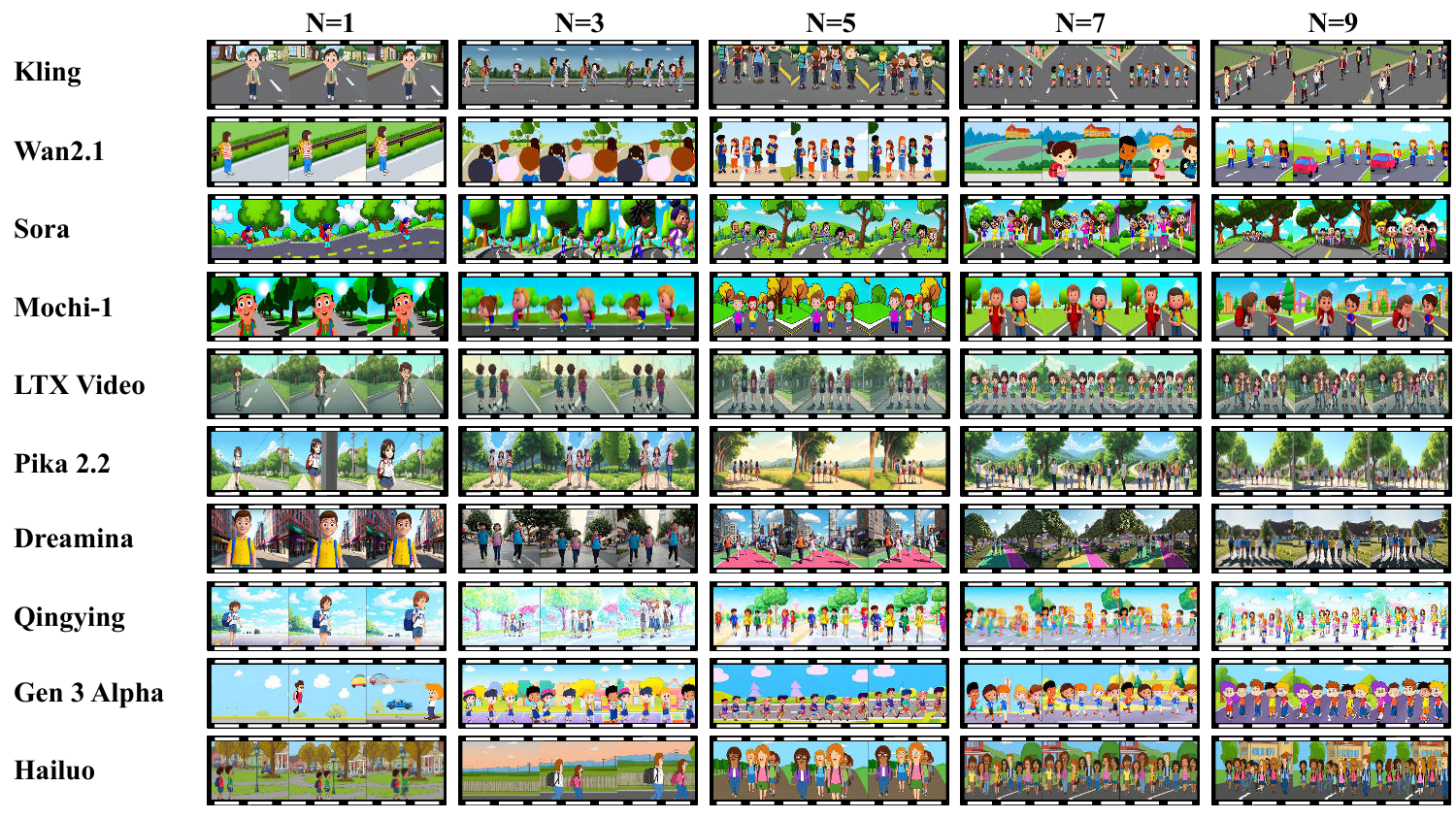}
    \caption{
    {\bf Counting Students Results on 10 Models}. }
    \label{fig:2_1}
\end{figure}

\begin{figure}[!ht]
    \centering
    \includegraphics[width=1.0\linewidth]{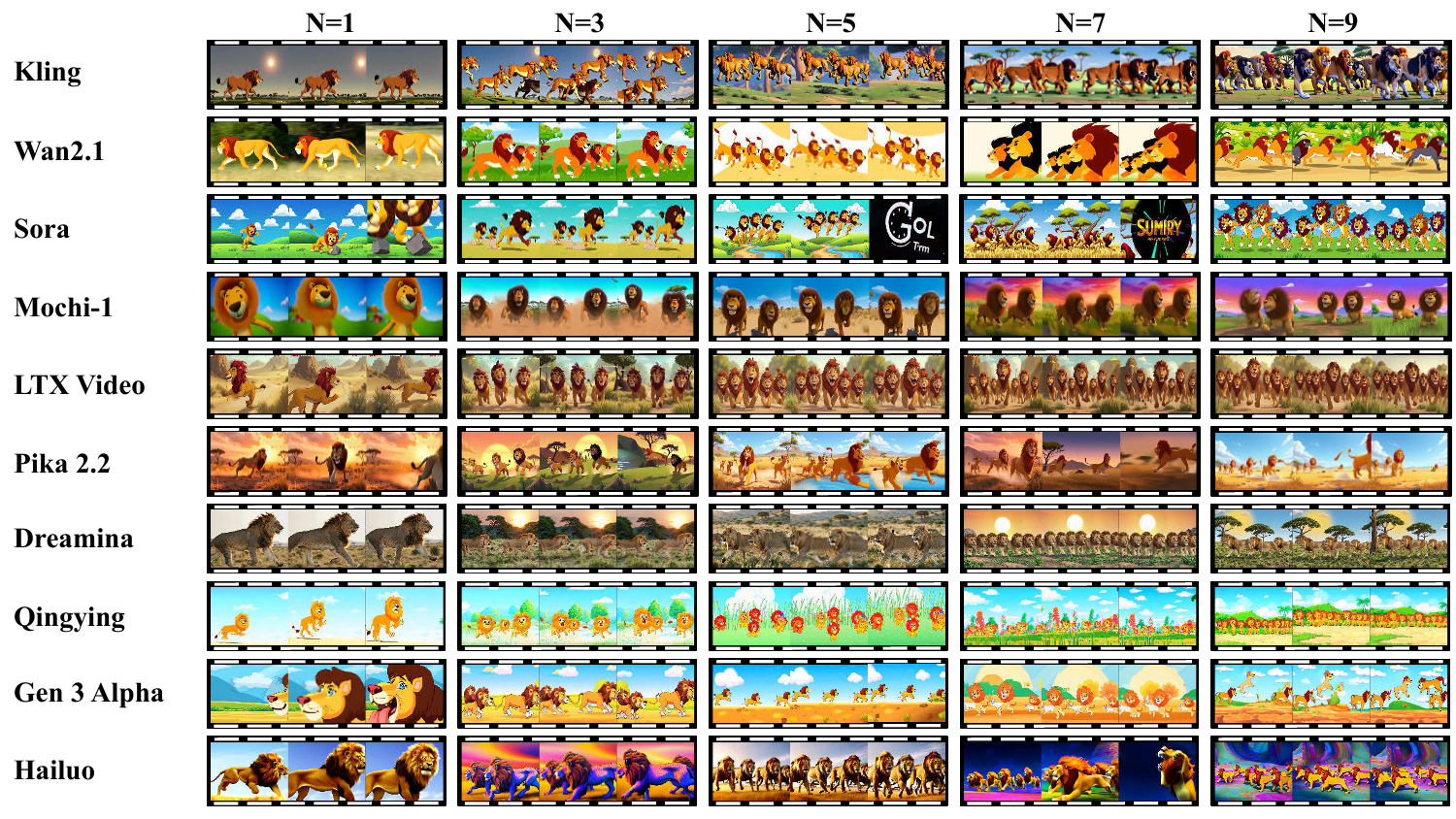}
    \caption{
    {\bf Counting Lions Results on 10 Models}. }
    \label{fig:2_2}
\end{figure}

\begin{figure}[!ht]
    \centering
    \includegraphics[width=1.0\linewidth]{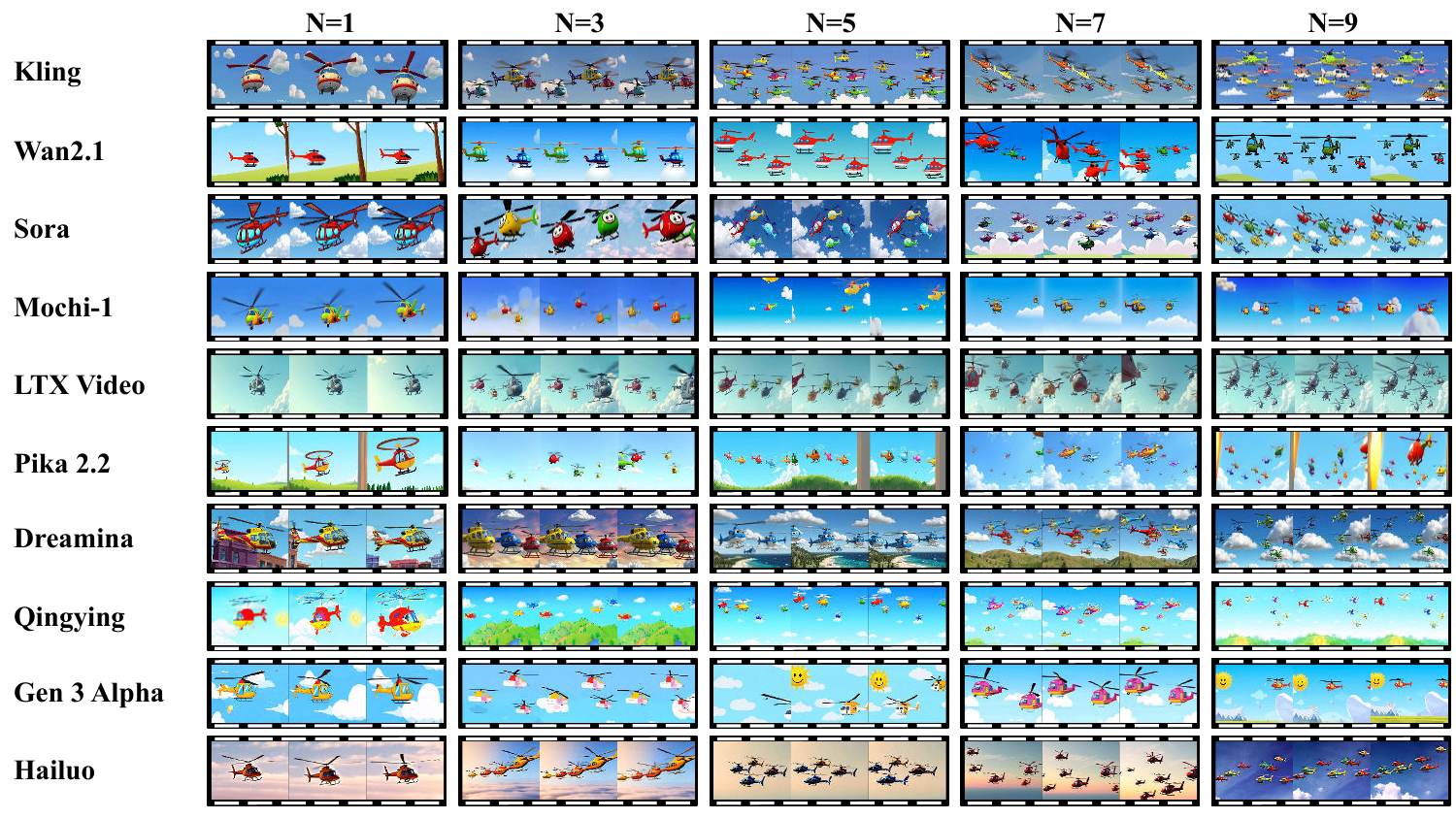}
    \caption{
    {\bf Counting Helicopters Results on 10 Models}. }
    \label{fig:2_3}
\end{figure}

\begin{figure}[!ht]
    \centering
    \includegraphics[width=1.0\linewidth]{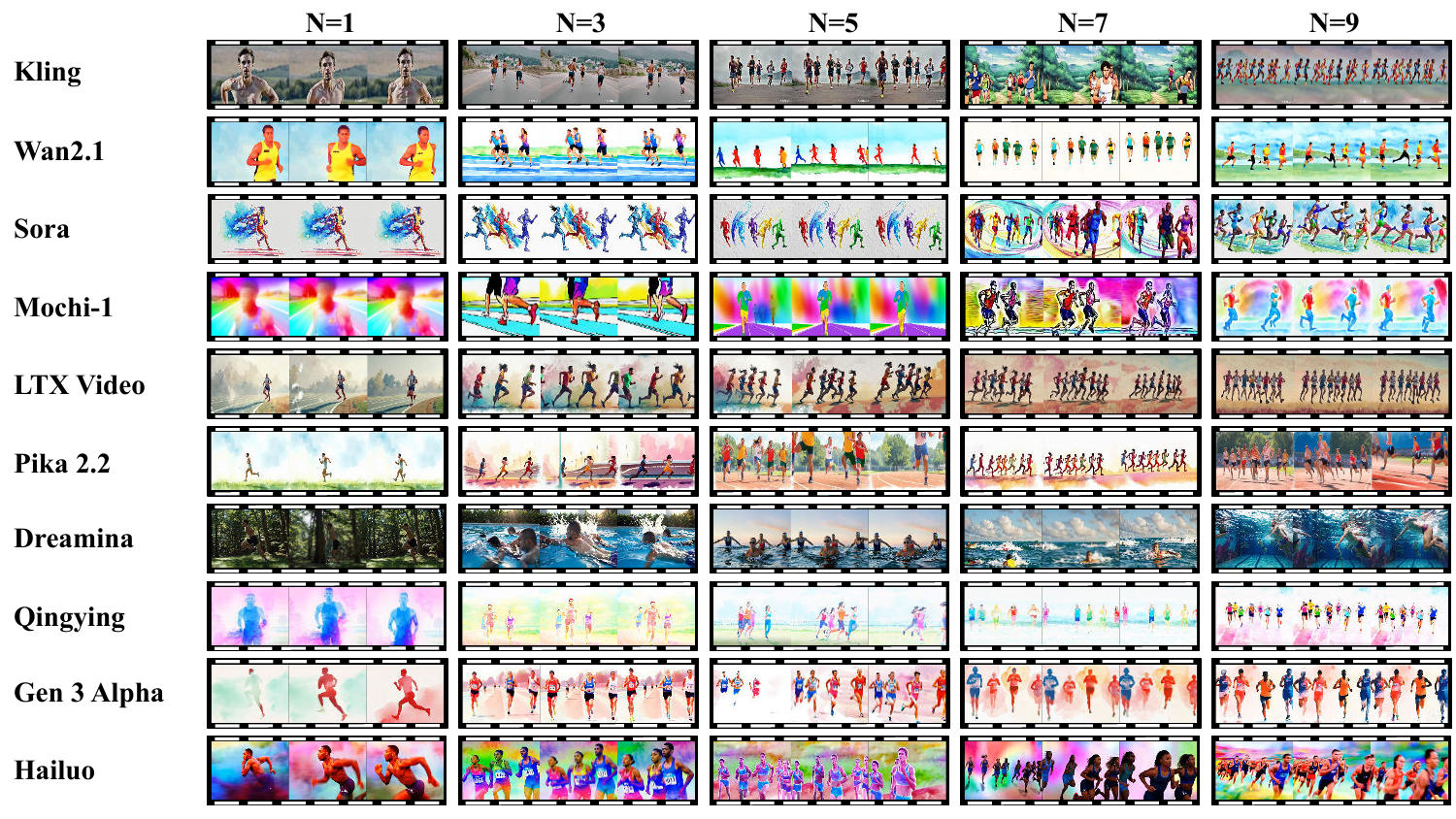}
    \caption{
    {\bf Counting Athletes Results on 10 Models}. }
    \label{fig:2_4}
\end{figure}

\begin{figure}[!ht]
    \centering
    \includegraphics[width=1.0\linewidth]{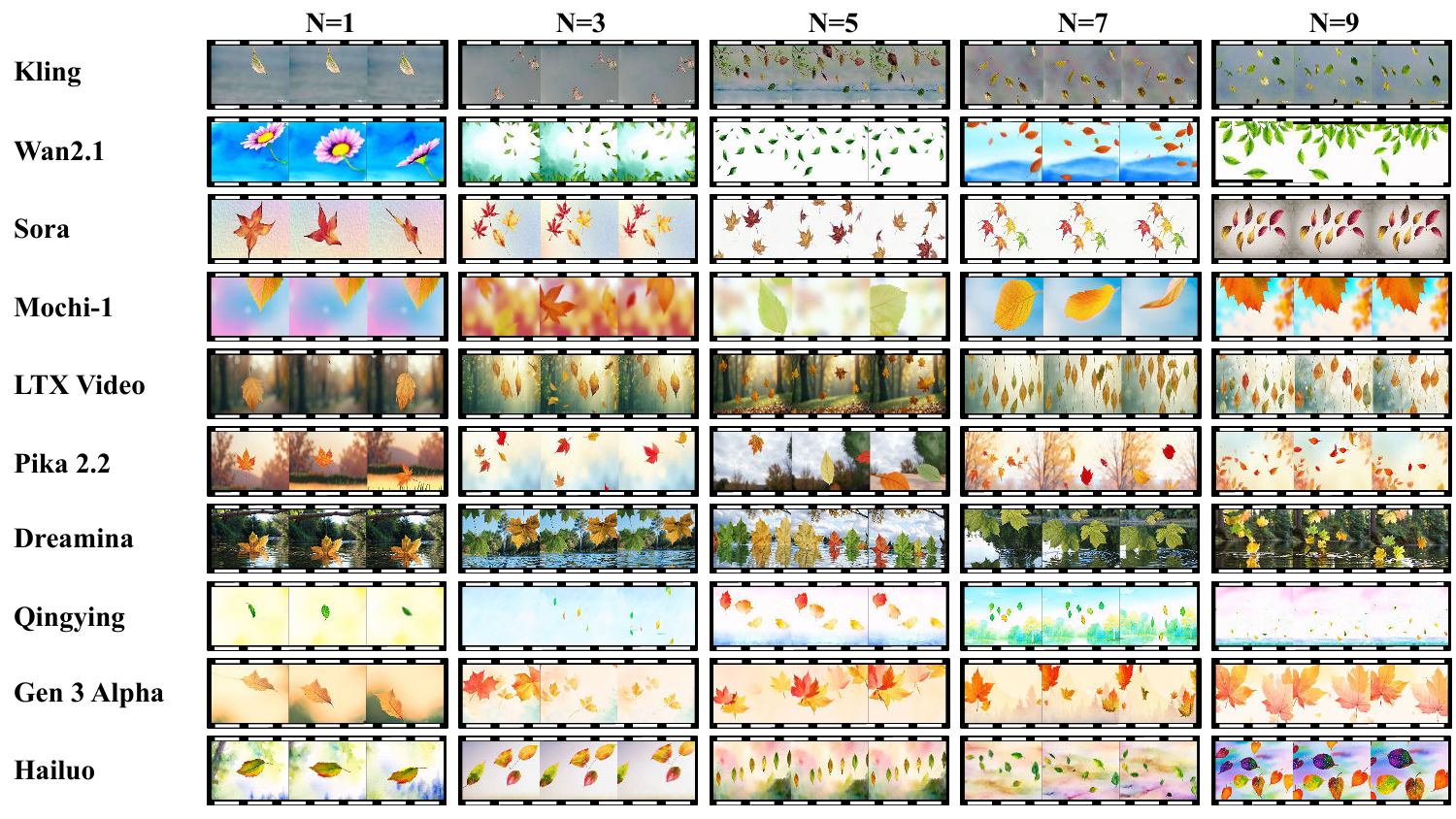}
    \caption{
    {\bf Counting Leaves Results on 10 Models}. }
    \label{fig:2_5}
\end{figure}

\begin{figure}[!ht]
    \centering
    \includegraphics[width=1.0\linewidth]{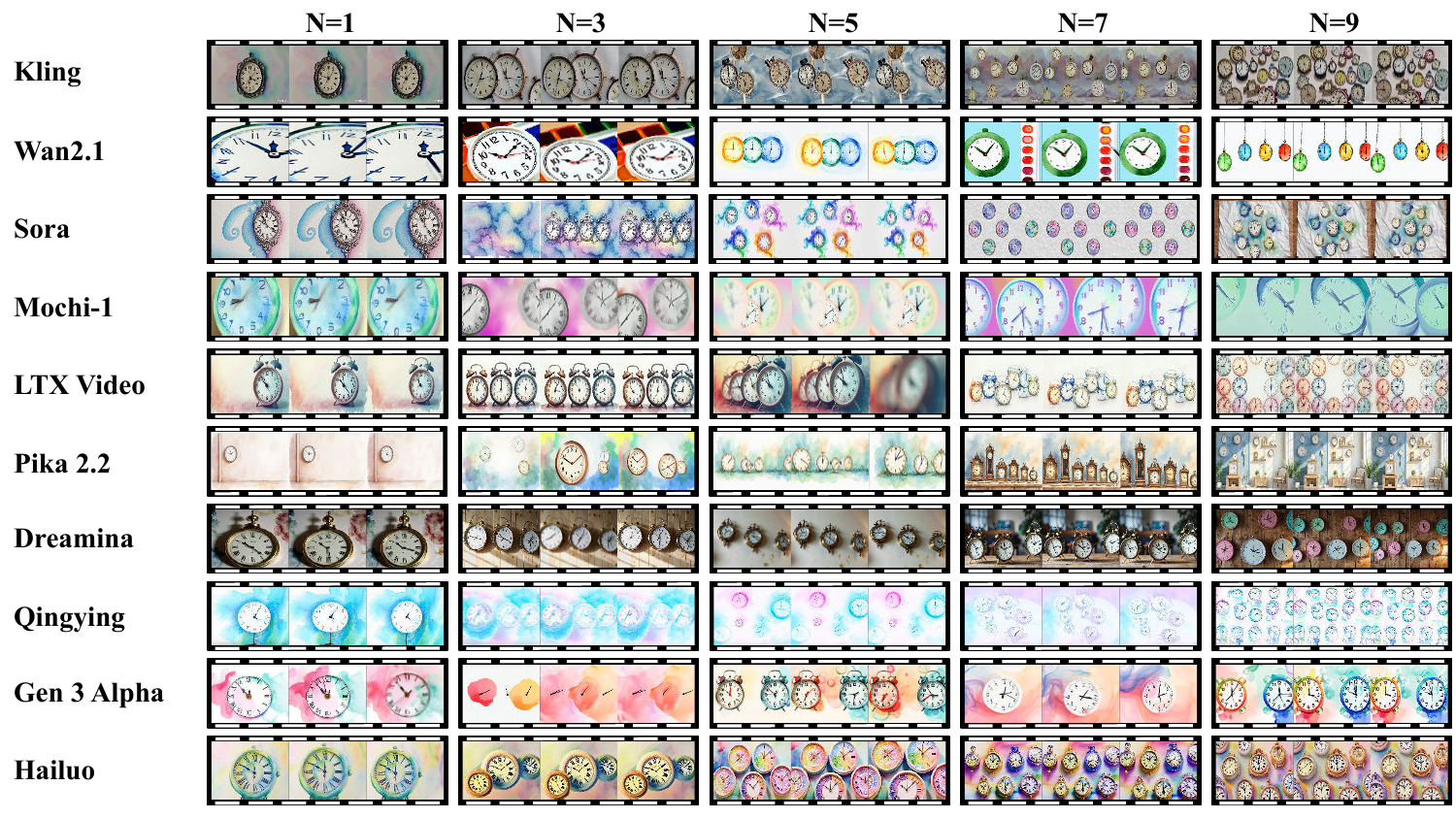}
    \caption{
    {\bf Counting Clocks Results on 10 Models}. }
    \label{fig:2_6}
\end{figure}

\begin{figure}[!ht]
    \centering
    \includegraphics[width=1.0\linewidth]{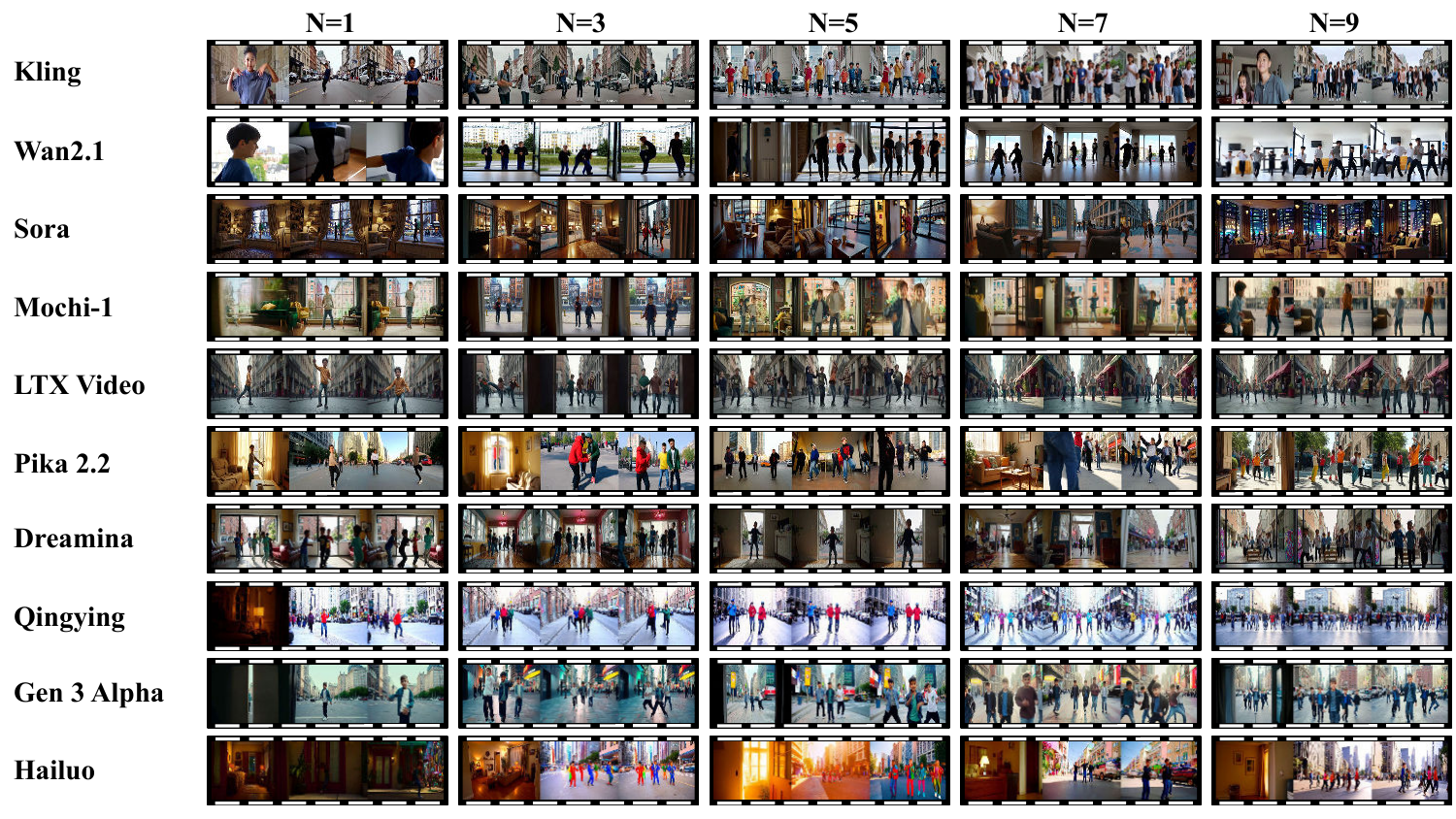}
    \caption{
    {\bf Counting Boys Results on 10 Models}. }
    \label{fig:3_1}
\end{figure}

\begin{figure}[!ht]
    \centering
    \includegraphics[width=1.0\linewidth]{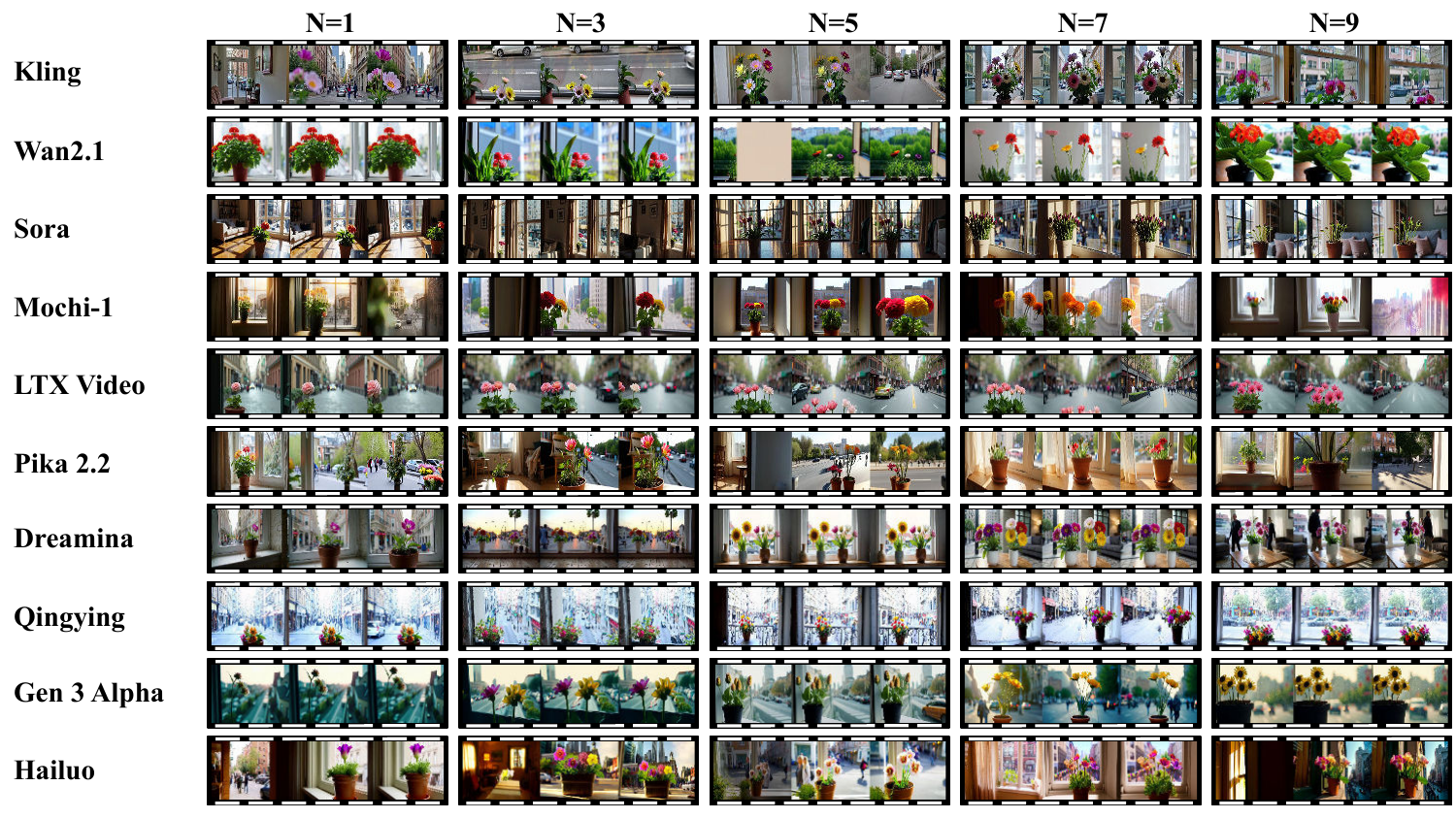}
    \caption{
    {\bf Counting Flowers Results on 10 Models}. }
    \label{fig:3_2}
\end{figure}

\begin{figure}[!ht]
    \centering
    \includegraphics[width=1.0\linewidth]{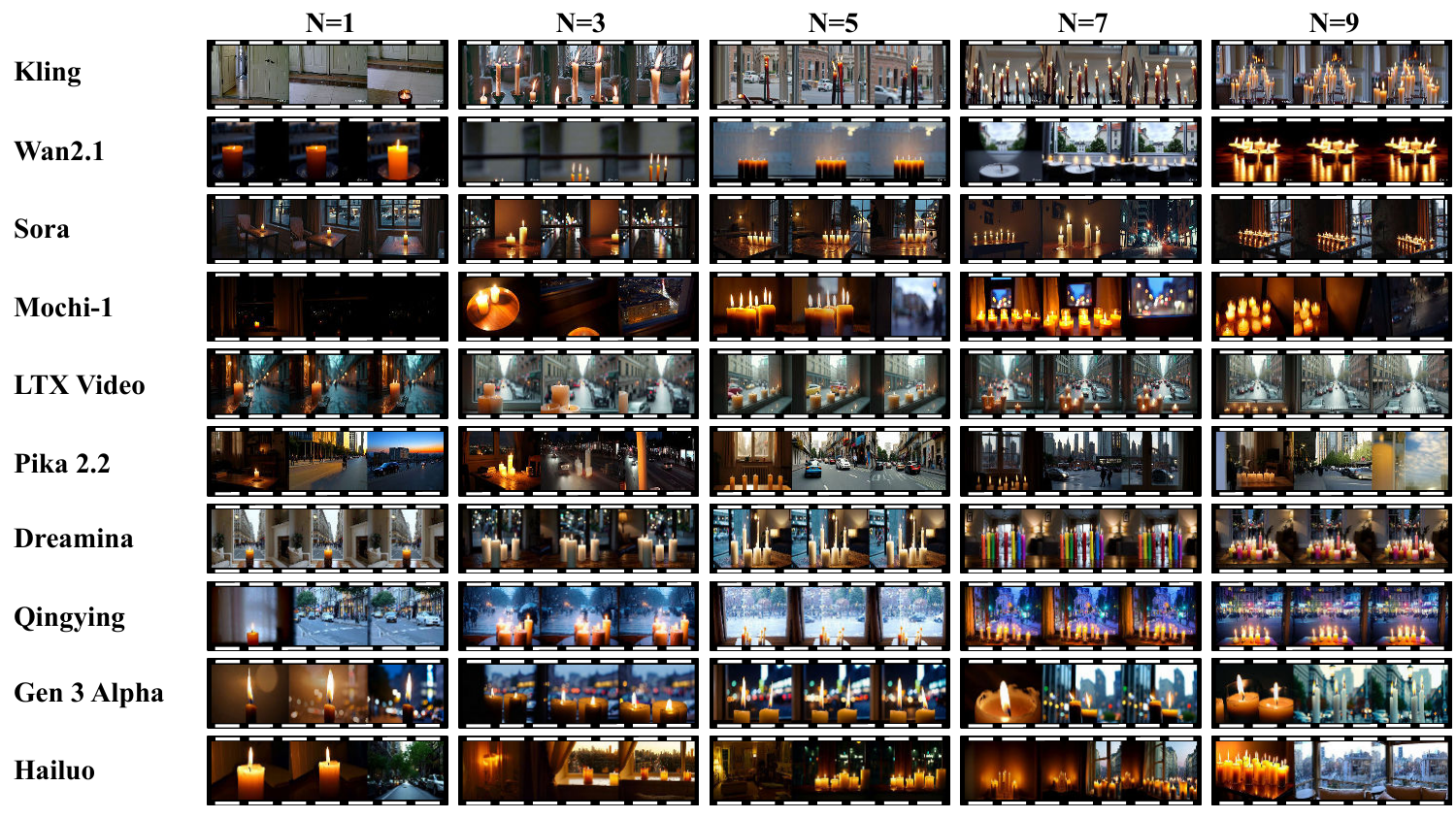}
    \caption{
    {\bf Counting Candles Results on 10 Models}. }
    \label{fig:3_3}
\end{figure}

\begin{figure}[!ht]
    \centering
    \includegraphics[width=1.0\linewidth]{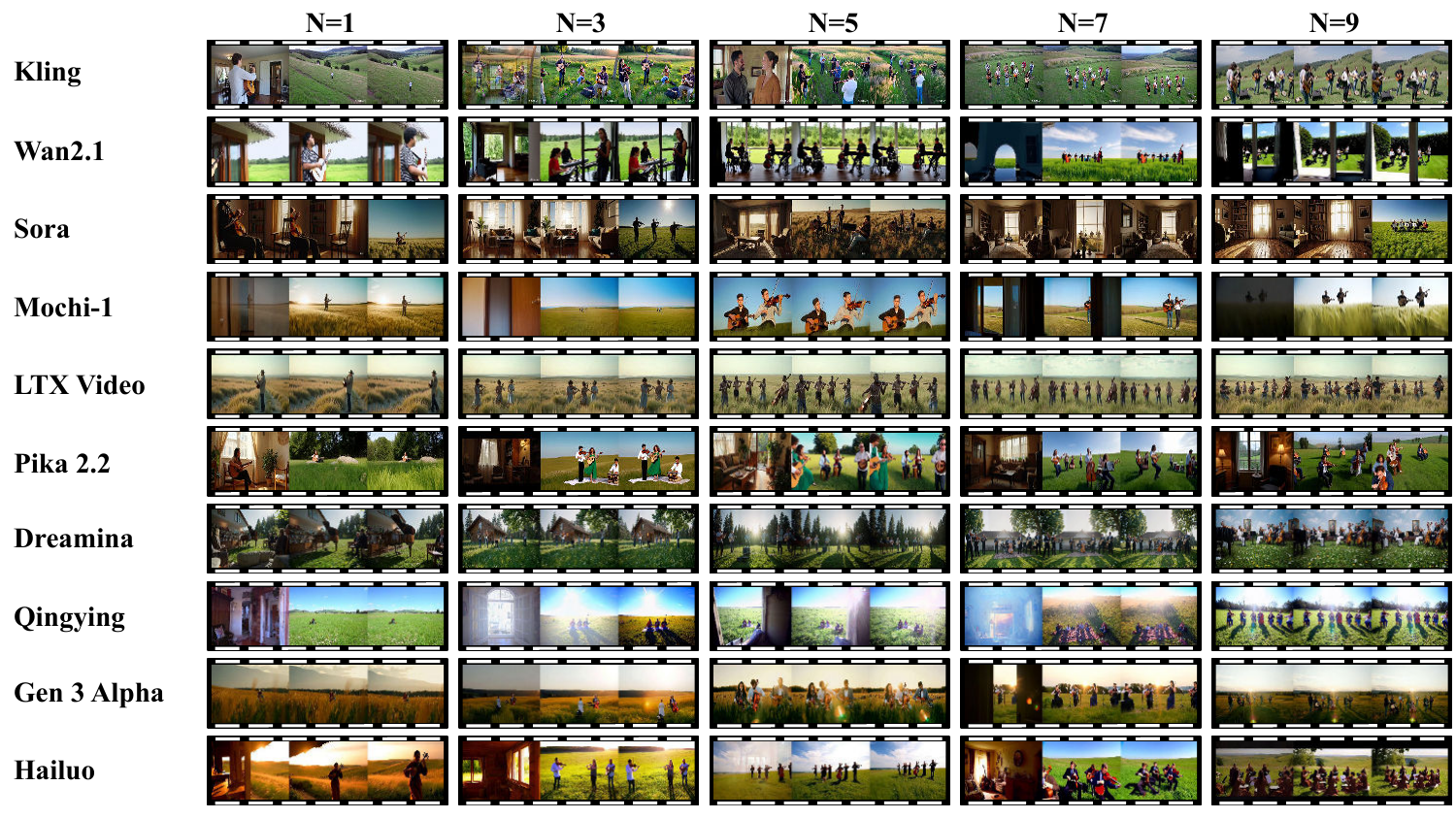}
    \caption{
    {\bf Counting Musicians Results on 10 Models}. }
    \label{fig:3_4}
\end{figure}

\begin{figure}[!ht]
    \centering
    \includegraphics[width=1.0\linewidth]{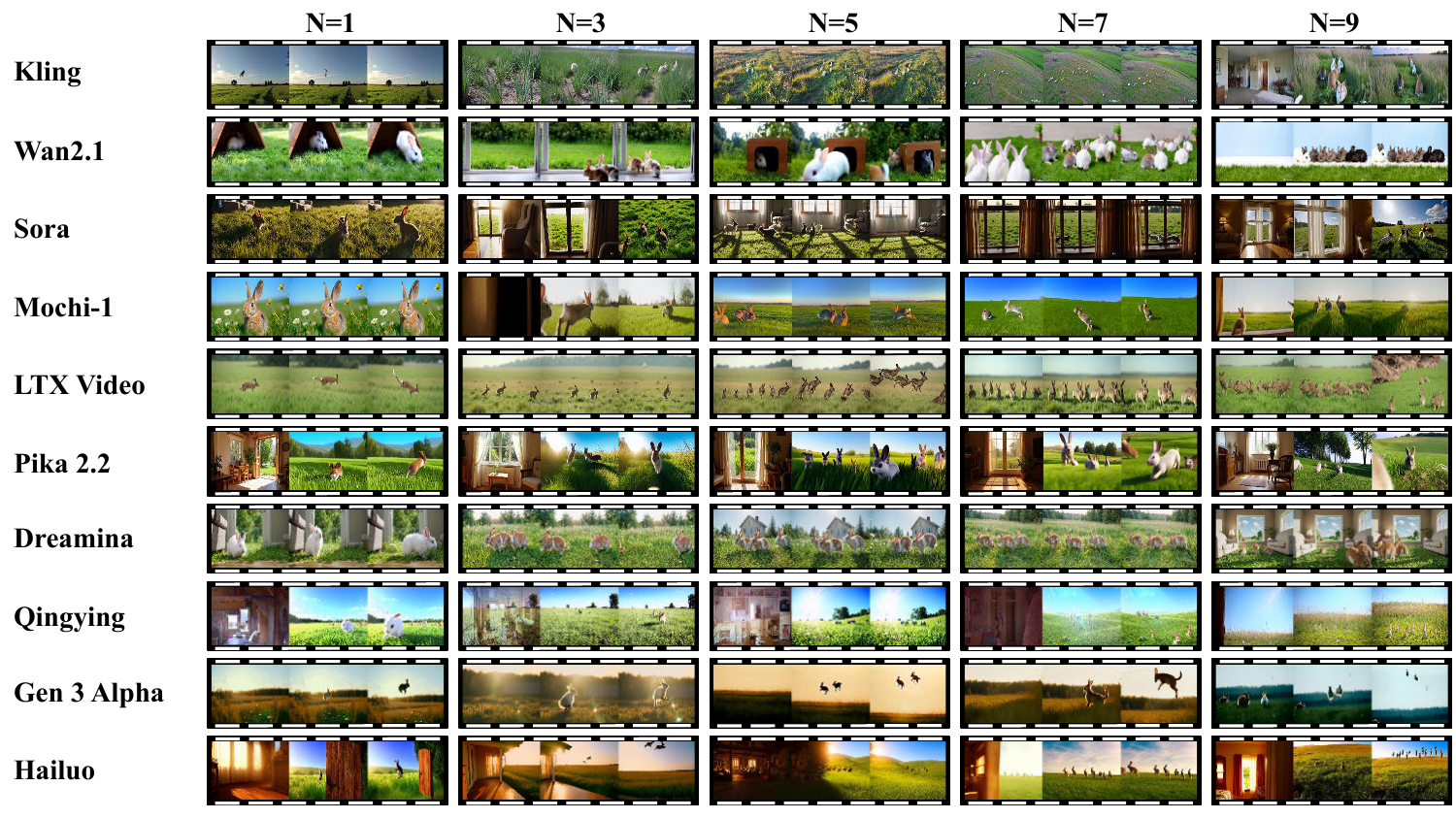}
    \caption{
    {\bf Counting Rabbits Results on 10 Models}. }
    \label{fig:3_5}
\end{figure}

\begin{figure}[!ht]
    \centering
    \includegraphics[width=1.0\linewidth]{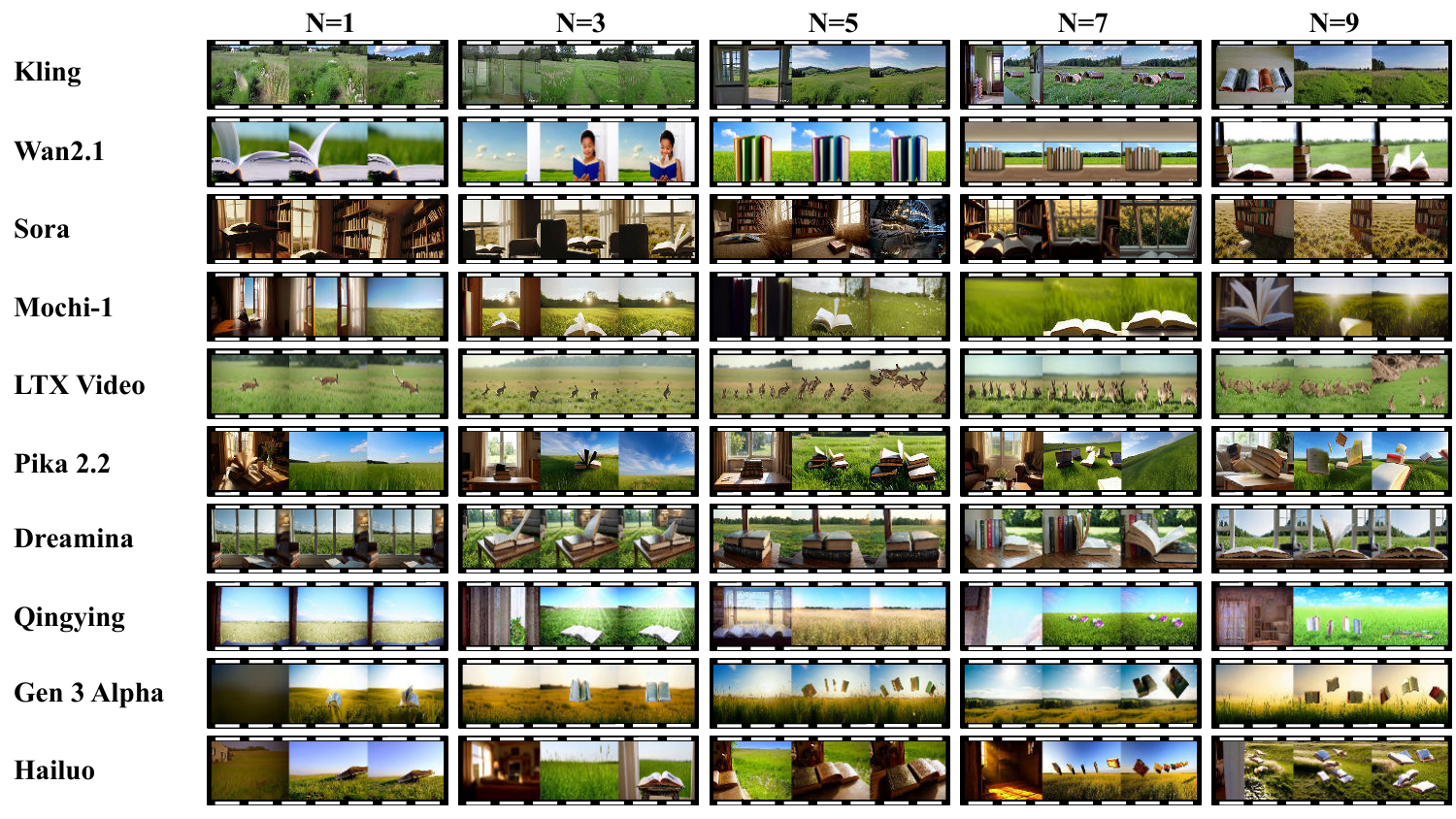}
    \caption{
    {\bf Counting Books Results on 10 Models}. }
    \label{fig:3_6}
\end{figure}

\begin{figure}[!ht]
    \centering
    \includegraphics[width=1.0\linewidth]{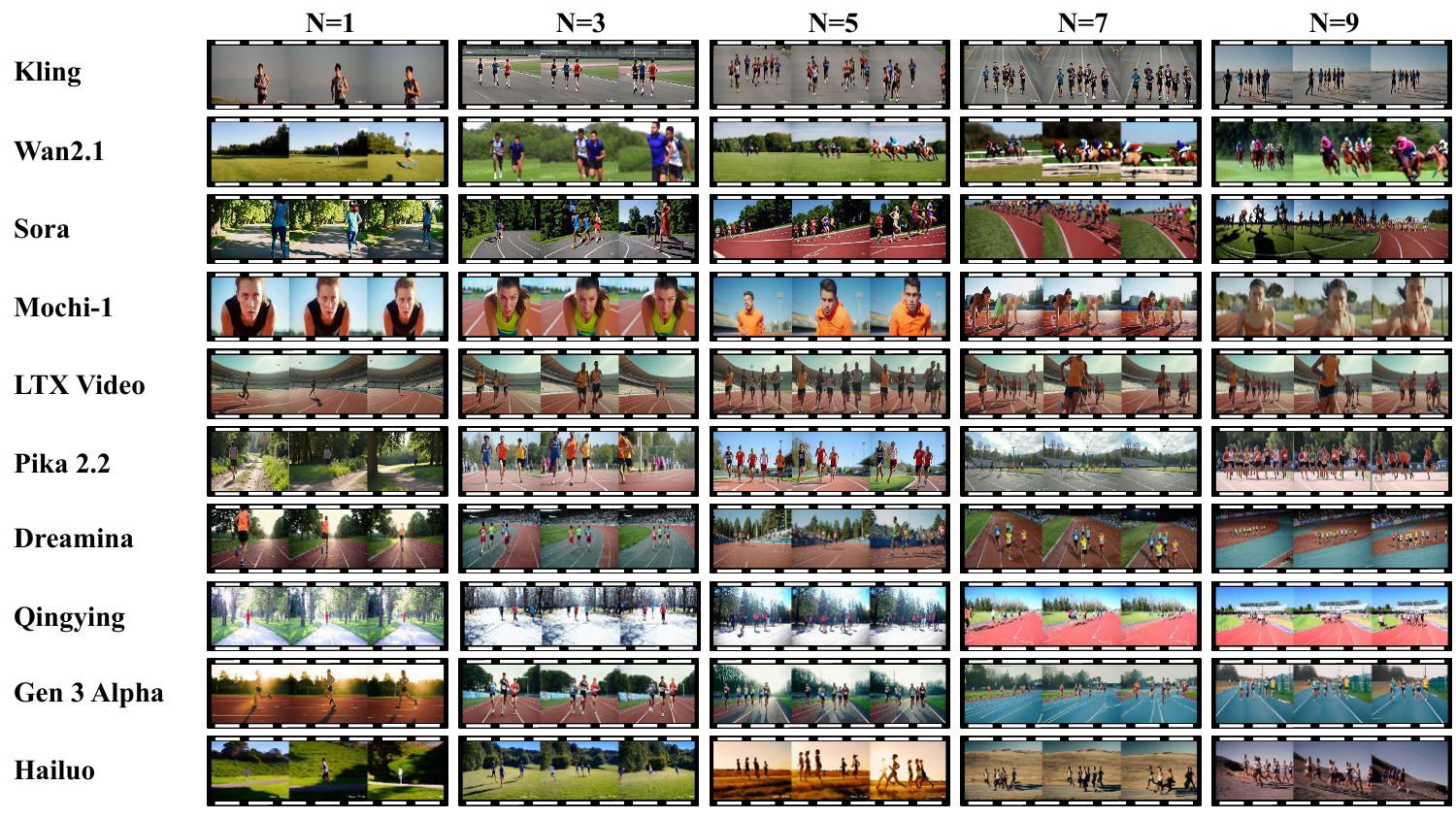}
    \caption{
    {\bf Counting Runners Results on 10 Models}. }
    \label{fig:4_1}
\end{figure}

\begin{figure}[!ht]
    \centering
    \includegraphics[width=1.0\linewidth]{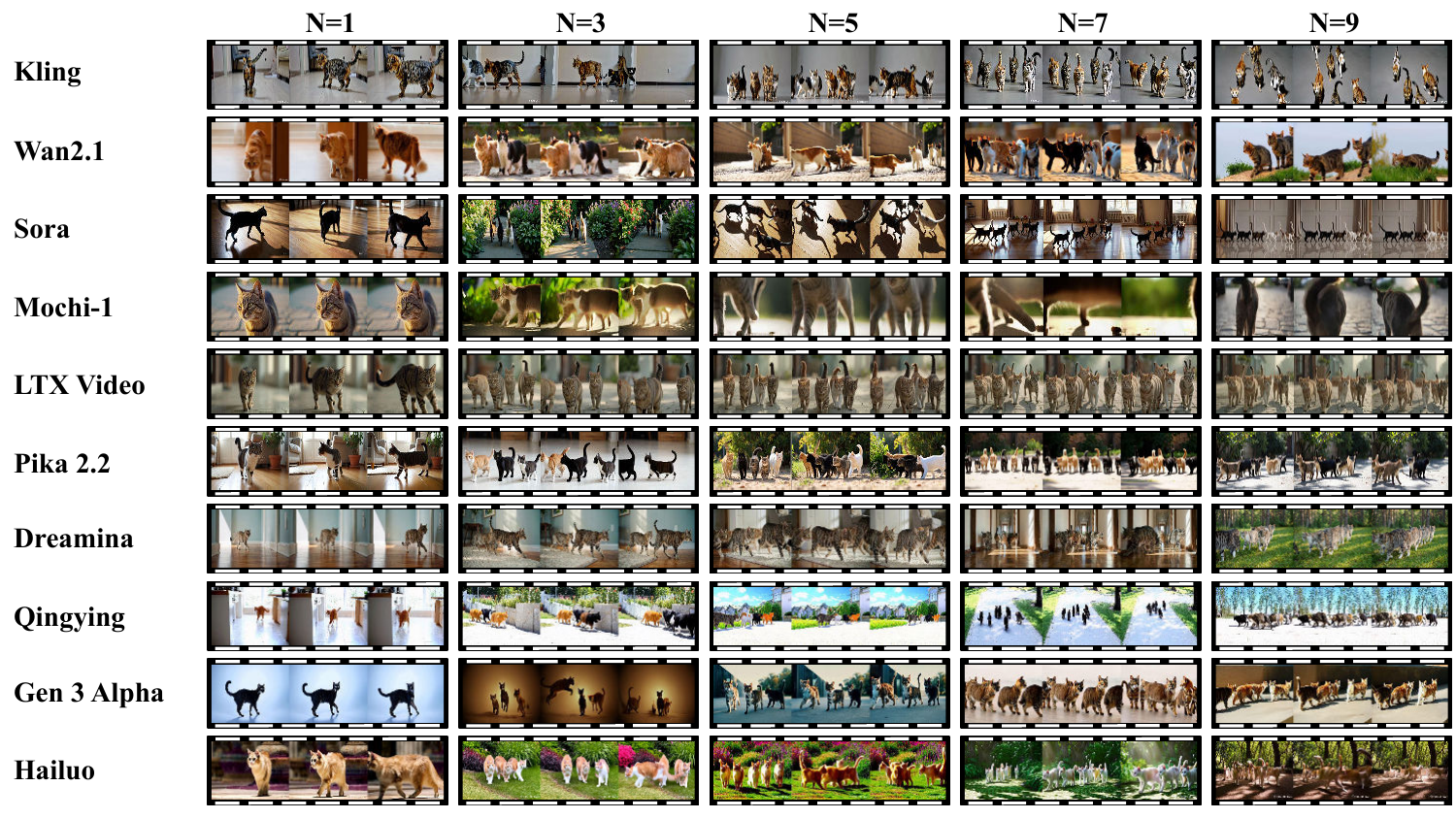}
    \caption{
    {\bf Counting Cats Results on 10 Models}. }
    \label{fig:4_2}
\end{figure}

\begin{figure}[!ht]
    \centering
    \includegraphics[width=1.0\linewidth]{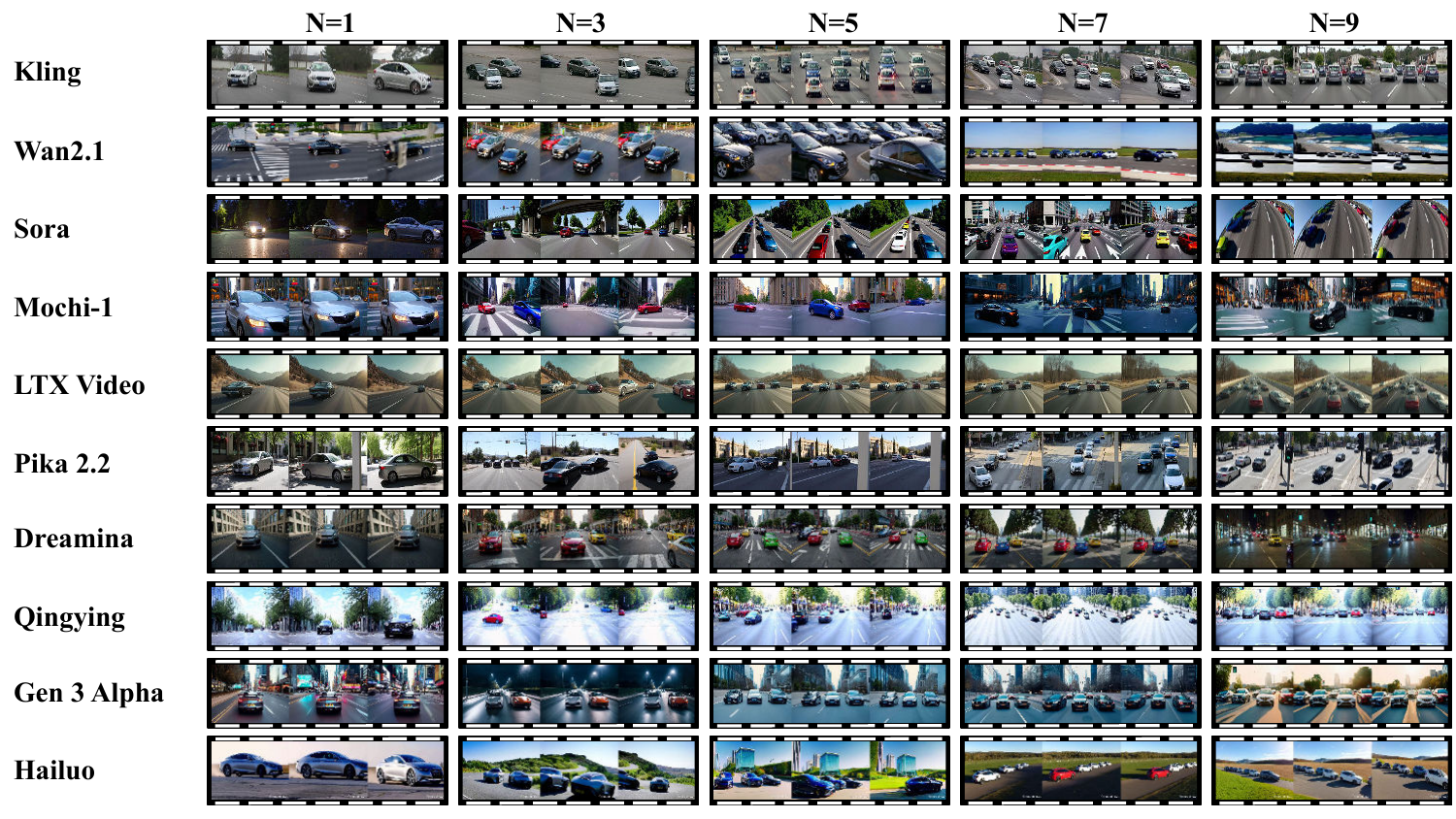}
    \caption{
    {\bf Counting Cars Results on 10 Models}. }
    \label{fig:4_3}
\end{figure}

\begin{figure}[!ht]
    \centering
    \includegraphics[width=1.0\linewidth]{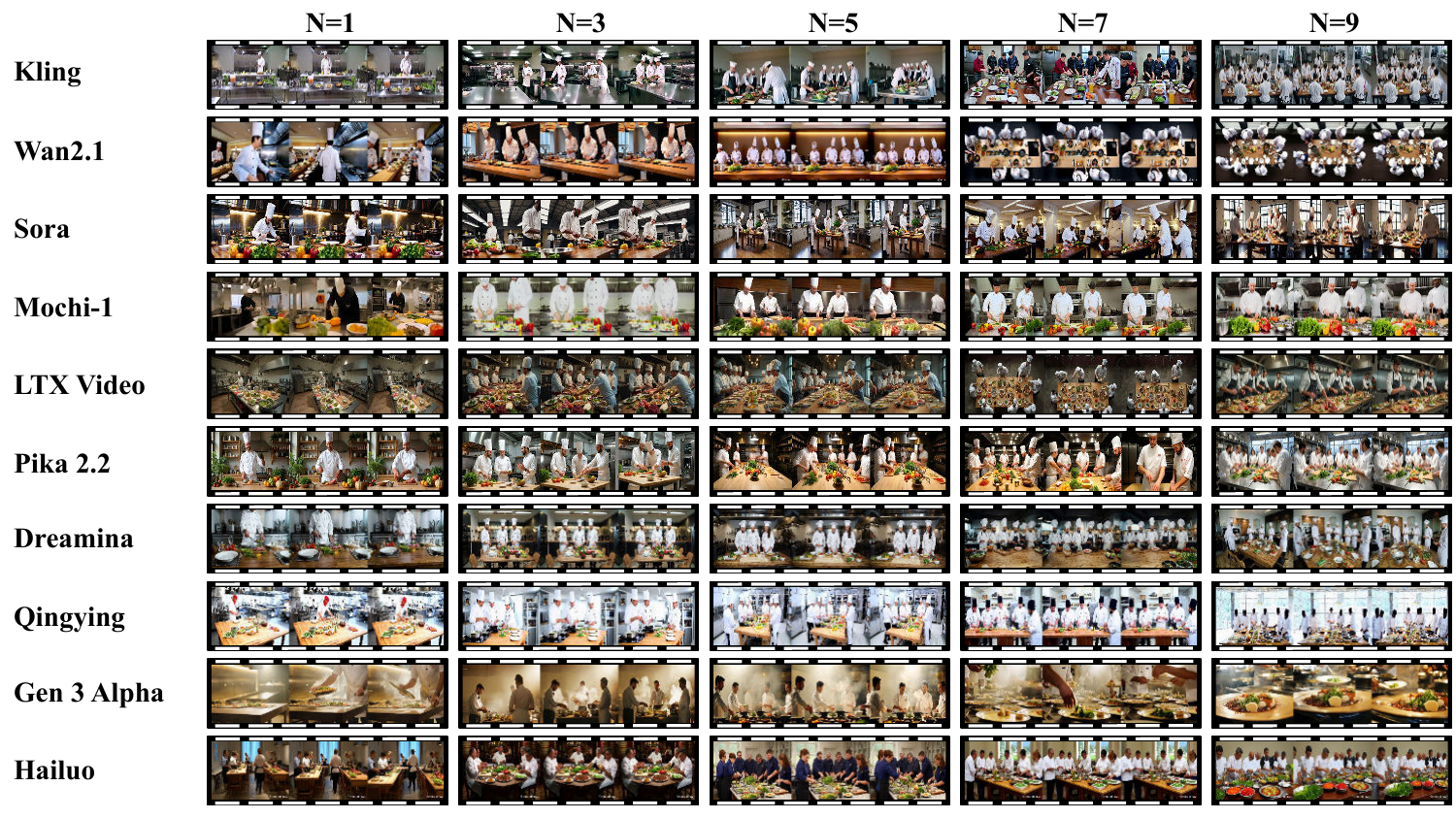}
    \caption{
    {\bf Counting Chefs Results on 10 Models}. }
    \label{fig:4_4}
\end{figure}

\begin{figure}[!ht]
    \centering
    \includegraphics[width=1.0\linewidth]{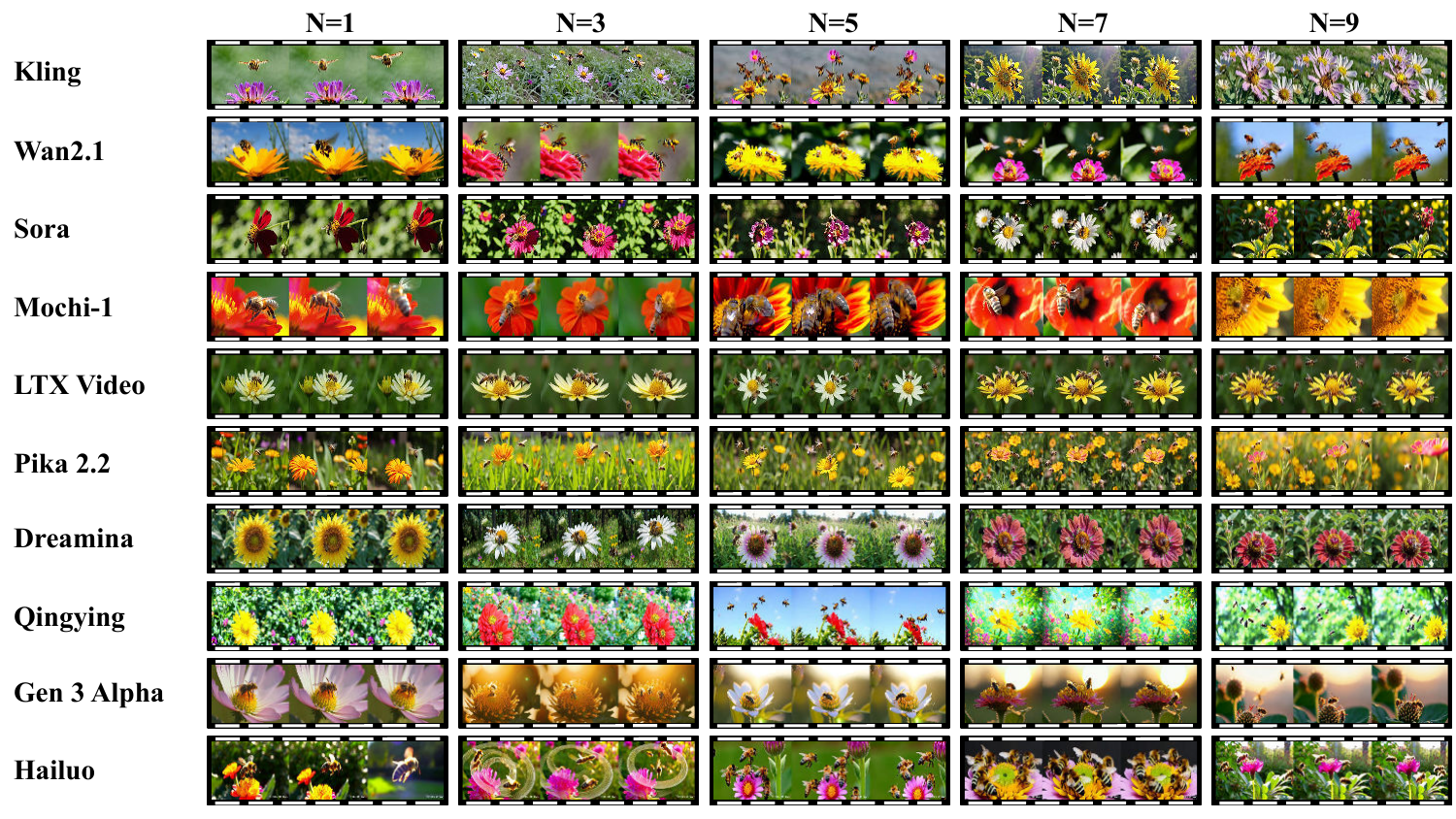}
    \caption{
    {\bf Counting Bees Results on 10 Models}. }
    \label{fig:4_5}
\end{figure}

\begin{figure}[!ht]
    \centering
    \includegraphics[width=1.0\linewidth]{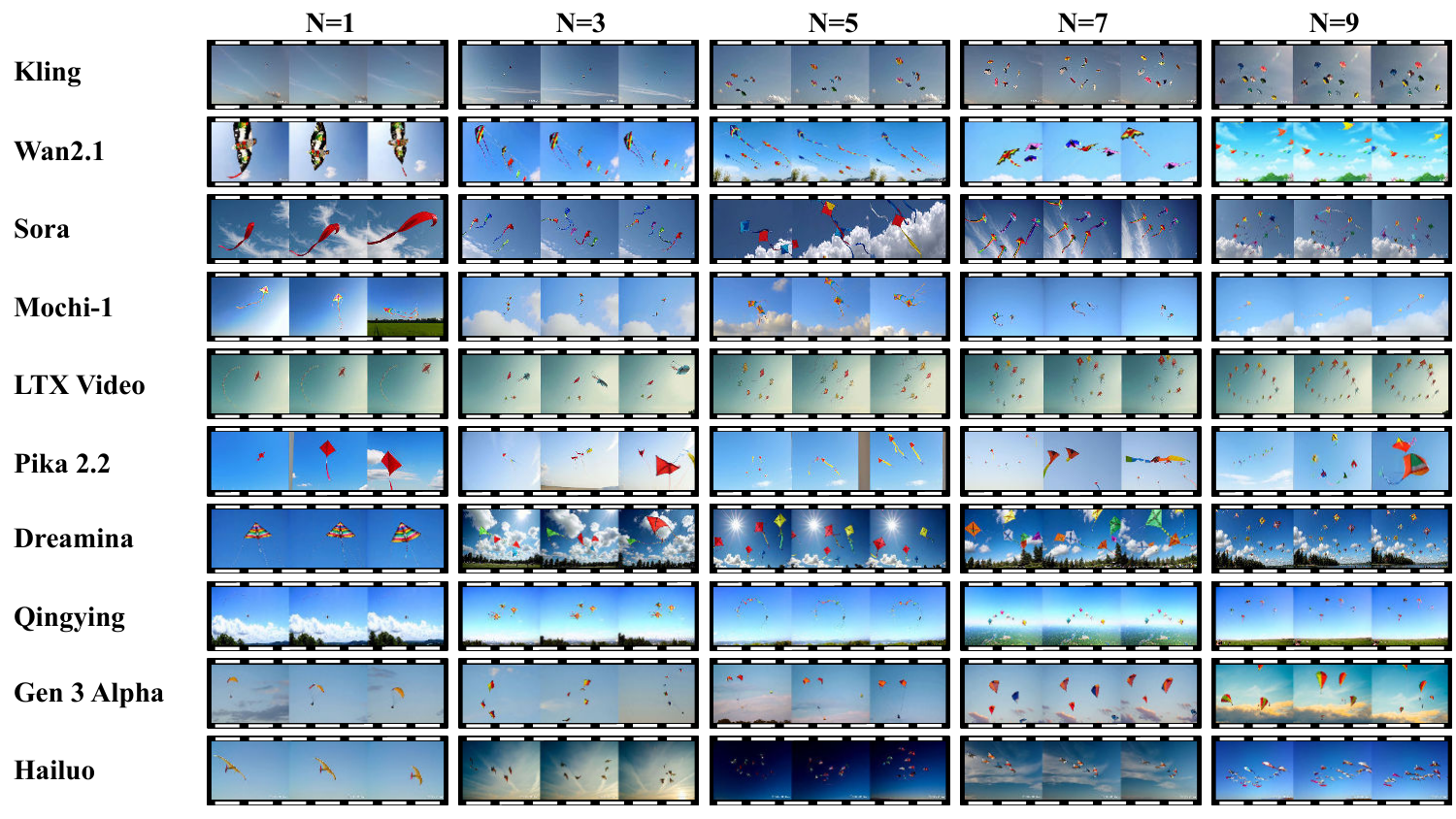}
    \caption{
    {\bf Counting Kites Results on 10 Models}. }
    \label{fig:4_6}
\end{figure}

\begin{figure}[!ht]
    \centering
    \includegraphics[width=1.0\linewidth]{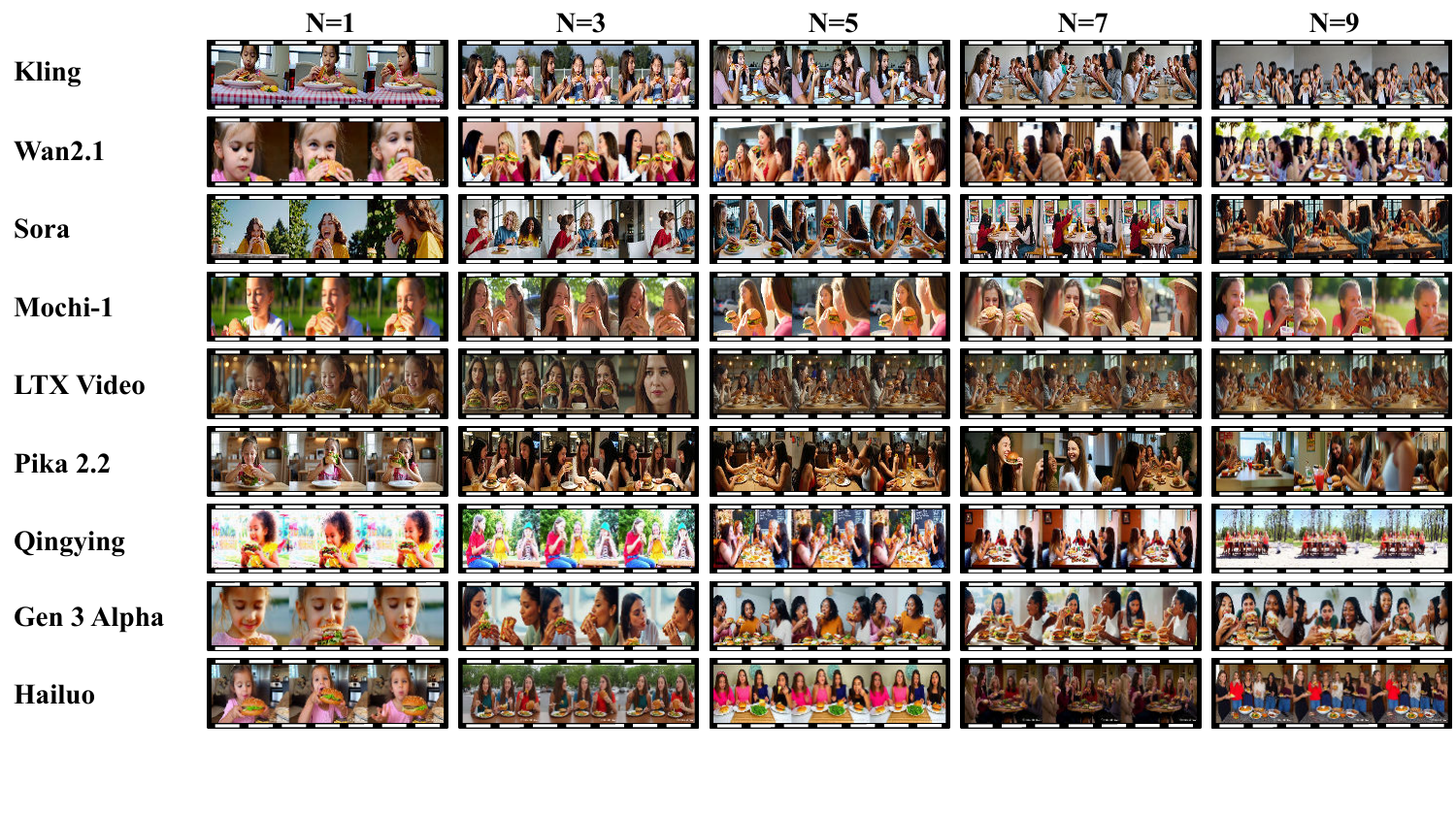}
    \caption{
    {\bf Counting Girls Results on 9 Models}. }
    \label{fig:5_1}
\end{figure}

\begin{figure}[!ht]
    \centering
    \includegraphics[width=1.0\linewidth]{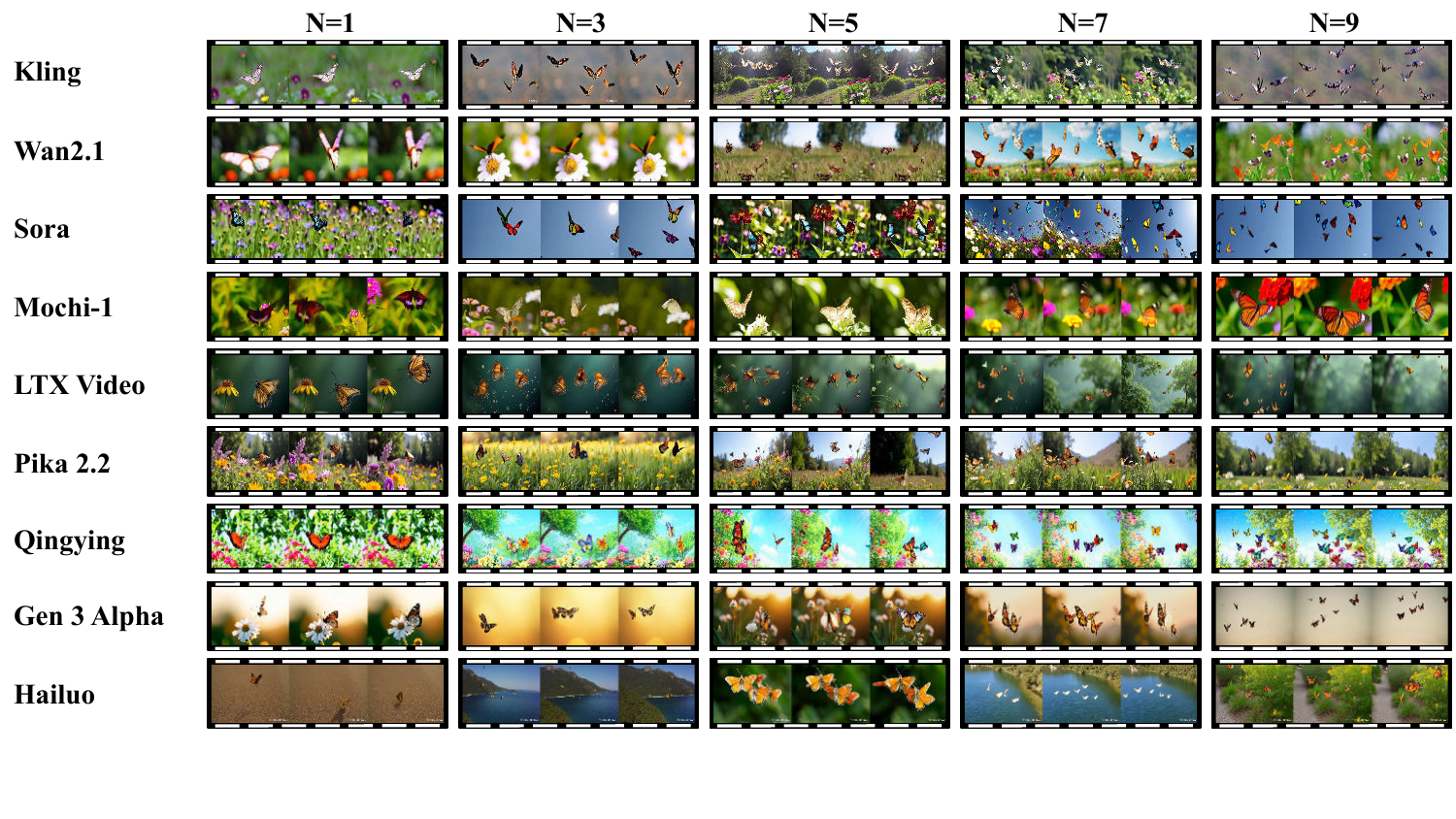}
    \caption{
    {\bf Counting Butterflies Results on 9 Models}. }
    \label{fig:5_2}
\end{figure}

\begin{figure}[!ht]
    \centering
    \includegraphics[width=1.0\linewidth]{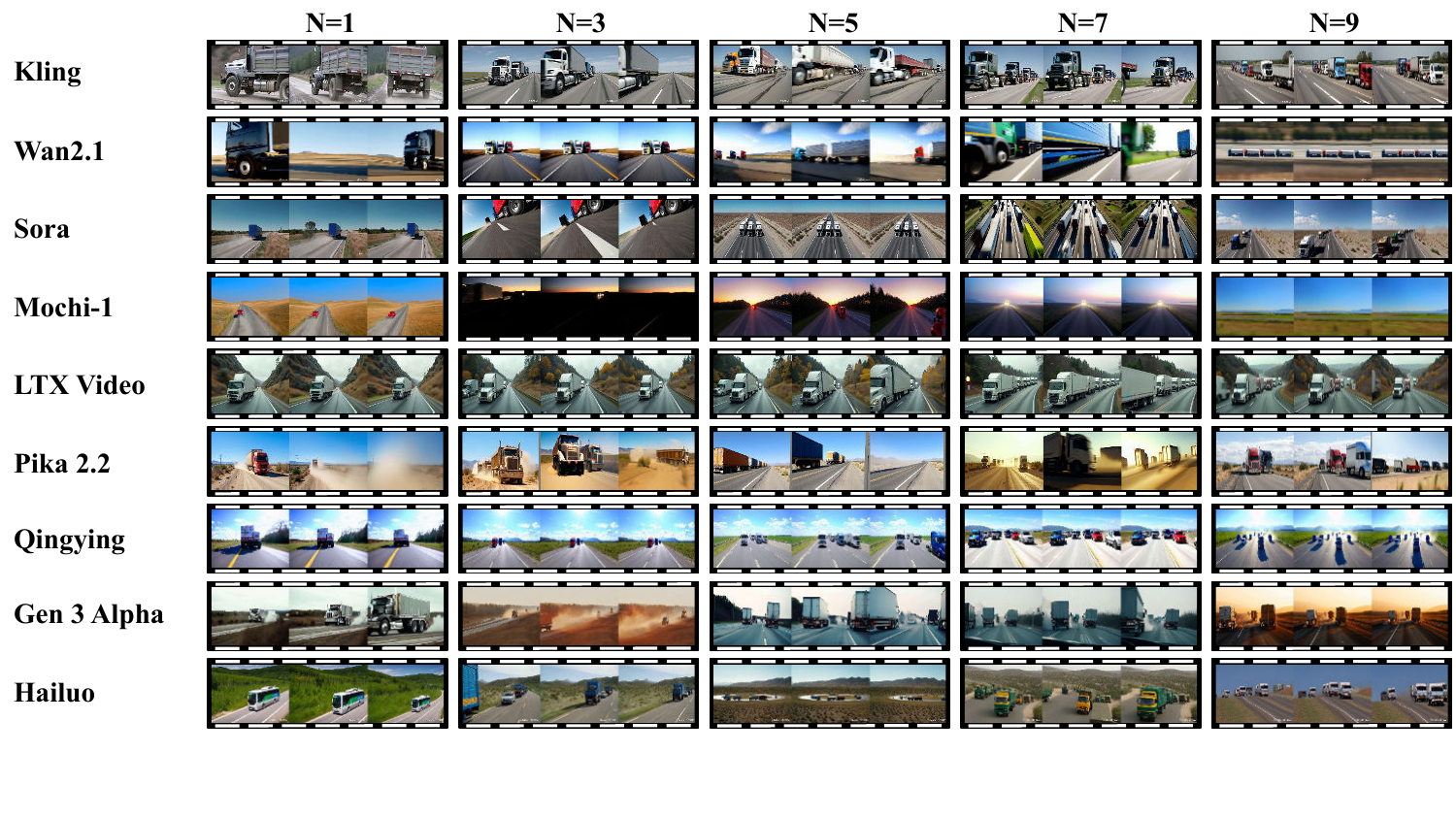}
    \caption{
    {\bf Counting Trucks Results on 9 Models}. }
    \label{fig:5_3}
\end{figure}

\begin{figure}[!ht]
    \centering
    \includegraphics[width=1.0\linewidth]{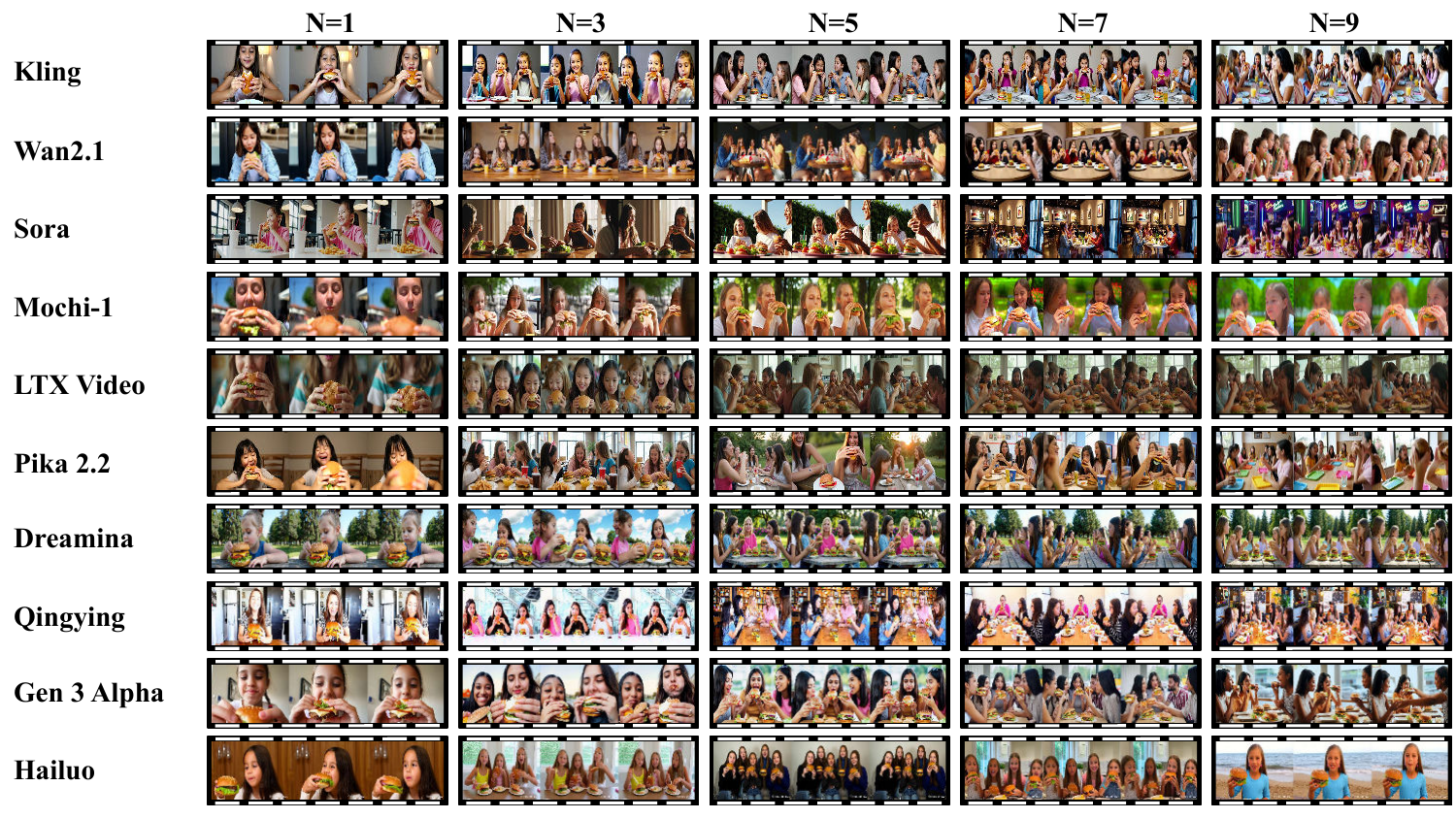}
    \caption{
    {\bf Counting Girls Results on 10 Models}. }
    \label{fig:5_4}
\end{figure}

\begin{figure}[!ht]
    \centering
    \includegraphics[width=1.0\linewidth]{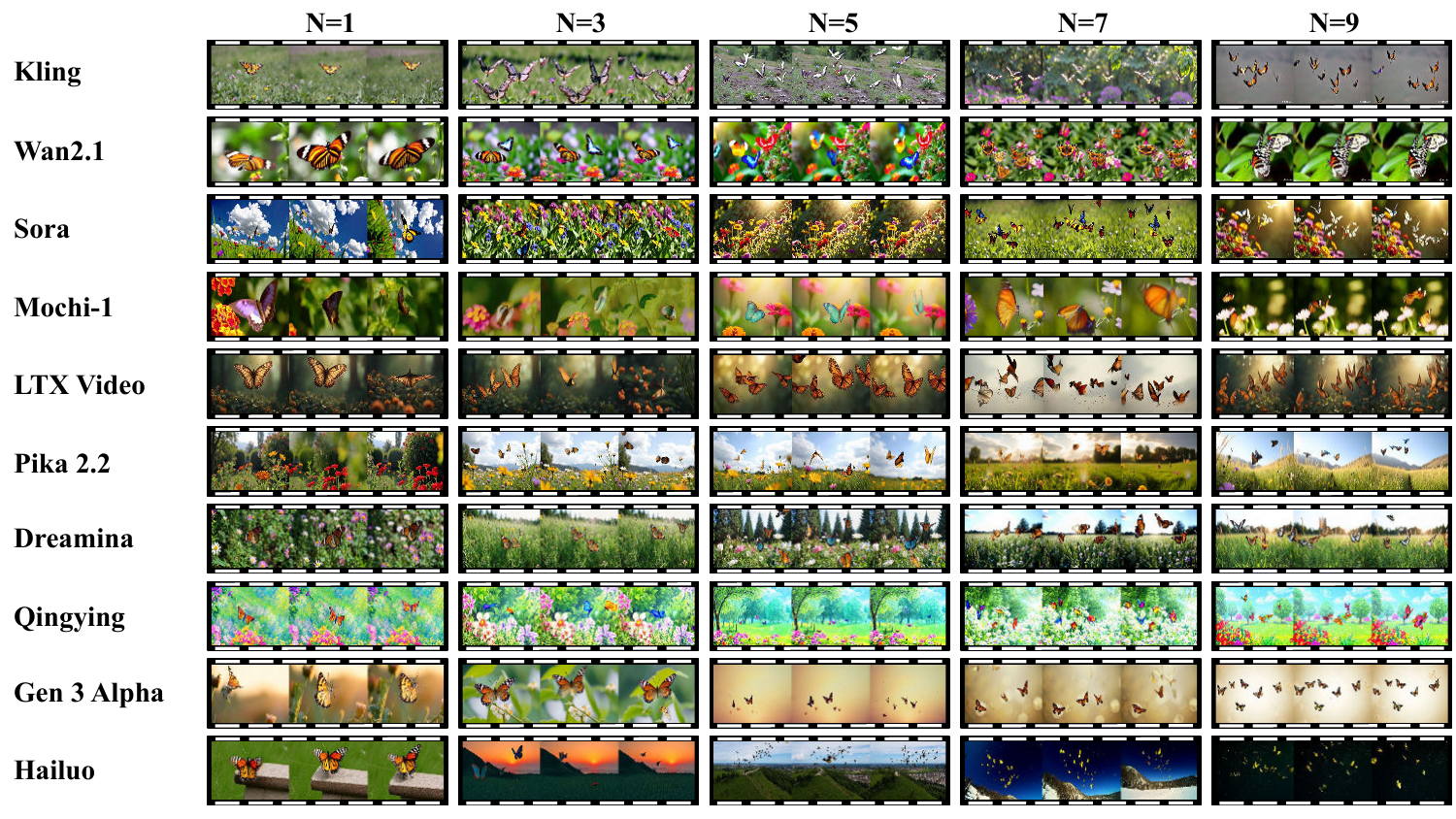}
    \caption{
    {\bf Counting Butterflies Results on 10 Models}. }
    \label{fig:5_5}
\end{figure}

\begin{figure}[!ht]
    \centering
    \includegraphics[width=1.0\linewidth]{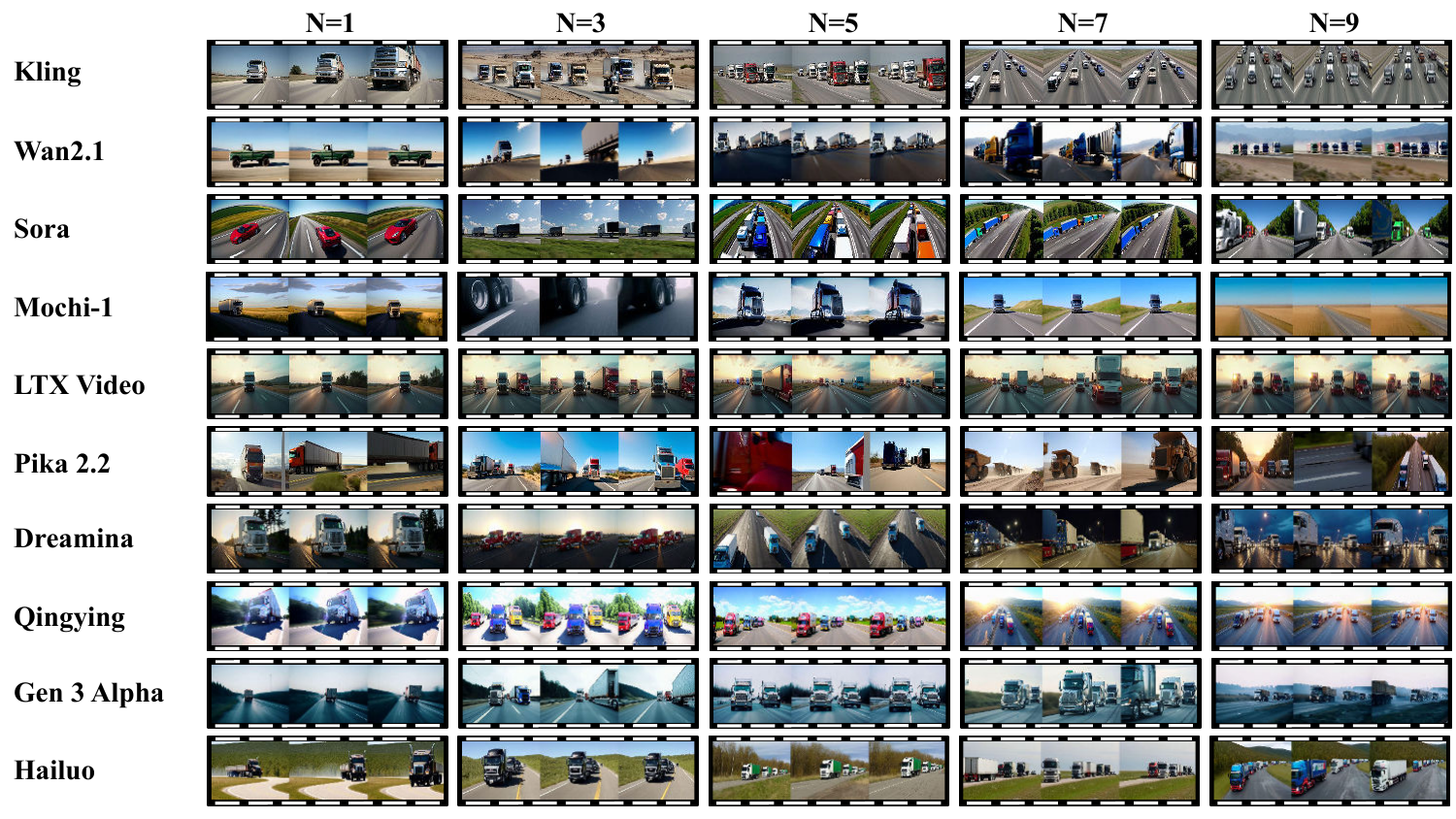}
    \caption{
    {\bf Counting Trucks Results on 10 Models}. }
    \label{fig:5_6}
\end{figure}

\begin{figure}[!ht]
    \centering
    \includegraphics[width=1.0\linewidth]{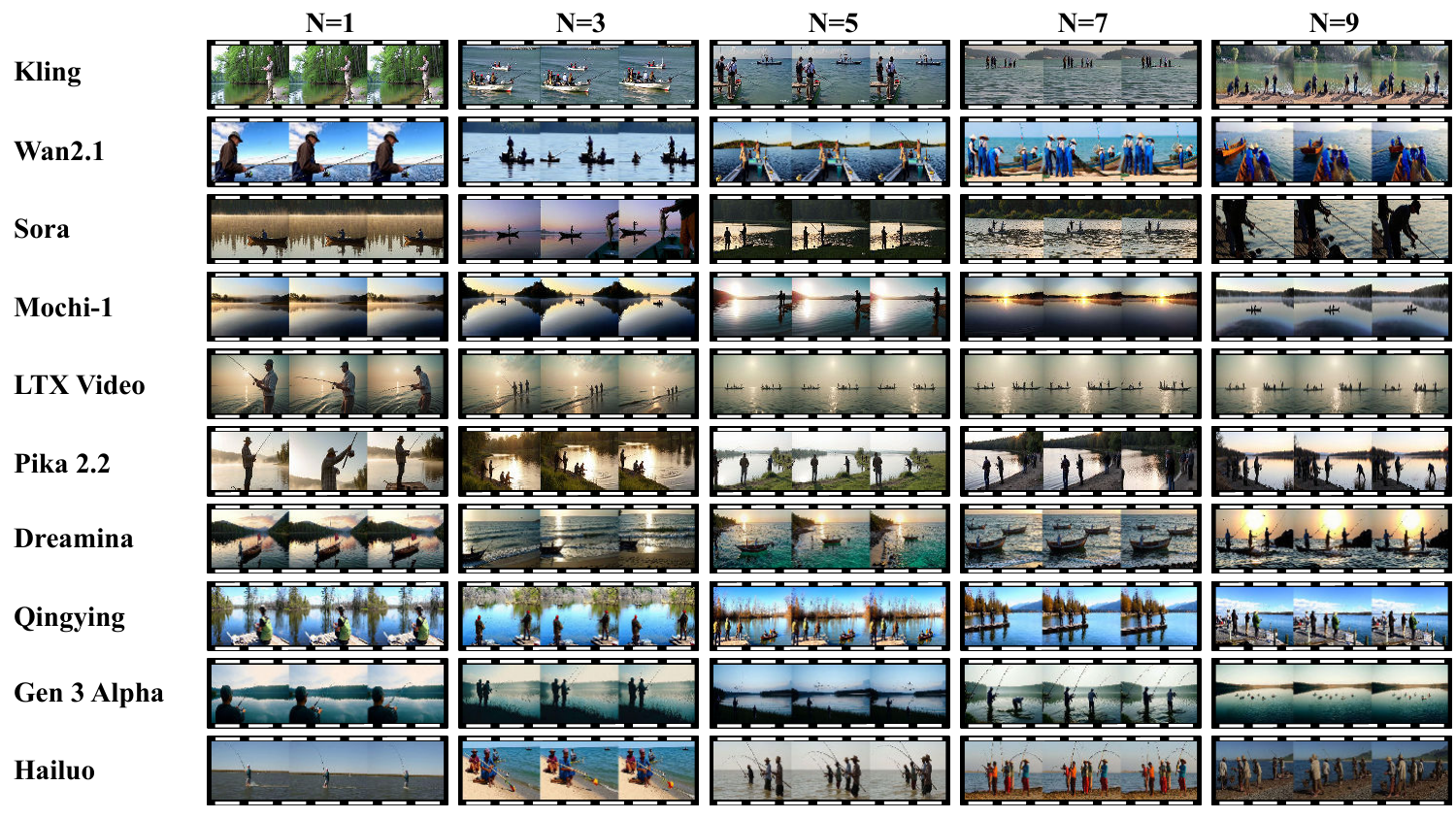}
    \caption{
    {\bf Counting Fishermen Results on 10 Models}. }
    \label{fig:6_1}
\end{figure}

\begin{figure}[!ht]
    \centering
    \includegraphics[width=1.0\linewidth]{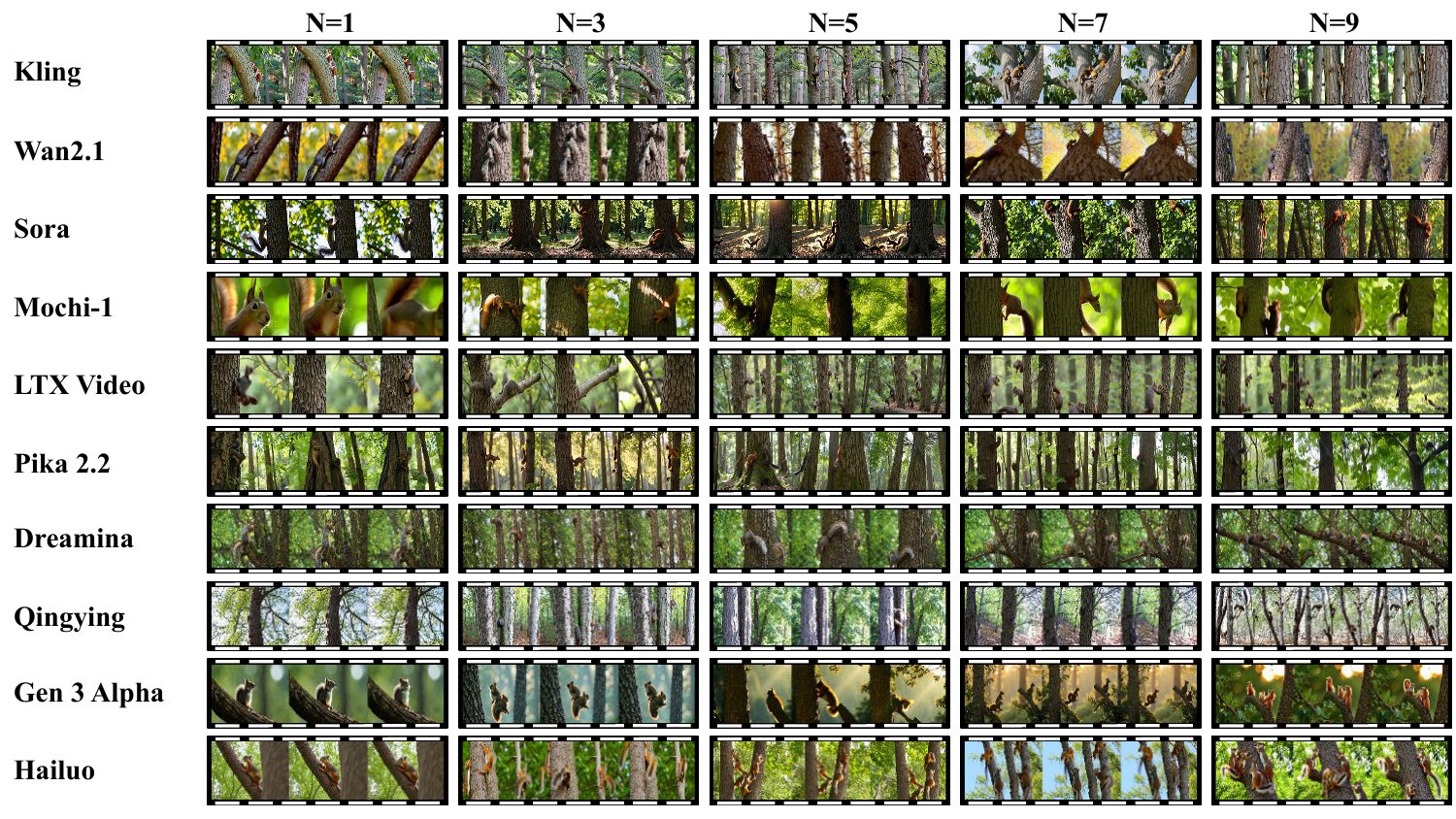}
    \caption{
    {\bf Counting Squirrels Results on 10 Models}. }
    \label{fig:6_2}
\end{figure}

\begin{figure}[!ht]
    \centering
    \includegraphics[width=1.0\linewidth]{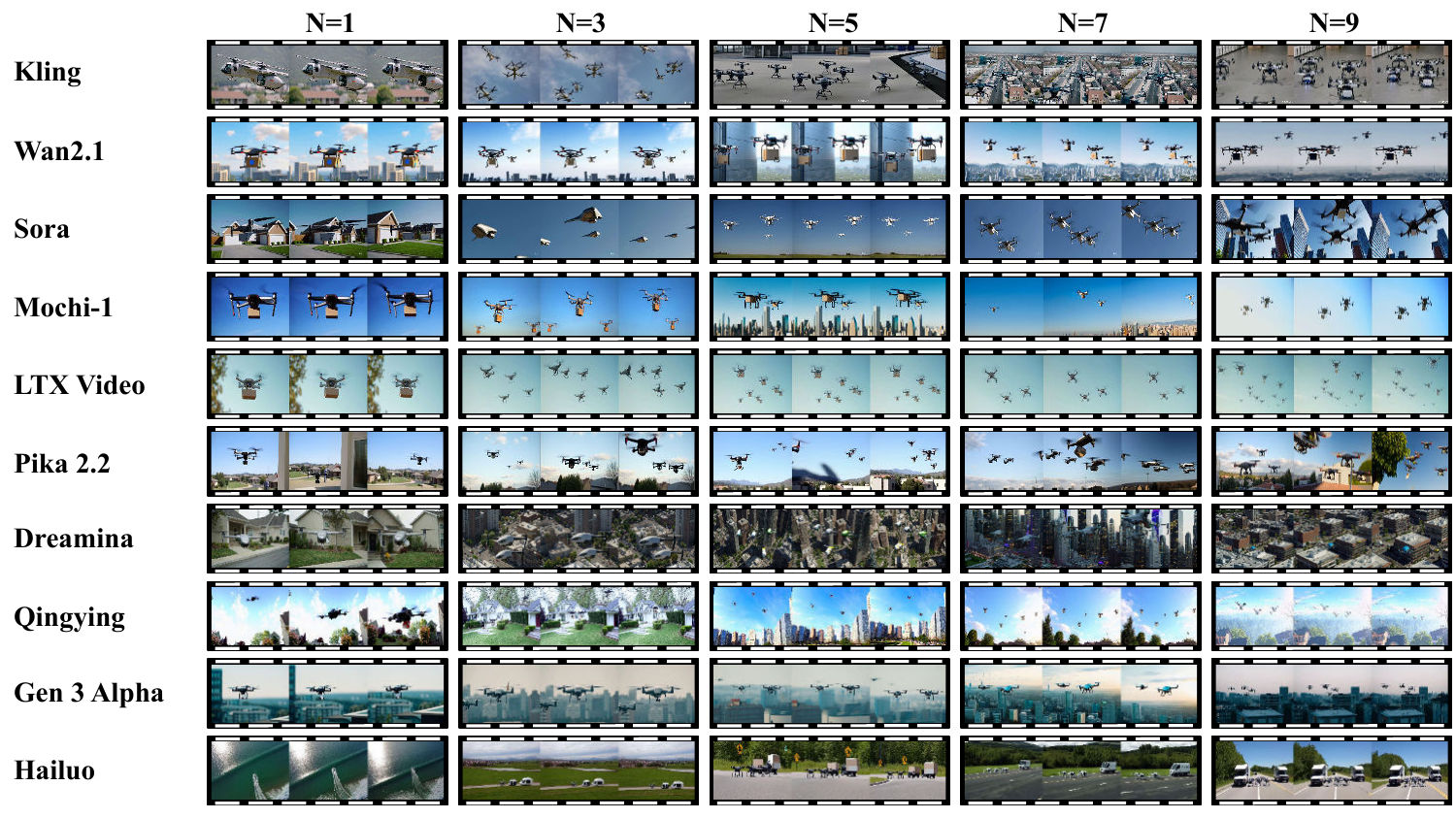}
    \caption{
    {\bf Counting Drones Results on 10 Models}. }
    \label{fig:6_3}
\end{figure}

\begin{figure}[!ht]
    \centering
    \includegraphics[width=1.0\linewidth]{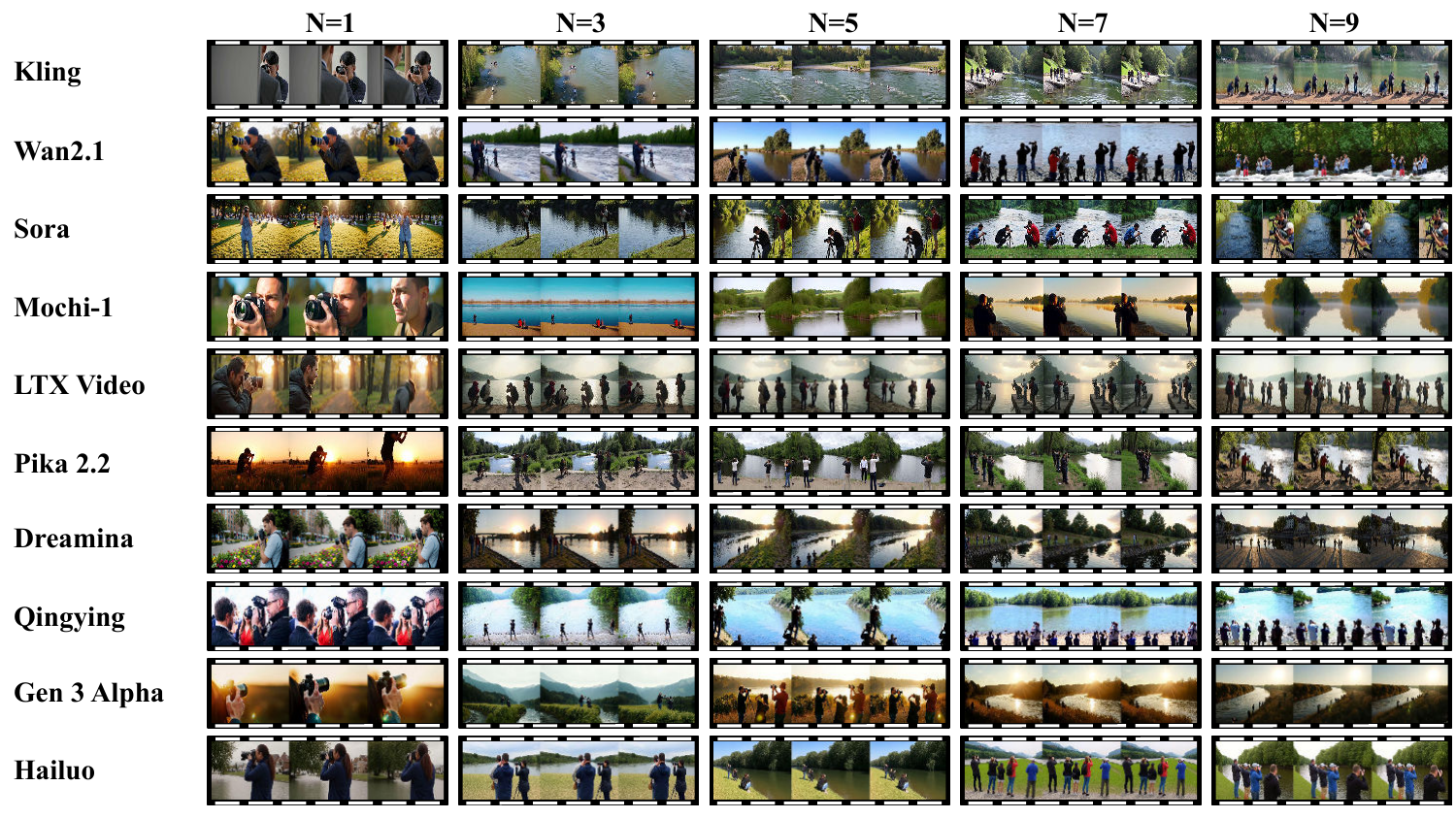}
    \caption{
    {\bf Counting Photographers Results on 10 Models}. }
    \label{fig:6_4}
\end{figure}

\begin{figure}[!ht]
    \centering
    \includegraphics[width=1.0\linewidth]{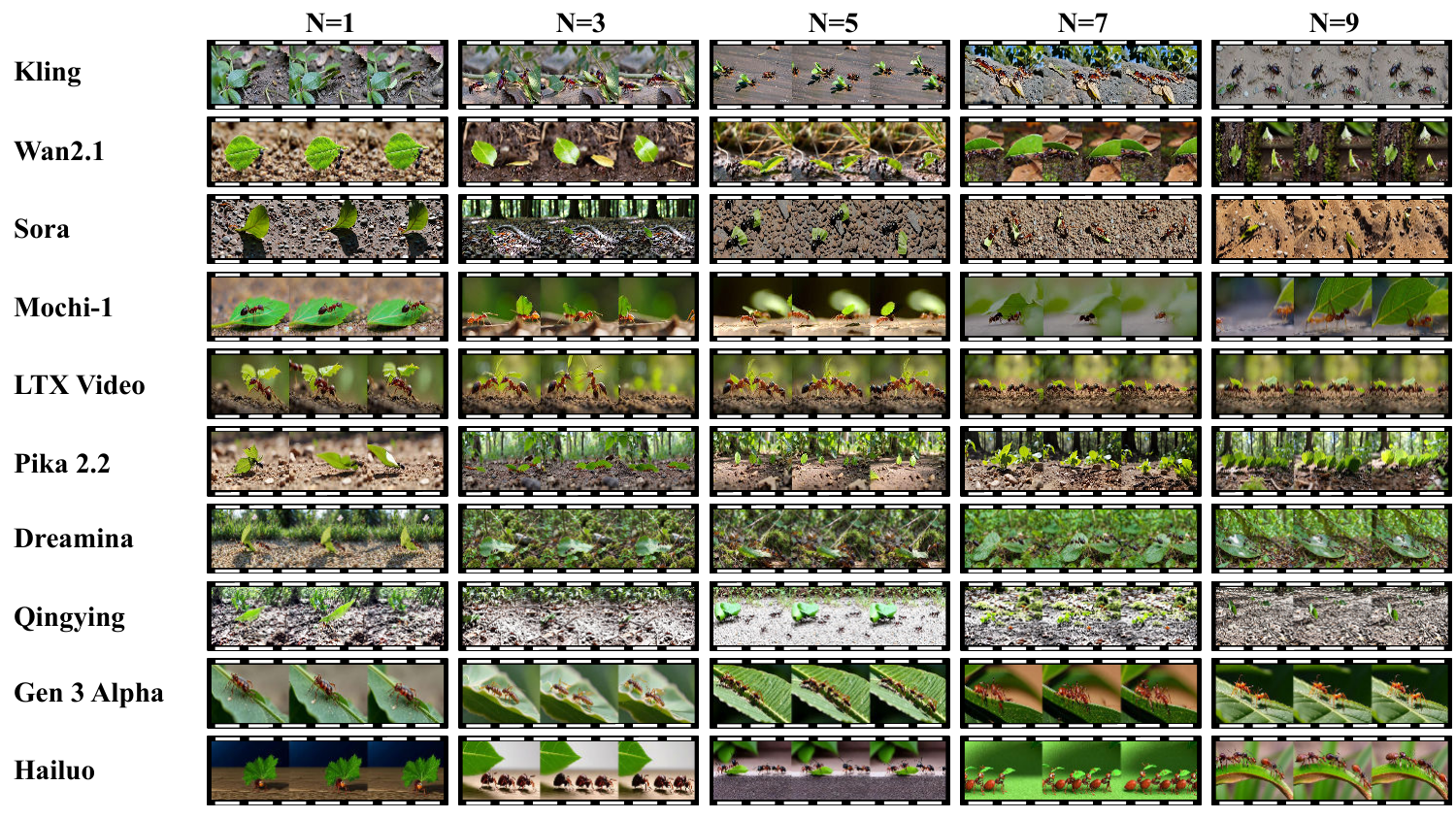}
    \caption{
    {\bf Counting Ants Results on 10 Models}. }
    \label{fig:6_5}
\end{figure}

\begin{figure}[!ht]
    \centering
    \includegraphics[width=1.0\linewidth]{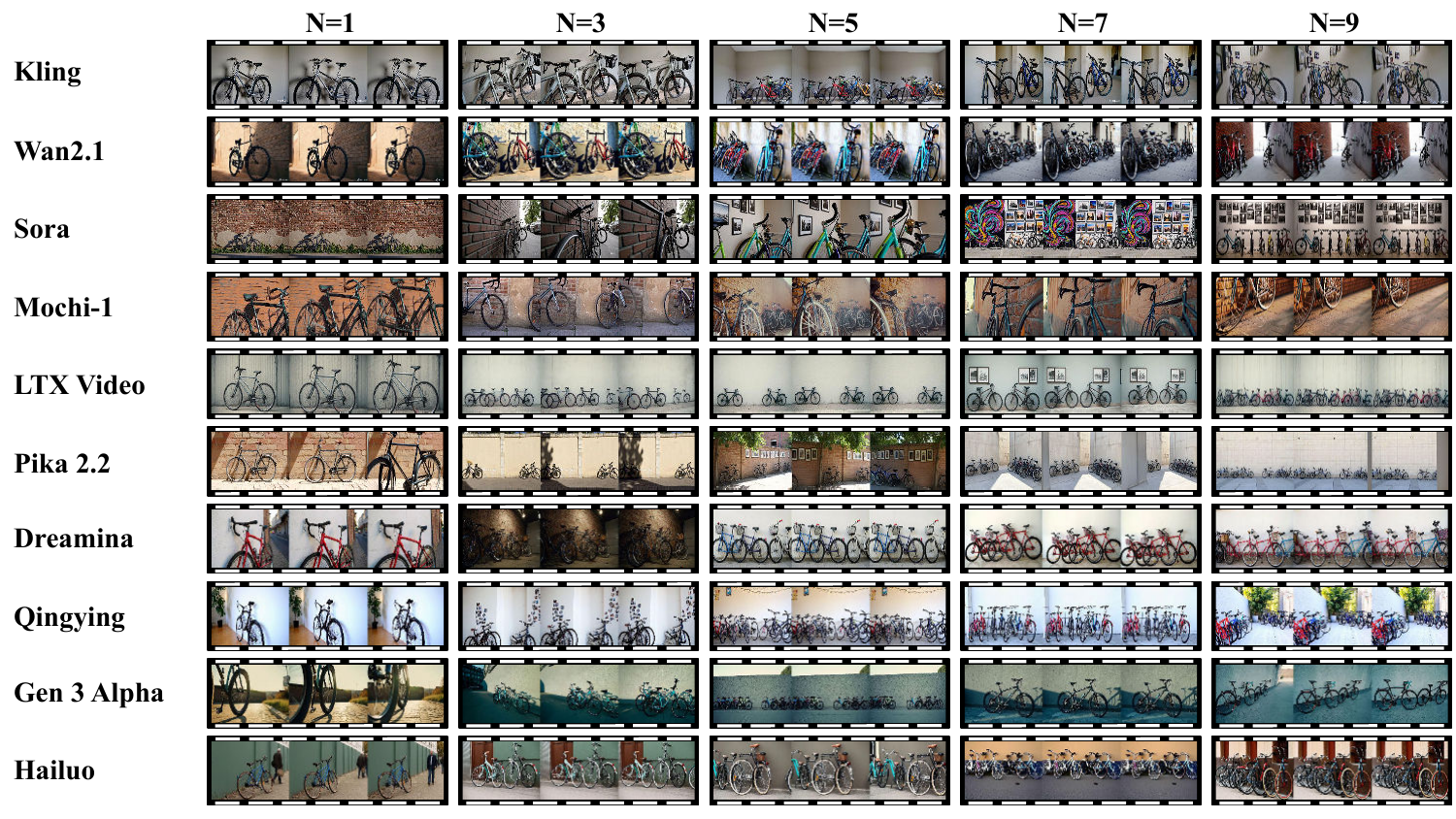}
    \caption{
    {\bf Counting Bicycles Results on 10 Models}. }
    \label{fig:6_6}
\end{figure}

In this subsection, we present a diverse set of video samples generated using the prompts from this benchmark, as illustrated in Figures~\ref{fig:1_1}--\ref{fig:6_6}. For each video sample, three key frames are selected to show its temporal dynamics. Our presented image samples cover all experiments detailed in Section~\ref{sec:experiments} and Appendix~\ref{sec:append_more_experiments}.

